\documentclass{article}

\PassOptionsToPackage{numbers,compress,sort}{natbib}

\usepackage[preprint]{neurips_2026}

\usepackage[utf8]{inputenc}
\usepackage[T1]{fontenc}
\usepackage{hyperref}
\usepackage{url}
\usepackage{booktabs}
\usepackage{amsfonts}
\usepackage{amsmath}
\usepackage{nicefrac}
\usepackage{microtype}
\usepackage{xcolor}
\usepackage{colortbl}
\definecolor{notrun}{rgb}{1.0,0.92,0.92}
\definecolor{anomaly}{rgb}{1.0,0.85,0.55}
\definecolor{corcolor}{rgb}{1, 0.95, 0.85}     
\definecolor{oodcolor}{rgb}{0.85, 0.98, 0.80}  
\definecolor{imbcolor}{rgb}{0.95, 0.90, 0.95}  
\definecolor{faircolor}{rgb}{0.85, 0.95, 1}    
\definecolor{intcolor}{rgb}{1, 0.85, 0.85}     
\usepackage{tcolorbox}
\DeclareRobustCommand{\colorhl}[2]{%
  {\setlength{\fboxsep}{2pt}\colorbox{#1}{\strut#2}}%
}
\NewEnviron{findingbox}[1][corcolor]{%
  \begin{tcolorbox}[colback=#1,colframe=black,arc=4pt,boxsep=0.5pt,
                    left=8pt,right=8pt,top=4pt,bottom=4pt,
                    before skip=4pt,after skip=4pt]
  \BODY
  \end{tcolorbox}
}
\usepackage{graphicx}
\usepackage{multirow}
\usepackage{longtable}
\usepackage{subcaption}
\usepackage{array}
\usepackage{float}
\usepackage{xspace}
\providecommand{\FloatBarrier}{\clearpage}

\newcommand{\benchname}{GRL-Safety\xspace}

\title{On the Safety of Graph Representation Learning}

\author{%
    \textbf{Xiaoguang Guo\textsuperscript{1}},
    \textbf{Zehong Wang\textsuperscript{2}},
    \textbf{Ziming Li\textsuperscript{1}},
    \textbf{Shawn Spitzel\textsuperscript{1}},
    \textbf{Soonwoo Kwon\textsuperscript{1}},
    \textbf{Tianyi Ma\textsuperscript{2}}, \\
    \textbf{Yanfang Ye\textsuperscript{2}},
    \textbf{Chuxu Zhang\textsuperscript{1}\thanks{Corresponding author.}}\\
    \small\texttt{\{xiaoguang.guo,ziming.li,shawn.spitzel,soonwoo.kwon,chuxu.zhang\}@uconn.edu} \\
    \small\texttt{\{zwang43,tma2,yye7\}@nd.edu} \\
    \textsuperscript{1}{University of Connecticut},
    \textsuperscript{2}{University of Notre Dame}
}

\begin{document}

\maketitle

\begin{abstract}
Graph representation learning (GRL) has evolved from topology-only graph
embeddings to task-specific supervised GNNs, and more recently to reusable
representations and graph foundation models (GFMs). However, existing
evaluations mainly measure clean transfer, adaptation, and task coverage. It
remains unclear whether GRL methods stay reliable when deployment stresses
affect graph signals, graph contexts, label support, structural groups, or
predictive evidence. We introduce \benchname, a multi-axis safety evaluation
benchmark for GRL. \benchname evaluates twelve representative methods, spanning
topology-only embedding methods, supervised GNNs, self-supervised graph models,
and GFMs, on twenty-five graph datasets under standardized evaluation conditions
while preserving method-native adaptation. The evaluation covers five safety
axes: corruption robustness, OOD generalization, class imbalance, fairness, and
interpretation, with per-axis and subcondition reporting rather than a single
aggregate score. Our analysis yields three cross-axis insights that can inspire
future research. First, safety behavior is shaped by the interaction between
representation design and the stressed graph factor, rather than by method
family alone. Second, foundation-era methods show axis-specific strengths
rather than broad safety dominance.
Third, several deployment regimes remain difficult even for the best evaluated
method, revealing capability gaps that require new robustness, adaptation, or
training objectives beyond model selection. The benchmark, evaluation protocols,
and code are available at:
\url{https://github.com/GXG-CS/GRL-Safety}.
\end{abstract}
\section{Introduction}
\label{sec:introduction}

Foundation models (FMs)~\citep{bommasani2021foundation} have shifted
machine learning in natural language
processing~\citep{devlin2019bert} and computer
vision~\citep{radford2021clip} from task-specific training toward
transferable representations that can be adapted across tasks and
domains. Graph learning~\citep{wu2020comprehensive}, which underlies
applications including molecular design~\citep{stokes2020antibiotic},
recommendation systems~\citep{ying2018pinsage}, and financial
networks~\citep{dou2020caregnn}, is undergoing a parallel shift.
In particular, graph representation learning (GRL) has progressed from
topology-only graph embeddings~\citep{perozzi2014deepwalk,grover2016node2vec}
and task-specific supervised graph neural networks
(GNNs)~\citep{kipf2017gcn,velickovic2018gat,hamilton2017sage}
toward transferable graph representations, including self-supervised
pretrained models~\citep{thakoor2022bgrl,hou2022graphmae} and graph
foundation models (GFMs)~\citep{gfmsurvey2025}, that can generalize
across graphs, domains, and task formats.

As these representations are deployed in diverse downstream settings,
deployment conditions expose safety risks that clean-setting
performance does not capture~\citep{dai2024trustworthy}. These risks
can enter through graph-specific channels, including corrupted
attributes or edges~\citep{zheng2021grb}, distribution shifts across
graph contexts~\citep{gui2022good}, skewed label
support~\citep{ma2025cilg}, structural or demographic group
disparity~\citep{dong2024fairgraph}, and unreliable predictive
evidence~\citep{agarwal2023graphxai}. In financial graphs, for
example, corrupted links or rare fraud labels can affect risk
prediction in ways that clean aggregate performance may hide
~\citep{dou2020caregnn}; in scientific or molecular graphs,
unreliable attribution evidence can undermine the trustworthiness of
model outputs~\citep{amara2022graphframex}. It remains an open
question how modern GRL methods behave across these
representative safety-relevant axes under comparable conditions.

Existing evaluation efforts only partially answer this question.
GRL and GFM benchmarks have made substantial progress
in measuring clean-setting performance, transfer, adaptation, and task
coverage across datasets, tasks, and
domains~\citep{hu2020ogb,xu2024graphfm,chen2024tsgfm,yu2026evalgfm}.
In parallel, trustworthy GRL evaluation has developed
axis-specific benchmarks and protocols for many of these
concerns~\citep{zheng2021grb,gui2022good,ma2025cilg,dong2024fairgraph,agarwal2023graphxai}.
However, these two lines remain disconnected: GRL
benchmarks typically do not evaluate safety-relevant axes, while
safety evaluation efforts are often organized one axis at a time,
under narrower method families or different protocols. What remains
missing is a unified, multi-axis safety evaluation of the modern GRL spectrum under comparable conditions.

Building such an evaluation requires handling several sources of
heterogeneity. First, safety-relevant axes stress different parts of
the GRL pipeline, so their results should be reported
by axis and subcondition rather than collapsed into a single
score~\citep{liang2022helm,wang2023decodingtrust}. Second, GRL methods expose different input requirements and
adaptation interfaces: topology-only embeddings consume only graph
structure, supervised GNNs consume features and topology,
self-supervised models add pretraining objectives, and GFMs introduce
broader adaptation mechanisms. A comparable benchmark must therefore
standardize the evaluation substrate while preserving each method's
native design~\citep{xu2024graphfm,chen2024tsgfm,yu2026evalgfm}.
Finally, because safety risks can manifest differently across graph
domains and task formats, the benchmark must cover diverse datasets
rather than a single graph family~\citep{hu2020ogb,gui2022good,feng2024taglas}.

Guided by these requirements, we introduce \benchname, a multi-axis
safety benchmark for GRL. \benchname
evaluates twelve representative GRL methods on
twenty-five graph datasets spanning diverse domains and task formats.
Under standardized evaluation conditions, feature-consuming methods use
shared text-attribute embeddings, topology-only embedding methods use
shared graph structure on compatible task formats, and methods with a
pretraining stage share a common source corpus while preserving their
original pretraining objectives and native downstream adaptation
pipelines. Overall, our contributions of this work are:
\begin{itemize}
    \item \textbf{Benchmark scope.} \benchname positions safety
    evaluation at the level of GRL, covering
    topology-only graph embeddings, supervised GNNs, self-supervised
    GRL methods, and GFMs across diverse graph tasks
    and domains.

    \item \textbf{Multi-axis safety protocol and evaluation.} \benchname evaluates
    five deployment-relevant safety axes: corruption robustness, OOD
    generalization, class imbalance, fairness, and interpretation. It
    reports per-axis and subcondition results rather than collapsing
    graph safety into a single aggregate score.

    \item \textbf{Cross-axis empirical insights.} \benchname reveals
    that safety behavior depends on the interaction between
    representation design and the stressed graph factor, that
    foundation-era methods show axis-specific strengths rather than
    broad safety dominance, and that some deployment regimes expose capability gaps
    beyond model selection.
\end{itemize}

\section{Related Work}
\label{app:related}

\paragraph{Graph representation learning.}
Graph representation learning (GRL) has progressed from topology-only graph
embeddings and task-specific message-passing models toward transferable
pretrained representations and graph foundation models
(GFMs)~\citep{wu2020comprehensive,gfmsurvey2025}. Early topology-only
embedding methods learn node representations directly from graph
structure, through random-walk contexts~\citep{perozzi2014deepwalk,grover2016node2vec},
proximity preservation~\citep{tang2015line}, or structural-role
similarity~\citep{ribeiro2017struc2vec}. Supervised GNNs learn
task-specific representations by iteratively aggregating neighborhood
information and training end-to-end with downstream
labels~\citep{kipf2017gcn,velickovic2018gat,hamilton2017sage,zhang2019heterogeneous}.
Self-supervised GRL relaxes this
task-specific training pattern through pretraining objectives such as
mutual-information maximization~\citep{velickovic2019dgi},
contrastive learning with augmentations~\citep{you2020graphcl,ju2023multi},
bootstrap latent prediction~\citep{thakoor2022bgrl}, and masked
reconstruction~\citep{hou2022graphmae,tian2023heterogeneous,wang2025beyond,wanggenerative}, so that learned
representations can be adapted to downstream tasks. More recently,
GFMs push this transferability goal further through broader design and
adaptation interfaces, including task and model
unification~\citep{liu2024ofa,wang2024gft,wang2024git}, graph
prompting~\citep{sun2022gppt,fang2023gpf}, cross-graph pretraining and
domain alignment~\citep{qiu2020gcc,he2025unigraph2,xia2024anygraph,yuan2024graphcsl,stemgnn2026},
in-context or zero-shot adaptation~\citep{huang2023prodigy,li2024zerog},
and graph-language interfaces~\citep{chen2024llaga,tang2024graphgpt}.
Existing evaluations of this spectrum mainly measure clean-setting
performance, adaptation, and task-format coverage on standard datasets
and splits~\citep{hu2020ogb,xu2024graphfm,chen2024tsgfm,yu2026evalgfm}.

\paragraph{Safety evaluation in GRL.}
Complementing evaluations of clean-setting performance, safety-oriented
evaluation work~\cite{wang2026safety} studies representative safety-relevant axes in GRL. For corruption robustness, GRB~\citep{zheng2021grb} and
DeepRobust~\citep{li2020deeprobust} provide benchmark and toolkit
support for feature and topology perturbations, while
NoisyGL~\citep{wang2024noisygl} provides a unified protocol for graph
learning under label noise. For out-of-distribution (OOD)
generalization, GOOD~\citep{gui2022good,wang2025grm} introduces controlled
covariate and concept shifts under a per-method evaluation protocol
spanning node and graph classification. For class imbalance, recent
surveys~\citep{ma2025cilg} consolidate protocols and metrics for
imbalanced node classification, while works such as
GraphSMOTE~\citep{zhao2021graphsmote} and
GraphENS~\citep{park2022graphens} commonly evaluate under controlled
imbalance-ratio settings. For fairness, the FairGraph
benchmark~\citep{dong2024fairgraph} and the PyGDebias
toolkit~\citep{dong2024pygdebias} consolidate datasets, metrics, and
evaluations of demographic and structural group disparity. For
interpretation, GraphXAI~\citep{agarwal2023graphxai},
GraphFramEx~\citep{amara2022graphframex}, and
GInX-Eval~\citep{amara2023ginxeval} provide complementary evaluation
frameworks for saliency, post-hoc-explainer fidelity, and
in-distribution explanation quality. Broader trustworthy graph
learning work also covers additional reliability concerns beyond
these representative axes, including adversarial and prompt-level
robustness for graph-LLM systems~\citep{dai2024trustworthy,zhang2025trustglm,yuan2024dragon}.
These efforts establish important axis-specific evaluation protocols,
but they are usually organized one axis at a time, leaving limited
evidence about how safety behavior changes across axes. Moreover,
they are often developed under different task scopes, input
representations, evaluation definitions, and method families, leaving
open how the GRL spectrum behaves under comparable safety
evaluation conditions.
\vspace{-1.0em}
\section{\benchname Benchmark and Evaluation}
\label{sec:benchmark}
\subsection{Setup}
\label{sec:bench:setup}
\benchname evaluates whether graph representation methods remain
reliable under graph-specific deployment stresses. It standardizes
inputs, splits, stress operators, and metrics while preserving each
method's native adaptation pipeline, and reports behavior by safety
axis rather than as a single aggregate leaderboard.

These choices address the three design requirements outlined
above. First, safety axes are evaluated separately because they
stress different parts of the GRL pipeline and can induce different
subcondition-level rankings. Second, evaluation conditions are
standardized while method-native adaptation is preserved, so methods
are compared under shared splits, stress operators, and metrics
without forcing a common architecture or adaptation head. Finally,
datasets span multiple domains and task formats so that safety
behavior is not tied to a single graph family.

\begin{figure*}[t]
  \centering
  \includegraphics[width=\textwidth]{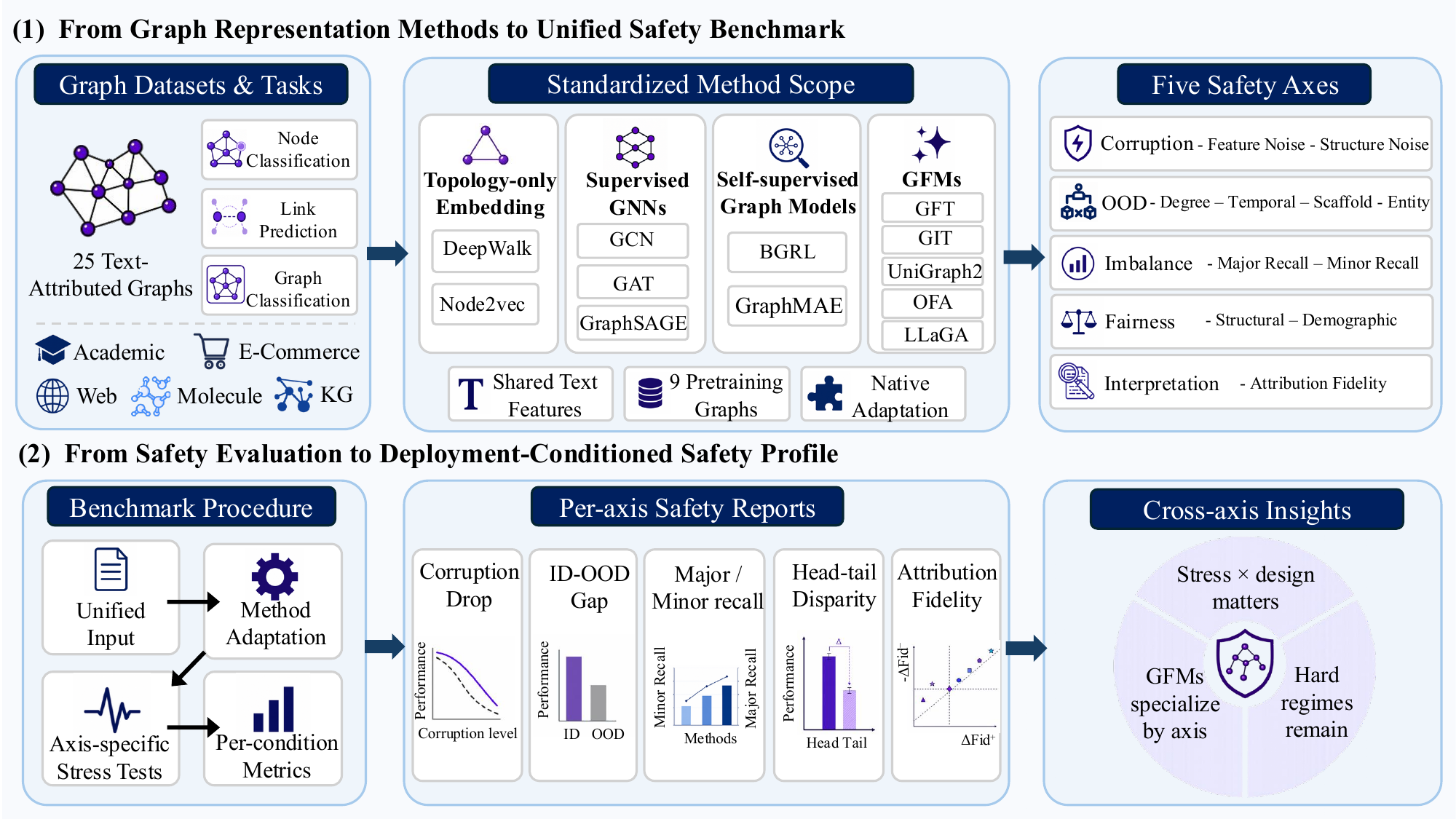}
  \caption{Overview of \benchname. We evaluate twelve GRL methods across five safety axes under standardized
  evaluation conditions.}
  \label{fig:overview}
\end{figure*}

\textbf{Method Scope.}
\label{sec:bench:models}
\benchname covers twelve methods spanning four tiers. Topology-only
baselines include DeepWalk~\citep{perozzi2014deepwalk} and
Node2vec~\citep{grover2016node2vec}. Supervised message-passing
baselines include GCN~\citep{kipf2017gcn}, GAT~\citep{velickovic2018gat},
and GraphSAGE~\citep{hamilton2017sage}. Self-supervised graph
representation methods include BGRL~\citep{thakoor2022bgrl} and
GraphMAE~\citep{hou2022graphmae}. Foundation-era graph models include
GFT~\citep{wang2024gft}, GIT~\citep{wang2024git},
UniGraph2~\citep{he2025unigraph2}, OFA~\citep{liu2024ofa}, and
LLaGA~\citep{chen2024llaga}, covering structured task abstraction,
cross-domain alignment, prompt-graph reformulation, and graph-language
adaptation.


\textbf{Datasets and Protocol.}
\label{sec:bench:datasets}
\benchname evaluates twenty-five text-attributed graphs covering node
classification, link prediction, and graph classification across
academic, e-commerce, web, knowledge-graph, recommendation, and
molecular domains; dataset statistics are reported in
Appendix~\ref{app:datasets}. For feature-consuming methods, each
node, edge, atom, or relation text field is embedded with
Sentence-BERT~\citep{reimers2019sbert} into a shared 768-dimensional
feature space. Topology-only methods use the same graph structure on
compatible task formats. Pretrained methods use a common nine-graph
pretraining corpus following GFT~\citep{wang2024gft}, while
supervised and topology-only baselines are fitted directly on each
downstream graph. All methods are evaluated under shared downstream
splits, stress operators, and metric definitions whenever the task
interface permits. Consequently, each safety axis is evaluated on the
subset of datasets whose task format, label structure, split metadata,
or attribution interface supports the corresponding stress operator;
Appendix~\ref{app:safety} summarizes this axis-specific dataset
applicability.

\textbf{Safety Axes.}
\label{sec:bench:dim}
\benchname evaluates five representative safety axes:
\colorhl{corcolor}{corruption robustness},
\colorhl{oodcolor}{OOD generalization},
\colorhl{imbcolor}{class imbalance},
\colorhl{faircolor}{fairness}, and
\colorhl{intcolor}{interpretation}.
Corruption separates feature perturbation from edge
deletion; OOD separates degree, temporal, scaffold, and inductive
entity shifts; imbalance reports major- and minor-class recall;
fairness reports structural and demographic group disparities; and
interpretation measures whether attributed substructures support the
prediction better than random evidence. Because these subconditions
can produce different method rankings, \benchname reports per-axis
results rather than collapsing them into a single safety score.
Capability transfer, calibration, privacy, and adaptive adversarial
robustness are outside the current scope.

\vspace{-1.0em}

\subsection{Per-Axis Safety Benchmark and Evaluation}
\label{sec:results}

The following subsections instantiate these five axes as concrete
evaluation protocols, report results across compatible methods, and
summarize the main empirical finding.

\vspace{-1.0em}

\subsubsection{\colorhl{corcolor}{Corruption Robustness}: Feature and Structure Noise}
\label{sec:results:corruption}

Corruption evaluates input-side stability under two noise channels:
feature noise, which degrades node attributes, and structure noise,
which removes observed edges. Feature noise applies to
feature-consuming methods, while structure noise applies to methods
relying on topology. This axis asks whether a graph representation
remains useful when attribute signal or graph structure is corrupted.

\noindent\textbf{Benchmark.}
We instantiate two corruption channels following common graph
robustness stress tests~\citep{liu2022airgnn,rong2020dropedge}.
\textbf{Feature noise} adds Gaussian perturbations to node features,
with severity controlled by relative scale
$\sigma_{\rm rel} \in \{0.1, 0.25, 0.5, 1.0, 2.0\}$; it is
inapplicable to topology-only embeddings. \textbf{Structure noise}
randomly removes edges, with severity controlled by deletion
probability $p \in \{0.05, 0.10, 0.20, 0.30, 0.50\}$.
Table~\ref{tab:robustness} reports the strongest setting for each
channel, $\sigma_{\rm rel}{=}2.0$ for feature noise and $p{=}0.5$ for
structure noise, on the shared node-classification datasets. We report
perturbed accuracy and Drop $=$ Clean $-$ Perturbed; lower drop
indicates stronger corruption stability. Full severity sweeps are
provided in Appendix~\ref{app:corruption_full}.

\noindent\textbf{Evidence.}
Table~\ref{tab:robustness} shows that feature noise and structure
noise induce different robustness profiles across method families.
Feature noise is inapplicable to topology-only embeddings; under edge
deletion, however, DeepWalk and Node2vec show large average drops of
9.63 and 9.84 points. In contrast, many supervised and self-supervised
GNNs remain comparatively stable across both channels. Foundation-era
methods do not form a uniformly robust tier: OFA drops under both
channels, while LLaGA is more stable under feature noise but drops
under structure noise, and UniGraph2 shows the reverse pattern. Full
sweeps in
Appendix~\ref{app:corruption_full} confirm that corruption sensitivity
depends on which graph signal is damaged.

\begin{findingbox}
\textbf{\textit{Finding 1.}} Corruption robustness is deployment-specific,
not a single aggregate property. Noisy attributes and missing links
induce different failure profiles, so stability under one signal
failure need not transfer to the other. Model selection should
therefore depend on the expected graph signal failure rather than clean
accuracy or a single corruption score.
\end{findingbox}

\begin{table}[H]
\centering
\caption{\textbf{Sev5 \colorhl{corcolor}{corruption} summary.}
Accuracy (\%) and clean-to-corrupted drop on four shared NC datasets.
Feature noise uses Gaussian perturbation with $\sigma{=}2.0$; structure
noise uses random edge deletion with $p{=}0.5$. Sev5 is the strongest
severity; $\text{Drop}=\text{Clean}-\text{Sev5}$ ($\downarrow$ is more
robust). Cells show mean$_{\pm\text{std}}$ over 5 seeds; aggregate std
is across datasets. \textbf{Bold}/\underline{underline}: best/second
per column. ``--'' indicates feature noise is inapplicable.}
\label{tab:robustness}
\scriptsize
\setlength{\tabcolsep}{2.6pt}
\renewcommand{\arraystretch}{1.1}
\resizebox{\textwidth}{!}{%
\begin{tabular}{l|cccc|cc|cccc|cc}
\toprule
 & \multicolumn{6}{c|}{\textbf{Feature Noise ($\sigma{=}2.0$)}} & \multicolumn{6}{c}{\textbf{Structure Noise ($p{=}0.5$)}} \\
\cmidrule(lr){2-7} \cmidrule(lr){8-13}
\textbf{Method} & \textbf{Citeseer} & \textbf{DBLP} & \textbf{Pubmed} & \textbf{Wikics} & \textbf{Sev5} & \textbf{Drop}\,$\downarrow$ & \textbf{Citeseer} & \textbf{DBLP} & \textbf{Pubmed} & \textbf{Wikics} & \textbf{Sev5} & \textbf{Drop}\,$\downarrow$ \\
\midrule
DeepWalk~\citep{perozzi2014deepwalk} & -- & -- & -- & -- & -- & -- & 31.7$_{\pm1.7}$ & 50.3$_{\pm0.5}$ & 52.0$_{\pm1.6}$ & 66.8$_{\pm1.0}$ & 50.2$_{\pm14.4}$ & 9.63$_{\pm7.3}$ \\
Node2vec~\citep{grover2016node2vec} & -- & -- & -- & -- & -- & -- & 31.0$_{\pm1.5}$ & 48.3$_{\pm1.3}$ & 52.6$_{\pm1.2}$ & 66.7$_{\pm0.8}$ & 49.6$_{\pm14.7}$ & 9.84$_{\pm6.5}$ \\
\midrule
GCN~\citep{kipf2017gcn} & 67.1$_{\pm1.6}$ & \underline{73.3}$_{\pm0.2}$ & \textbf{77.4}$_{\pm2.1}$ & \underline{77.9}$_{\pm0.7}$ & \underline{73.9}$_{\pm5.0}$ & \textbf{1.18}$_{\pm0.2}$ & 66.8$_{\pm0.8}$ & \underline{73.5}$_{\pm0.3}$ & \textbf{77.2}$_{\pm1.8}$ & 76.9$_{\pm0.7}$ & \underline{73.6}$_{\pm4.8}$ & 1.44$_{\pm0.6}$ \\
GAT~\citep{velickovic2018gat} & \underline{67.3}$_{\pm1.4}$ & \textbf{73.7}$_{\pm0.4}$ & \underline{77.2}$_{\pm0.9}$ & 77.6$_{\pm0.6}$ & \textbf{74.0}$_{\pm4.8}$ & \underline{1.29}$_{\pm0.4}$ & \textbf{67.7}$_{\pm1.4}$ & \textbf{74.0}$_{\pm0.4}$ & \underline{76.6}$_{\pm1.0}$ & 76.6$_{\pm0.5}$ & \textbf{73.7}$_{\pm4.2}$ & 1.51$_{\pm0.7}$ \\
GraphSAGE~\citep{hamilton2017sage} & \textbf{67.3}$_{\pm2.0}$ & 70.4$_{\pm1.8}$ & 74.2$_{\pm1.4}$ & 75.5$_{\pm1.2}$ & 71.9$_{\pm3.7}$ & 2.41$_{\pm0.7}$ & \underline{67.0}$_{\pm0.5}$ & 72.6$_{\pm0.8}$ & 74.8$_{\pm2.0}$ & 76.0$_{\pm0.3}$ & 72.6$_{\pm4.0}$ & 1.84$_{\pm0.6}$ \\
\midrule
BGRL~\citep{thakoor2022bgrl} & 63.9$_{\pm0.7}$ & 69.6$_{\pm0.5}$ & 72.0$_{\pm2.3}$ & \textbf{78.1}$_{\pm0.7}$ & 70.9$_{\pm5.9}$ & 1.98$_{\pm1.0}$ & 64.8$_{\pm0.7}$ & 70.6$_{\pm0.3}$ & 73.4$_{\pm2.7}$ & \textbf{78.0}$_{\pm0.6}$ & 71.7$_{\pm5.5}$ & \textbf{1.16}$_{\pm1.0}$ \\
GraphMAE~\citep{hou2022graphmae} & 63.8$_{\pm0.8}$ & 68.7$_{\pm1.3}$ & 76.2$_{\pm0.7}$ & 77.7$_{\pm0.9}$ & 71.6$_{\pm6.5}$ & 2.99$_{\pm1.3}$ & 65.5$_{\pm0.9}$ & 71.5$_{\pm0.7}$ & 76.4$_{\pm0.8}$ & \underline{77.8}$_{\pm0.5}$ & 72.8$_{\pm5.6}$ & 1.83$_{\pm0.7}$ \\
\midrule
GFT~\citep{wang2024gft} & 59.7$_{\pm0.9}$ & 58.5$_{\pm1.6}$ & 73.0$_{\pm1.2}$ & 67.2$_{\pm1.3}$ & 64.6$_{\pm6.8}$ & 7.36$_{\pm3.3}$ & 57.7$_{\pm0.5}$ & 63.9$_{\pm1.2}$ & 74.4$_{\pm1.5}$ & 75.6$_{\pm0.8}$ & 67.9$_{\pm8.6}$ & 3.85$_{\pm3.7}$ \\
GIT~\citep{wang2024git} & 49.9$_{\pm1.5}$ & 71.5$_{\pm0.1}$ & 73.5$_{\pm1.6}$ & 63.0$_{\pm2.0}$ & 64.5$_{\pm10.7}$ & 9.47$_{\pm8.2}$ & 61.5$_{\pm1.0}$ & 72.8$_{\pm0.2}$ & 74.4$_{\pm1.4}$ & 75.8$_{\pm0.7}$ & 71.1$_{\pm6.5}$ & 3.00$_{\pm2.3}$ \\
UniGraph2~\citep{he2025unigraph2} & 60.4$_{\pm0.7}$ & 59.9$_{\pm3.4}$ & 71.4$_{\pm2.0}$ & 70.4$_{\pm2.1}$ & 65.5$_{\pm6.2}$ & 5.23$_{\pm1.8}$ & 64.2$_{\pm1.0}$ & 66.8$_{\pm2.3}$ & 72.6$_{\pm1.7}$ & 72.2$_{\pm2.0}$ & 68.9$_{\pm4.1}$ & \underline{1.42}$_{\pm1.7}$ \\
OFA~\citep{liu2024ofa} & 50.8$_{\pm1.6}$ & 54.2$_{\pm0.7}$ & 62.3$_{\pm3.0}$ & 64.2$_{\pm1.2}$ & 57.9$_{\pm6.4}$ & 11.98$_{\pm0.9}$ & 53.5$_{\pm2.3}$ & 64.2$_{\pm1.8}$ & 69.1$_{\pm3.4}$ & 69.1$_{\pm0.8}$ & 64.0$_{\pm7.3}$ & 5.89$_{\pm3.9}$ \\
LLaGA~\citep{chen2024llaga} & 61.5$_{\pm0.7}$ & 68.2$_{\pm1.9}$ & 71.9$_{\pm2.8}$ & 73.2$_{\pm0.9}$ & 68.7$_{\pm5.2}$ & 2.00$_{\pm1.0}$ & 53.5$_{\pm2.2}$ & 68.3$_{\pm1.3}$ & 67.1$_{\pm2.6}$ & 72.2$_{\pm1.1}$ & 65.3$_{\pm8.1}$ & 5.44$_{\pm4.6}$ \\
\bottomrule
\end{tabular}%
}
\end{table}

\vspace{-1.2em}

\begin{figure}[H]
\centering
\includegraphics[width=\linewidth]{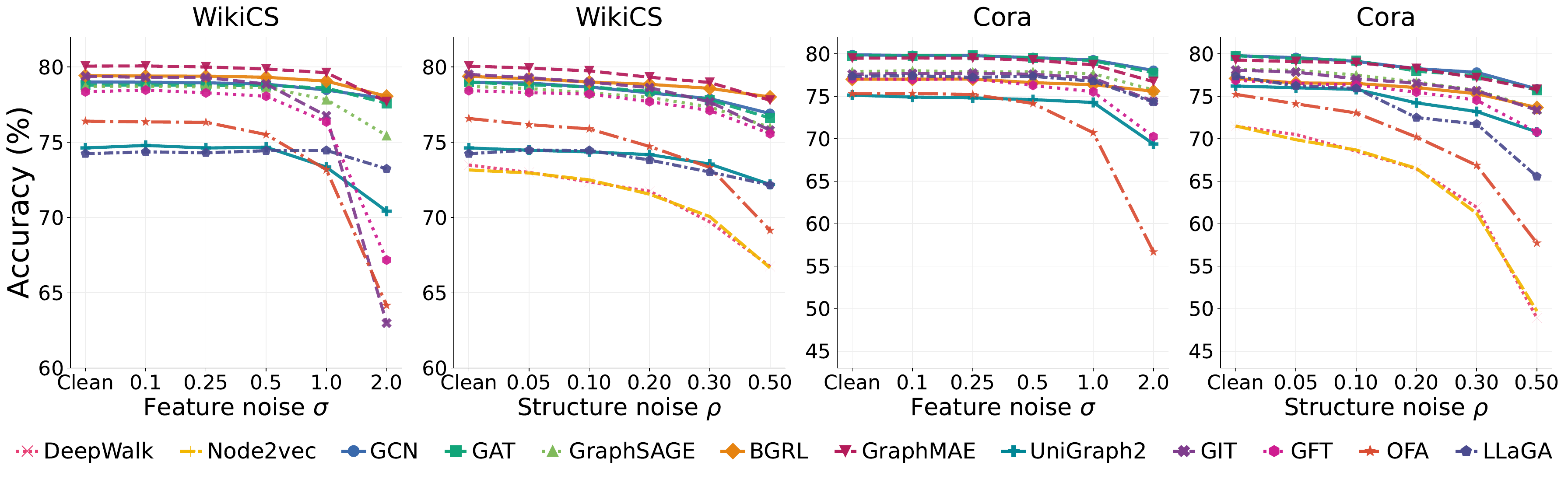}
\caption{\textbf{\colorhl{corcolor}{Corruption} severity trajectories.}
Test accuracy under feature- and structure-noise sweeps on WikiCS
and Cora. Curves denote methods; per-dataset curves in
Appendix Figures~\ref{fig:app_fn}--\ref{fig:app_ed}.}
\label{fig:corruption_curves}
\end{figure}

\subsubsection{\colorhl{oodcolor}{OOD Generalization}: Shifts Across Graph Contexts}
\label{sec:results:ood}

OOD generalization evaluates whether graph representations remain
useful when the graph context changes across structure, time,
molecular scaffold, or entity set. We study four OOD shift
mechanisms: degree shift, temporal shift, scaffold shift, and
inductive-entity shift.

\noindent\textbf{Benchmark.}
We instantiate four OOD shift mechanisms following mechanism-level
evaluation protocols in graph OOD benchmarks~\citep{gui2022good}.
\textbf{Degree shift} partitions each node-classification dataset by
node-degree quantile, using the high-degree majority as the ID split
and the low-degree quantile as the OOD split. \textbf{Temporal shift}
trains on arXiv papers with publication year $\leq 2010$ and evaluates
on papers from $\geq 2017$. \textbf{Scaffold shift} compares random
and Bemis--Murcko scaffold splits on BACE and
Tox21~\citep{hu2020ogb}. \textbf{Inductive-entity shift} is
operationalized as KG completion on WN18RR and FB15K237, where
test-time candidate entities are held out from downstream KG training.
We report ID-to-OOD accuracy drop for degree and temporal shifts,
$\Delta\text{AUC} = \text{AUC}_{\rm rand}-\text{AUC}_{\rm scaffold}$
for scaffold shift, and MRR / Hits@10 for inductive-entity shift.

\begin{figure}[t]
  \centering
  \begin{subfigure}[t]{0.49\linewidth}
    \vspace{0pt}
    \centering
    \includegraphics[width=\linewidth]{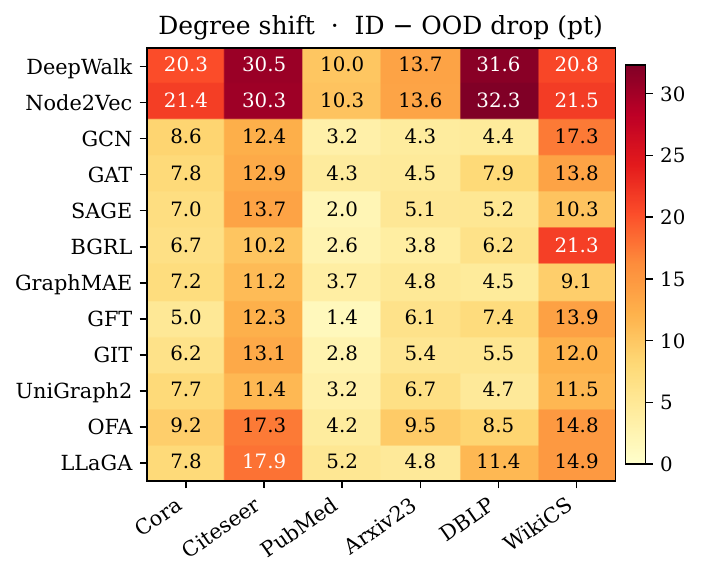}
    \caption{\colorhl{oodcolor}{\textbf{Degree shift.}} ID--OOD
    accuracy drop across 12 methods on 6 NC datasets. Full results are
    in Table~\ref{tab:app_ood_degree_combined}.}
    \label{fig:ood_degree_heatmap}
  \end{subfigure}\hfill
  \begin{subfigure}[t]{0.49\linewidth}
    \vspace{0pt}
    \centering
    \includegraphics[width=\linewidth]{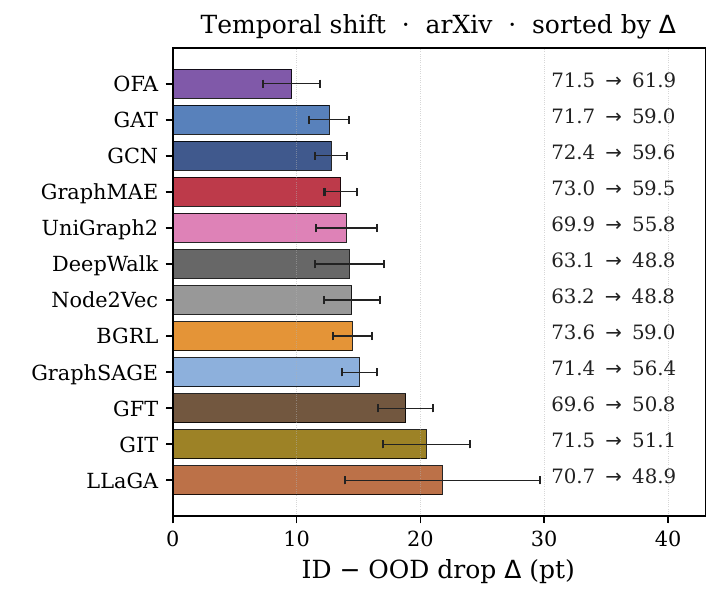}
    \caption{\colorhl{oodcolor}{\textbf{Temporal shift.}} ID--OOD
    accuracy drop on arXiv, sorted by drop size.}
    \label{fig:ood_temporal_bars}
  \end{subfigure}
  \caption{\textbf{Node-classification OOD shifts.} Larger values
  indicate weaker OOD generalization; darker cells and longer bars
  denote larger drops.}
  \label{fig:ood_nc_panels}
\end{figure}

\noindent\textbf{Evidence.}
Figure~\ref{fig:ood_nc_panels} and Table~\ref{tab:ood_beyond_nc}
show that OOD rankings change with the evaluated shift mechanism.
Under degree shift across six node-classification datasets,
topology-only embeddings have the largest average drops, while
GraphMAE has the smallest average drop (6.75 points across the six datasets in Figure~\ref{fig:ood_degree_heatmap}). On the arXiv
temporal split, OFA shows the smallest drop (9.61 points). Across
the two scaffold-shift datasets evaluated, the strongest method
differs: GraphSAGE shows the smallest BACE gap (9.7 $\Delta$AUC),
while UniGraph2 shows the smallest Tox21 gap (5.4 $\Delta$AUC).
Across the two inductive-entity KG datasets evaluated, the leader
also changes: BGRL leads WN18RR in MRR (0.231) but falls to 0.088
on FB15K237, whereas OFA leads FB15K237 (0.186) but drops to 0.075
on WN18RR. These results illustrate subcondition sensitivity rather
than a single OOD ranking.

\begin{findingbox}[oodcolor]
\textbf{\textit{Finding 2.}} OOD reliability is application-shift
specific. In the evaluated settings, degree shifts in
node-classification graphs, the arXiv temporal split, scaffold splits
in molecules, and inductive-entity shifts in KGs expose different
representation behaviors. Model selection should therefore be
conditioned on the anticipated deployment shift rather than on a
single aggregate OOD score.
\end{findingbox}


\vspace{-0.8em}

\begin{table}[t]
\caption{\textbf{OOD shift mechanisms beyond node classification.}
(a) \colorhl{oodcolor}{\textbf{Scaffold shift}} on BACE / Tox21:
ROC-AUC under random versus Bemis--Murcko scaffold splits
(80/10/10, OGB convention~\citep{hu2020ogb});
$\Delta=\mathrm{AUC}_{\mathrm{rand}}-\mathrm{AUC}_{\mathrm{scaf}}$,
where larger $\Delta$ indicates a larger scaffold-generalization gap.
(b) \colorhl{oodcolor}{\textbf{Inductive-entity shift}} on WN18RR /
FB15K237, operationalized as KG completion with test-time candidate
entities held out from downstream KG training; MRR / Hits@10, higher
is better. \textbf{Bold}/\underline{underline}: best/second per
column. Per-cell values are mean$_{\pm\text{std}}$ over 5 seeds.}
\label{tab:ood_beyond_nc}
\centering
\setlength{\tabcolsep}{2.5pt}
\renewcommand{\arraystretch}{1.05}
\resizebox{\textwidth}{!}{%
\begin{tabular}{@{}c@{\hspace{14pt}}c@{}}
\multicolumn{1}{c}{\Large\itshape (a) Mol scaffold OOD (BACE / Tox21)} &
\multicolumn{1}{c}{\Large\itshape (b) Inductive KG (WN18RR / FB15K237)} \\[2pt]
\begin{tabular}{l|ccc|ccc}
\toprule
 & \multicolumn{3}{c|}{\textbf{BACE}} & \multicolumn{3}{c}{\textbf{Tox21}} \\
\textbf{Method} & rand. & scaf. & $\Delta\,\downarrow$ & rand. & scaf. & $\Delta\,\downarrow$ \\
\midrule
GCN~\citep{kipf2017gcn}            & 64.5$_{\pm0.4}$ & 52.1$_{\pm0.6}$ & 12.3 & 78.5$_{\pm0.2}$ & 72.5$_{\pm0.2}$ & \underline{5.9} \\
GAT~\citep{velickovic2018gat}      & 62.9$_{\pm0.3}$ & 48.1$_{\pm1.1}$ & 14.7 & 78.6$_{\pm0.4}$ & 71.2$_{\pm0.3}$ & 7.4 \\
GraphSAGE~\citep{hamilton2017sage} & 66.7$_{\pm0.9}$ & 56.9$_{\pm0.9}$ & \textbf{9.7} & 80.9$_{\pm0.2}$ & 74.9$_{\pm0.2}$ & 6.0 \\
\midrule
BGRL~\citep{thakoor2022bgrl}       & \textbf{85.3}$_{\pm0.8}$ & 61.7$_{\pm2.1}$ & 23.6 & \underline{82.4}$_{\pm0.3}$ & 76.0$_{\pm0.3}$ & 6.4 \\
GraphMAE~\citep{hou2022graphmae}   & 76.5$_{\pm0.8}$ & 54.3$_{\pm0.6}$ & 22.2 & 80.4$_{\pm0.5}$ & 73.1$_{\pm0.7}$ & 7.4 \\
\midrule
GFT~\citep{wang2024gft}            & 70.0$_{\pm5.8}$ & 54.3$_{\pm1.4}$ & 15.7 & 80.8$_{\pm0.6}$ & 73.4$_{\pm0.8}$ & 7.5 \\
GIT~\citep{wang2024git}            & 82.8$_{\pm0.6}$ & \underline{65.2}$_{\pm0.6}$ & 17.5 & \textbf{83.0}$_{\pm0.4}$ & \textbf{76.0}$_{\pm0.4}$ & 6.9 \\
UniGraph2~\citep{he2025unigraph2}  & 68.5$_{\pm2.1}$ & 56.7$_{\pm2.8}$ & \underline{11.7} & 75.5$_{\pm1.6}$ & 70.2$_{\pm0.9}$ & \textbf{5.4} \\
OFA~\citep{liu2024ofa}             & \underline{84.0}$_{\pm1.3}$ & \textbf{66.6}$_{\pm1.6}$ & 17.3 & 81.8$_{\pm0.5}$ & \underline{75.5}$_{\pm0.8}$ & 6.3 \\
LLaGA~\citep{chen2024llaga}        & 79.0$_{\pm0.7}$ & 56.0$_{\pm0.9}$ & 23.0 & 79.8$_{\pm2.2}$ & 73.6$_{\pm0.9}$ & 6.2 \\
\bottomrule
\end{tabular}
&
\begin{tabular}{l|cc|cc}
\toprule
 & \multicolumn{2}{c|}{\textbf{WN18RR}} & \multicolumn{2}{c}{\textbf{FB15K237}} \\
\textbf{Method} & MRR\,$\uparrow$ & Hits@10\,$\uparrow$ & MRR\,$\uparrow$ & Hits@10\,$\uparrow$ \\
\midrule
GCN~\citep{kipf2017gcn}            & 0.192$_{\pm.014}$ & 0.346$_{\pm.015}$ & 0.098$_{\pm.006}$ & 0.179$_{\pm.005}$ \\
GAT~\citep{velickovic2018gat}      & 0.167$_{\pm.011}$ & 0.363$_{\pm.021}$ & 0.039$_{\pm.004}$ & 0.092$_{\pm.017}$ \\
GraphSAGE~\citep{hamilton2017sage} & 0.134$_{\pm.027}$ & 0.262$_{\pm.031}$ & 0.172$_{\pm.011}$ & \underline{0.378}$_{\pm.015}$ \\
\midrule
BGRL~\citep{thakoor2022bgrl}       & \textbf{0.231}$_{\pm.009}$ & \underline{0.443}$_{\pm.027}$ & 0.088$_{\pm.004}$ & 0.153$_{\pm.002}$ \\
GraphMAE~\citep{hou2022graphmae}   & \underline{0.224}$_{\pm.014}$ & \textbf{0.541}$_{\pm.025}$ & 0.121$_{\pm.009}$ & 0.256$_{\pm.015}$ \\
\midrule
GFT~\citep{wang2024gft}            & 0.157$_{\pm.005}$ & 0.339$_{\pm.013}$ & \underline{0.174}$_{\pm.005}$ & 0.364$_{\pm.009}$ \\
GIT~\citep{wang2024git}            & 0.094$_{\pm.013}$ & 0.212$_{\pm.010}$ & 0.161$_{\pm.005}$ & 0.346$_{\pm.010}$ \\
UniGraph2~\citep{he2025unigraph2}  & 0.148$_{\pm.007}$ & 0.315$_{\pm.015}$ & 0.064$_{\pm.008}$ & 0.127$_{\pm.012}$ \\
OFA~\citep{liu2024ofa}             & 0.075$_{\pm.011}$ & 0.139$_{\pm.013}$ & \textbf{0.186}$_{\pm.013}$ & \textbf{0.385}$_{\pm.019}$ \\
LLaGA~\citep{chen2024llaga}        & 0.152$_{\pm.019}$ & 0.270$_{\pm.027}$ & 0.082$_{\pm.012}$ & 0.155$_{\pm.019}$ \\
\bottomrule
\end{tabular}
\end{tabular}%
}
\end{table}

\subsubsection{\colorhl{imbcolor}{Class Imbalance}: Major and Minor Recall}
\label{sec:results:imbalance}

Class imbalance evaluates whether strong average performance reflects
balanced class-wise recall or is driven mainly by majority-class
predictions under skewed label support.

\begin{figure}[H]
\centering
\includegraphics[width=\linewidth]{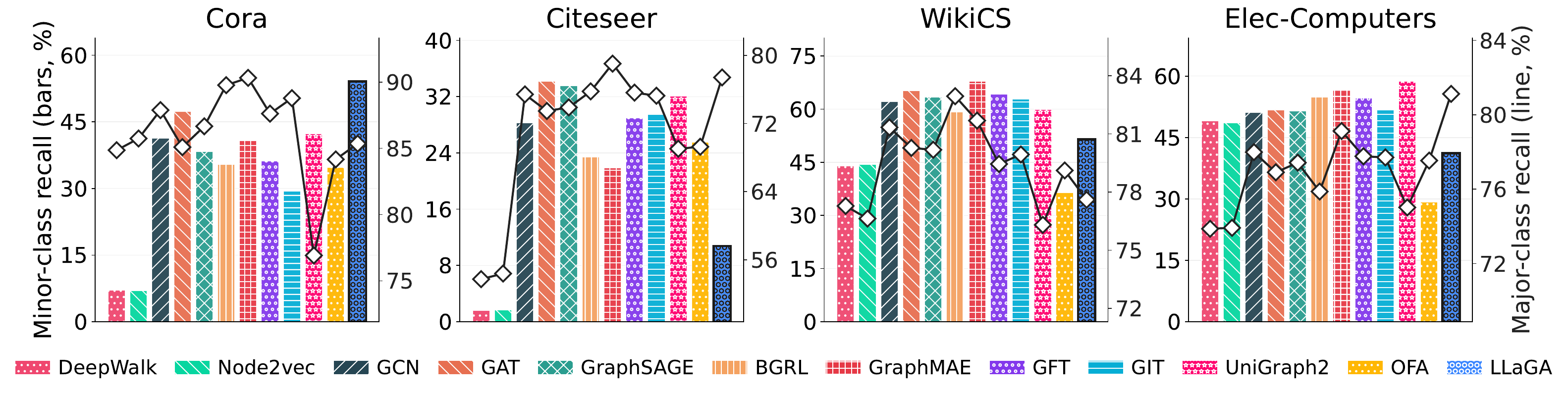}
\vspace{-1em}
\caption{\textbf{Major- and minor-class recall under \colorhl{imbcolor}{class
imbalance}.} At imbalance ratio $\rho{=}10$, bars show minor-class
recall and lines show major-class recall on four representative
node-classification datasets. High major-class recall does not
necessarily imply high minor-class recall.}
\label{fig:imbalance_v2_combo}
\vspace{-1em}
\end{figure}

\noindent\textbf{Benchmark.}
We measure class-imbalance behavior on node classification at
imbalance ratio $\rho{=}10$. Classes are sorted by their original
training-set sample count; the lower half are treated as minor classes
and downsampled to $n_{\rm major}/\rho$, where $n_{\rm major}$ is the
largest training count among the upper-half major classes. Full
protocol details are in Appendix~\ref{app:safety:imbalance}. Each
method is adapted on the imbalanced training graph using its native
pipeline. We report per-class recall on the
test split, averaged separately across major and minor classes.
Figure~\ref{fig:imbalance_v2_combo} shows four representative datasets
where all twelve methods are covered; full per-dataset results for
eight categorical-label NC datasets and $\rho\in\{5,10,20\}$ are
reported in Appendix Table~\ref{tab:app_imb_per_class}.

\noindent\textbf{Evidence.}
Figure~\ref{fig:imbalance_v2_combo} shows that major-class and
minor-class recall separate under skewed label support. This separation
is most visible for topology-only graph embeddings, which have
near-zero minor-class recall on Cora and CiteSeer despite much higher
major-class recall. Among feature-consuming methods, the strongest
minor-class method changes by dataset: LLaGA leads on Cora, GAT on
CiteSeer, GraphMAE on WikiCS, and UniGraph2 on Elec-Computers.
Minor-class gains also need not align with majority-class recall: on
Elec-Computers, UniGraph2 has the best minor recall but lower major
recall than LLaGA ($75.0\%$ vs.\ $81.1\%$), whereas OFA keeps
competitive major recall while dropping to $26.2\%$ minor recall on
WikiCS. Overall, class imbalance exposes a major-minor coverage
trade-off rather than a uniform accuracy degradation pattern.

\begin{findingbox}[imbcolor]
\textbf{\textit{Finding 3.}} Class imbalance creates a rare-class
coverage failure in graph prediction. Under skewed label support,
strong majority-class recall can hide severe minority-class collapse,
while methods with better minority coverage may sacrifice majority
performance. For graph tasks where rare classes carry the primary
risk, model selection should prioritize minority coverage and balanced
recall rather than average accuracy.
\end{findingbox}


\vspace{-1.0em}

\subsubsection{\colorhl{faircolor}{Fairness}: Structural Head--Tail Disparity}
\label{sec:results:fairness}

Fairness evaluates whether prediction quality differs across
groups. In GRL, a natural group axis is
structural position: high-degree nodes receive more neighborhood
evidence, while low-degree nodes have thinner local context.

\begin{figure}[!htbp]
\centering
\includegraphics[width=\linewidth]{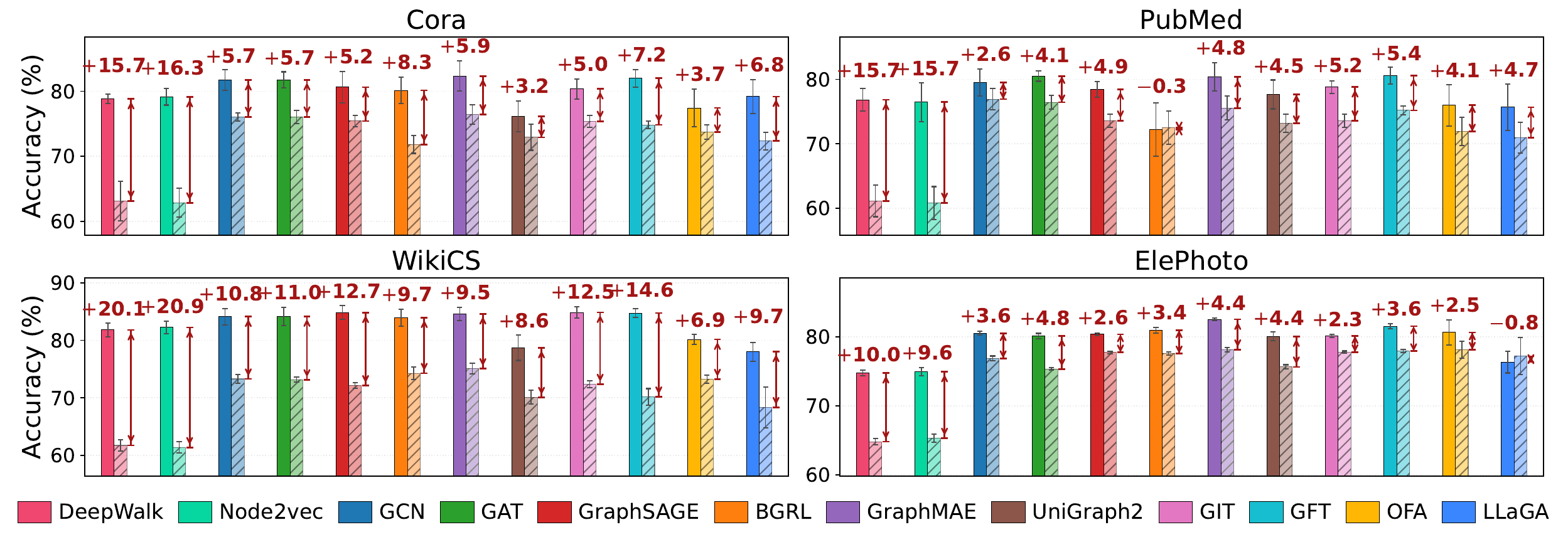}
\caption{\textbf{\colorhl{faircolor}{Structural head--tail
fairness.}} Accuracy across all 12 methods on four
node-classification datasets. Each method has two bars: head nodes
(solid, top-degree quintile) and tail nodes (hatched, bottom-degree
quintile). Red labels report the signed gap
$\Delta = \mathrm{Acc}_{\rm head} - \mathrm{Acc}_{\rm tail}$ in
percentage points; positive values indicate head-favoring behavior
and negative values indicate tail-favoring behavior. Error bars
show standard deviation across 5 seeds.}
\label{fig:fair_structural_v2}
\end{figure}

\noindent\textbf{Benchmark.}
We evaluate structural fairness on node classification. For each
dataset, test nodes are split into \emph{head} nodes (top 20\% by
degree) and \emph{tail} nodes (bottom 20\% by degree), and we
report the signed head-minus-tail accuracy gap
$\Delta = \mathrm{Acc}_{\rm head} - \mathrm{Acc}_{\rm tail}$.
A smaller $|\Delta|$ indicates more balanced head/tail performance;
$\Delta > 0$ indicates head-favoring behavior, while $\Delta < 0$
indicates tail-favoring behavior. Each method is adapted with its
native pipeline, and no fairness-aware training intervention is
applied. Figure~\ref{fig:fair_structural_v2} shows four
representative datasets where all twelve methods are covered. Full
structural-fairness results on eight NC datasets and
demographic-fairness results on Tolokers are reported in
Appendix~\ref{app:fair_full}.

\noindent\textbf{Evidence.}
Figure~\ref{fig:fair_structural_v2} shows that head--tail disparity
is graph-conditioned rather than tier-conditioned. Most methods favor
head nodes on Cora, PubMed, WikiCS, and ElePhoto, but the size of this
advantage varies substantially across graphs and methods. The
topology-only graph embeddings often show large head-favoring gaps,
indicating that purely structural representations can amplify
degree-position disparity. The smallest absolute gap also rotates
across datasets: UniGraph2 is smallest on Cora, BGRL on PubMed, OFA
on WikiCS, and LLaGA on ElePhoto. Individual methods do not carry a
fixed fairness profile: for example, GFT has a large gap on WikiCS
($+14.58$ pp) but a much smaller gap on ElePhoto ($+3.62$ pp).
Overall, structural fairness depends on the interaction between graph
degree structure and the representation interface.

\begin{findingbox}[faircolor]
\textbf{\textit{Finding 4.}} Structural fairness is a deployment-relevant
degree-position risk in graph prediction. In degree-skewed graphs,
high-degree and low-degree nodes can experience different error
profiles, so fairness assessment should report whether a representation
amplifies, reduces, or reverses head--tail disparity.
\end{findingbox}



\vspace{-0.5em}

\subsubsection{\colorhl{intcolor}{Interpretation}: Attribution Fidelity}
\label{sec:results:interpret}

Interpretation evaluates whether the input elements highlighted by a
post-hoc attribution method are actually predictive. This axis
does not ask whether a method is accurate, but whether its
attributed substructures support the prediction better than a random
baseline at the same sparsity.

\noindent\textbf{Benchmark.}
We evaluate node-level attribution fidelity using gradient-saliency
edge attribution under subgraph ablation~\citep{agarwal2023graphxai}.
For each test node, we mask the top-$k{=}10\%$ edges in its
$K{=}2$-hop receptive field and compute the GraphFramEx
characterization score~\citep{amara2022graphframex}, the harmonic
mean of $\text{Fid}^{+}$ (confidence drop when salient edges are
masked) and $1{-}\text{Fid}^{-}$ (prediction preservation when
non-salient edges are masked). We report
$\Delta_{\rm char}{\times}100 = \text{char}_{\rm sal} -
\text{char}_{\rm rand}$, the lift over a random-edge baseline; a
positive value means saliency identifies more predictive edges than
guessing. This evaluation covers eight methods on eight
NC datasets; OFA and LLaGA are not included because their
adaptation paths do not expose the same per-edge gradient interface
under this protocol.

\noindent\textbf{Evidence.}
Table~\ref{tab:interp_v2_app} shows that node-level attribution
fidelity is concentrated in a small set of methods. GFT has the
highest cross-dataset mean ($31.30$), followed by GraphSAGE
($27.23$) and GIT ($24.26$); the remaining methods are at or below
$14.00$ on average. Per dataset, GFT leads six of eight columns,
while GIT leads the two e-commerce graphs. Appendix
Figure~\ref{fig:interp_main} shows that the two fidelity components
move together across representative NC datasets. Overall, node-level
attribution fidelity is concentrated rather than universal: only a
small subset of methods consistently produces edge saliency that
outperforms random-edge ablation. The evaluation is also
interface-dependent, since methods without a shared per-edge gradient
interface require separate attribution protocols.


\begin{findingbox}[intcolor]
\textbf{\textit{Finding 5.}} Explanation availability is not the same
as attribution fidelity. For graph domains where explanations are used
as scientific or molecular evidence, safety evaluation should test
whether highlighted edges or subgraphs are actually predictive under
the stated attribution protocol, rather than assuming that a saliency
interface yields faithful explanations.
\end{findingbox}

\begin{table}[t]
\centering
\caption{\textbf{Node-level subgraph-ablation \colorhl{intcolor}{fidelity}.}
$\Delta_{\text{char}}\times100$ measures the lift of gradient-saliency
edge attribution over a random-edge baseline under top-$k{=}10\%$
subgraph ablation; higher is better and $0$ means no lift over random.
Per-dataset cells report mean$\pm$std over 5 seeds; the Avg column
reports the cross-dataset mean with between-dataset std.
\textbf{Bold}/\underline{underline}: best/second per column.}
\label{tab:interp_v2_app}
\footnotesize
\setlength{\tabcolsep}{2.5pt}
\renewcommand{\arraystretch}{1.0}
\resizebox{\linewidth}{!}{%
\begin{tabular}{lcccccccc|c}
\toprule
\textbf{Method} & \textbf{Cora} & \textbf{Citeseer} & \textbf{PubMed} & \textbf{WikiCS} & \textbf{Elec-Computers} & \textbf{Elec-Photo} & \textbf{Amazon-Ratings} & \textbf{Tolokers} & \textbf{Avg}$\uparrow$ \\
\midrule
GCN & 11.56$_{\pm2.18}$ & 5.43$_{\pm5.69}$ & 12.66$_{\pm2.66}$ & 23.72$_{\pm2.09}$ & 16.13$_{\pm5.23}$ & 16.53$_{\pm2.64}$ & 3.53$_{\pm2.56}$ & 22.42$_{\pm0.24}$ & 14.00$_{\pm7.24}$ \\
GAT & 11.25$_{\pm1.75}$ & 3.40$_{\pm1.80}$ & 18.44$_{\pm3.32}$ & 25.82$_{\pm4.55}$ & 8.12$_{\pm3.25}$ & 9.61$_{\pm3.56}$ & 2.01$_{\pm3.71}$ & 0.17$_{\pm0.03}$ & 9.85$_{\pm8.71}$ \\
GraphSAGE & \underline{30.21}$_{\pm0.72}$ & 11.97$_{\pm7.87}$ & \underline{33.74}$_{\pm4.00}$ & 51.84$_{\pm1.58}$ & \underline{25.69}$_{\pm5.52}$ & \underline{24.83}$_{\pm4.44}$ & \underline{16.00}$_{\pm5.48}$ & \underline{23.56}$_{\pm0.48}$ & \underline{27.23}$_{\pm12.17}$ \\
\midrule
BGRL & 14.24$_{\pm1.06}$ & 8.95$_{\pm0.64}$ & 11.02$_{\pm4.75}$ & 20.14$_{\pm2.26}$ & 18.22$_{\pm5.81}$ & 16.98$_{\pm2.05}$ & 5.89$_{\pm3.12}$ & 8.16$_{\pm1.73}$ & 12.95$_{\pm5.21}$ \\
GraphMAE & 12.81$_{\pm0.19}$ & 6.58$_{\pm3.60}$ & 17.34$_{\pm0.94}$ & 25.36$_{\pm1.82}$ & 9.13$_{\pm5.39}$ & 10.55$_{\pm2.42}$ & 6.71$_{\pm2.67}$ & 1.21$_{\pm0.81}$ & 11.21$_{\pm7.43}$ \\
\midrule
GFT & \textbf{30.64}$_{\pm3.00}$ & \textbf{19.52}$_{\pm21.72}$ & \textbf{40.11}$_{\pm5.03}$ & \textbf{63.65}$_{\pm1.52}$ & 24.87$_{\pm4.48}$ & 24.17$_{\pm1.20}$ & \textbf{17.25}$_{\pm0.93}$ & \textbf{30.22}$_{\pm1.40}$ & \textbf{31.30}$_{\pm14.90}$ \\
GIT & 21.26$_{\pm18.28}$ & \underline{13.59}$_{\pm3.75}$ & 30.12$_{\pm5.96}$ & \underline{53.50}$_{\pm2.93}$ & \textbf{26.54}$_{\pm4.84}$ & \textbf{25.00}$_{\pm2.42}$ & 14.22$_{\pm2.81}$ & 9.85$_{\pm0.98}$ & 24.26$_{\pm13.77}$ \\
UniGraph2 & 5.38$_{\pm1.65}$ & 2.75$_{\pm2.43}$ & 10.73$_{\pm7.27}$ & 24.58$_{\pm4.37}$ & 7.33$_{\pm4.42}$ & 6.40$_{\pm2.09}$ & 5.02$_{\pm1.50}$ & 14.19$_{\pm3.57}$ & 9.55$_{\pm7.05}$ \\
\bottomrule
\end{tabular}%
}
\end{table}

\subsection{Cross-Axis Discussion}
\label{sec:cross_dim}

The per-axis results in \S\ref{sec:results} show that safety failures
are not isolated to a single stressor. We synthesize them into three
cross-axis insights.

\textbf{Insight 1: Safety behavior is shaped by stressed graph factors and representation design.}
Across corruption, OOD generalization, and fairness, method rankings
and failure modes change with the interaction between the stressed graph
factor and the underlying representation design. Topology-dependent
methods are fragile under edge deletion but feature noise is
inapplicable to topology-only embeddings; BGRL and OFA take leadership
on different KG OOD datasets; and GFT's structural-fairness gap is much
larger on WikiCS than on ElePhoto. These observations suggest that
safety outcomes are governed by which graph factor a representation
uses and which factor fails at deployment, rather than by supervised,
self-supervised, or foundation-era status alone.

\textbf{Insight 2: Foundation-era methods show axis-specific strengths, not broad safety dominance.}
Foundation-era methods lead in some safety regimes but do not dominate
across the benchmark. GFT has the strongest node-level attribution
fidelity, OFA has the smallest drop on the arXiv temporal split, and
UniGraph2 is competitive for minority-class recall on multiple
imbalance datasets; however, each also has weaker axes or channels. Thus,
adopting a foundation-era graph model is not a general safety recipe:
the relevant question is which safety mechanism its interface supports
or weakens.

\textbf{Insight 3: Some deployment regimes remain difficult for the evaluated method set.}
Several settings expose limitations beyond choosing the best method in
the current cohort. Even the strongest method on the arXiv temporal
split still drops by about ten accuracy points, and class imbalance
produces a major--minor recall trade-off rather than a uniformly good
operating point. These regimes
point to method-capability gaps that require new robustness,
adaptation, or training objectives, not only better model selection.

\section{Limitations and Future Work}
\label{sec:limitations}
\benchname covers representative graph-safety axes rather than an
exhaustive taxonomy. Calibration, adaptive adversarial robustness, and
privacy are outside the current scope, and method-axis coverage remains
limited by adapter compatibility, such as the per-edge gradient
interface required for node-level attribution fidelity. Future versions
can add domains, adapters, and stress operators while preserving the
same condition-aware reporting format.

Beyond these scope and coverage limitations, \benchname also exposes
method-level open challenges. The current evaluation identifies safety
failures through fixed stress operators and evaluation metrics. A next
step is to develop methods that explicitly optimize for the expected
deployment stressor, such as missing links, temporal drift, rare classes,
or structurally disadvantaged nodes.

\section{Conclusion}
\benchname shows that GRL safety is
condition-dependent: clean accuracy or method family alone does not
reliably predict behavior under deployment-relevant stress. Across
corruption, OOD generalization, class imbalance, fairness, and
interpretation, method rankings vary with the interaction between the
stressed graph factor and the underlying representation design. These results suggest that graph
safety should be evaluated through per-axis and subcondition reports
rather than collapsed into a single aggregate score. The main purpose of
\benchname is therefore not to produce a universal safety leaderboard,
but to provide a diagnostic view of which methods fail under which graph
conditions. This view can guide deployment-time model selection by
matching methods to expected stressors, such as missing links, temporal
drift, rare classes, or structurally disadvantaged nodes, while also
exposing capability gaps that require future GRL
methods to optimize for safety-relevant graph conditions rather than
clean performance alone.


\clearpage
\bibliographystyle{plainnat}
\bibliography{references}

@article{liang2022helm,
  title={Holistic Evaluation of Language Models},
  author={Liang, Percy and Bommasani, Rishi and Lee, Tony and Tsipras, Dimitris and Soylu, Dilara and Yasunaga, Michihiro and Zhang, Yian and Narayanan, Deepak and Wu, Yuhuai and Kumar, Ananya and Newman, Benjamin and others},
  journal={Transactions on Machine Learning Research},
  year={2023},
  eprint={2211.09110},
  archivePrefix={arXiv}
}

@article{wu2020comprehensive,
  title={A comprehensive survey on graph neural networks},
  author={Wu, Zonghan and Pan, Shirui and Chen, Fengwen and Long, Guodong and Zhang, Chengqi and Yu, Philip S},
  journal={IEEE transactions on neural networks and learning systems},
  volume={32},
  number={1},
  pages={4--24},
  year={2020},
  publisher={IEEE}
}

@inproceedings{wang2023decodingtrust,
  title={{DecodingTrust}: A Comprehensive Assessment of Trustworthiness in {GPT} Models},
  author={Wang, Boxin and Chen, Weixin and Pei, Hengzhi and Xie, Chulin and Kang, Mintong and Zhang, Chenhui and Xu, Chejian and Xiong, Zidi and Dutta, Ritik and Schaeffer, Rylan and others},
  booktitle={Advances in Neural Information Processing Systems (NeurIPS) Datasets and Benchmarks Track},
  year={2023},
  eprint={2306.11698},
  archivePrefix={arXiv}
}

@inproceedings{liu2024ofa,
  title={One for All: Towards Training One Graph Model for All Classification Tasks},
  author={Liu, Hao and Feng, Jiarui and Kong, Lecheng and Liang, Ningyue and Tao, Dacheng and Chen, Yixin and Zhang, Muhan},
  booktitle={International Conference on Learning Representations (ICLR)},
  year={2024},
  eprint={2310.00149},
  archivePrefix={arXiv}
}

@inproceedings{wang2024gft,
  title={{GFT}: Graph Foundation Model with Transferable Tree Vocabulary},
  author={Wang, Zehong and Zhang, Zheyuan and Chawla, Nitesh V. and Zhang, Chuxu and Ye, Yanfang},
  booktitle={Advances in Neural Information Processing Systems (NeurIPS)},
  year={2024},
  eprint={2411.06070},
  archivePrefix={arXiv}
}

@inproceedings{wang2024git,
  title={Towards Graph Foundation Models: Learning Generalities Across Graphs via Task-Trees},
  author={Wang, Zehong and Zhang, Zheyuan and Ma, Tianyi and Chawla, Nitesh V. and Zhang, Chuxu and Ye, Yanfang},
  booktitle={International Conference on Machine Learning (ICML)},
  year={2025},
  eprint={2412.16441},
  archivePrefix={arXiv}
}

@inproceedings{he2025unigraph2,
  title={{UniGraph2}: Learning a Unified Embedding Space to Bind Multimodal Graphs},
  author={He, Yufei and Sui, Yuan and He, Xiaoxin and Liu, Yue and Sun, Yifei and Hooi, Bryan},
  booktitle={Proceedings of the ACM Web Conference (WWW)},
  year={2025},
  eprint={2502.00806},
  archivePrefix={arXiv}
}

@inproceedings{thakoor2022bgrl,
  title={Large-Scale Representation Learning on Graphs via Bootstrapping},
  author={Thakoor, Shantanu and Tallec, Corentin and Azar, Mohammad Gheshlaghi and Azabou, Mehdi and Dyer, Eva L. and Munos, R{\'e}mi and Veli{\v{c}}kovi{\'c}, Petar and Valko, Michal},
  booktitle={International Conference on Learning Representations (ICLR)},
  year={2022},
  eprint={2102.06514},
  archivePrefix={arXiv}
}

@inproceedings{hou2022graphmae,
  title={{GraphMAE}: Self-Supervised Masked Graph Autoencoders},
  author={Hou, Zhenyu and Liu, Xiao and Cen, Yukuo and Dong, Yuxiao and Yang, Hongxia and Wang, Chunjie and Tang, Jie},
  booktitle={Proceedings of the 28th ACM SIGKDD Conference on Knowledge Discovery and Data Mining (KDD)},
  year={2022},
  eprint={2205.10803},
  archivePrefix={arXiv}
}

@inproceedings{chen2024llaga,
  title={{LLaGA}: Large Language and Graph Assistant},
  author={Chen, Runjin and Zhao, Tong and Jaiswal, Ajay and Shah, Neil and Wang, Zhangyang},
  booktitle={Proceedings of the 41st International Conference on Machine Learning (ICML)},
  year={2024},
  eprint={2402.08170},
  archivePrefix={arXiv}
}

@article{gfmsurvey2025,
  title={Graph Foundation Models: A Comprehensive Survey},
  author={Wang, Zehong and Liu, Zheyuan and Ma, Tianyi and Li, Jiazheng and Zhang, Zheyuan and Fu, Xingbo and Li, Yiyang and Yuan, Zhengqing and Song, Wei and Ma, Yijun and Zeng, Qingkai and Chen, Xiusi and Zhao, Jianan and Li, Jundong and Jiang, Meng and Lio, Pietro and Chawla, Nitesh and Zhang, Chuxu and Ye, Yanfang},
  journal={arXiv preprint arXiv:2505.15116},
  year={2025}
}

@article{xu2024graphfm,
  title={{GraphFM}: A Comprehensive Benchmark for Graph Foundation Model},
  author={Xu, Yuhao and Liu, Xinqi and Duan, Keyu and Fang, Yi and Chuang, Yu-Neng and Zha, Daochen and Tan, Qiaoyu},
  journal={arXiv preprint arXiv:2406.08310},
  year={2024}
}

@article{yu2026evalgfm,
  title={Evaluating Progress in Graph Foundation Models: A Comprehensive Benchmark and New Insights},
  author={Yu, Xingtong and Ye, Shenghua and Liang, Ruijuan and Zhou, Chang and Cheng, Hong and Zhang, Xinming and Fang, Yuan},
  journal={arXiv preprint arXiv:2603.10033},
  year={2026}
}

@article{chen2024tsgfm,
  title={Text-space Graph Foundation Models: Comprehensive Benchmarks and New Insights},
  author={Chen, Zhikai and Mao, Haitao and Liu, Jingzhe and Song, Yu and Li, Bingheng and Jin, Wei and Fatemi, Bahare and Tsitsulin, Anton and Perozzi, Bryan and Liu, Hui and Tang, Jiliang},
  journal={arXiv preprint arXiv:2406.10727},
  year={2024}
}

@article{dai2024trustworthy,
  title={A Comprehensive Survey on Trustworthy Graph Neural Networks: Privacy, Robustness, Fairness, and Explainability},
  author={Dai, Enyan and Zhao, Tianxiang and Zhu, Huaisheng and Xu, Junjie and Guo, Zhimeng and Liu, Hui and Tang, Jiliang and Wang, Suhang},
  journal={Machine Intelligence Research},
  year={2024},
  volume={21},
  number={6},
  pages={1011--1061}
}

@inproceedings{gui2022good,
  title={{GOOD}: A Graph Out-of-Distribution Benchmark},
  author={Gui, Shurui and Li, Xiner and Wang, Limei and Ji, Shuiwang},
  booktitle={Advances in Neural Information Processing Systems (NeurIPS) Datasets and Benchmarks Track},
  year={2022},
  eprint={2206.08452},
  archivePrefix={arXiv}
}

@inproceedings{wang2024noisygl,
  title={{NoisyGL}: A Comprehensive Benchmark for Graph Neural Networks under Label Noise},
  author={Wang, Zhonghao and Sun, Danyu and Zhou, Sheng and Wang, Haobo and Fan, Jiapei and Huang, Longtao and Bu, Jiajun},
  booktitle={Advances in Neural Information Processing Systems (NeurIPS) Datasets and Benchmarks Track},
  year={2024},
  eprint={2406.04299},
  archivePrefix={arXiv}
}

@article{agarwal2023graphxai,
  title={Evaluating Explainability for Graph Neural Networks},
  author={Agarwal, Chirag and Queen, Owen and Lakkaraju, Himabindu and Zitnik, Marinka},
  journal={Scientific Data},
  year={2023},
  volume={10},
  number={1},
  pages={144},
  eprint={2208.09339},
  archivePrefix={arXiv}
}

@article{dong2024fairgraph,
  title={A Benchmark for Fairness-Aware Graph Learning},
  author={Dong, Yushun and Wang, Song and Lei, Zhenyu and Zheng, Zaiyi and Ma, Jing and Chen, Chen and Li, Jundong},
  journal={arXiv preprint arXiv:2407.12112},
  year={2024}
}

@article{zhang2025trustglm,
  title={{TrustGLM}: Evaluating the Robustness of Graph {LLM}s Against Prompt, Text, and Structure Attacks},
  author={Zhang, Qihai and Sheng, Xinyue and Sun, Yuanfu and Tan, Qiaoyu},
  journal={arXiv preprint arXiv:2506.11844},
  year={2025}
}

@inproceedings{kipf2017gcn,
  title={Semi-Supervised Classification with Graph Convolutional Networks},
  author={Kipf, Thomas N. and Welling, Max},
  booktitle={International Conference on Learning Representations (ICLR)},
  year={2017},
  eprint={1609.02907},
  archivePrefix={arXiv}
}

@inproceedings{velickovic2018gat,
  title={Graph Attention Networks},
  author={Veli{\v{c}}kovi{\'c}, Petar and Cucurull, Guillem and Casanova, Arantxa and Romero, Adriana and Li{\`o}, Pietro and Bengio, Yoshua},
  booktitle={International Conference on Learning Representations (ICLR)},
  year={2018},
  eprint={1710.10903},
  archivePrefix={arXiv}
}

@inproceedings{hamilton2017sage,
  title={Inductive Representation Learning on Large Graphs},
  author={Hamilton, William L. and Ying, Rex and Leskovec, Jure},
  booktitle={Advances in Neural Information Processing Systems (NeurIPS)},
  year={2017},
  eprint={1706.02216},
  archivePrefix={arXiv}
}

@inproceedings{reimers2019sbert,
  title={{Sentence-BERT}: Sentence Embeddings using Siamese {BERT}-Networks},
  author={Reimers, Nils and Gurevych, Iryna},
  booktitle={Conference on Empirical Methods in Natural Language Processing (EMNLP)},
  pages={3982--3992},
  year={2019},
  eprint={1908.10084},
  archivePrefix={arXiv}
}

@misc{feng2024taglas,
  title={{TAGLAS}: An Atlas of Text-Attributed Graph Datasets in the Era of Large Graph and Language Models},
  author={Feng, Jiarui and Liu, Hao and Kong, Lecheng and Zhu, Mingfang and Chen, Yixin and Zhang, Muhan},
  year={2024},
  eprint={2406.14683},
  archivePrefix={arXiv},
  primaryClass={cs.LG}
}

@inproceedings{amara2022graphframex,
  title={{GraphFramEx}: Towards Systematic Evaluation of Explainability Methods for Graph Neural Networks},
  author={Amara, Kenza and Ying, Rex and Zhang, Zitao and Han, Zhihao and Shan, Yinan and Brandes, Ulrik and Schemm, Sebastian and Zhang, Ce},
  booktitle={Learning on Graphs Conference (LoG)},
  year={2022},
  eprint={2206.09677},
  archivePrefix={arXiv}
}

@inproceedings{amara2023ginxeval,
  title={{GInX-Eval}: Towards In-Distribution Evaluation of Graph Neural Network Explanations},
  author={Amara, Kenza and El-Assady, Mennatallah and Ying, Rex},
  booktitle={International Conference on Learning Representations (ICLR)},
  year={2024},
  eprint={2309.16223},
  archivePrefix={arXiv}
}

@inproceedings{hu2020ogb,
  title={Open Graph Benchmark: Datasets for Machine Learning on Graphs},
  author={Hu, Weihua and Fey, Matthias and Zitnik, Marinka and Dong, Yuxiao and Ren, Hongyu and Liu, Bowen and Catasta, Michele and Leskovec, Jure},
  booktitle={Advances in Neural Information Processing Systems (NeurIPS)},
  year={2020},
  eprint={2005.00687},
  archivePrefix={arXiv}
}

@inproceedings{liu2022airgnn,
  title={Graph Neural Networks with Adaptive Residual},
  author={Liu, Xiaorui and Ding, Jiayuan and Jin, Wei and Xu, Han and Ma, Yao and Liu, Zitao and Tang, Jiliang},
  booktitle={Advances in Neural Information Processing Systems (NeurIPS)},
  year={2021},
  eprint={2110.15064},
  archivePrefix={arXiv}
}

@inproceedings{rong2020dropedge,
  title={{DropEdge}: Towards Deep Graph Convolutional Networks on Node Classification},
  author={Rong, Yu and Huang, Wenbing and Xu, Tingyang and Huang, Junzhou},
  booktitle={International Conference on Learning Representations (ICLR)},
  year={2020},
  eprint={1907.10903},
  archivePrefix={arXiv}
}

@inproceedings{tang2020investigating,
  title={Investigating and Mitigating Degree-Related Biases in Graph Convolutional Networks},
  author={Tang, Xianfeng and Yao, Huaxiu and Sun, Yiwei and Wang, Yiqi and Tang, Jiliang and Aggarwal, Charu and Mitra, Prasenjit and Wang, Suhang},
  booktitle={Proceedings of the 29th ACM International Conference on Information and Knowledge Management (CIKM)},
  pages={1435--1444},
  year={2020}
}

@inproceedings{huang2023tgb,
  title={Temporal Graph Benchmark for Machine Learning on Temporal Graphs},
  author={Huang, Shenyang and Poursafaei, Farimah and Danovitch, Jacob and Fey, Matthias and Hu, Weihua and Rossi, Emanuele and Leskovec, Jure and Bronstein, Michael and Rabusseau, Guillaume and Rabbany, Reihaneh},
  booktitle={Advances in Neural Information Processing Systems (NeurIPS) Datasets and Benchmarks Track},
  year={2023}
}

@inproceedings{dong2024pygdebias,
  title={{PyGDebias}: A Python Library for Debiasing in Graph Learning},
  author={Dong, Yushun and Wang, Zhenyu and Liu, Zaiyi and Zhao, Han and Wang, Suhang and Li, Jundong},
  booktitle={Proceedings of the 17th ACM International Conference on Web Search and Data Mining (WSDM)},
  year={2024}
}

@inproceedings{sun2022gppt,
  title={{GPPT}: Graph Pre-training and Prompt Tuning to Generalize Graph Neural Networks},
  author={Sun, Mingchen and Zhou, Kaixiong and He, Xin and Wang, Ying and Wang, Xin},
  booktitle={Proceedings of the 28th ACM SIGKDD Conference on Knowledge Discovery and Data Mining (KDD)},
  year={2022}
}

@inproceedings{qiu2020gcc,
  title={{GCC}: Graph Contrastive Coding for Graph Neural Network Pre-Training},
  author={Qiu, Jiezhong and Chen, Qibin and Dong, Yuxiao and Zhang, Jing and Yang, Hongxia and Ding, Ming and Wang, Kuansan and Tang, Jie},
  booktitle={Proceedings of the 26th ACM SIGKDD Conference on Knowledge Discovery and Data Mining (KDD)},
  year={2020},
  eprint={2006.09963},
  archivePrefix={arXiv}
}

@article{xia2024anygraph,
  title={{AnyGraph}: Graph Foundation Model in the Wild},
  author={Xia, Lianghao and Huang, Chao},
  journal={arXiv preprint arXiv:2408.10700},
  year={2024},
  eprint={2408.10700},
  archivePrefix={arXiv}
}

@inproceedings{fang2023gpf,
  title={Universal Prompt Tuning for Graph Neural Networks},
  author={Fang, Taoran and Zhang, Yunchao and Yang, Yang and Wang, Chunping and Chen, Lei},
  booktitle={Advances in Neural Information Processing Systems (NeurIPS)},
  year={2023},
  eprint={2209.15240},
  archivePrefix={arXiv}
}

@inproceedings{tang2024graphgpt,
  title={{GraphGPT}: Graph Instruction Tuning for Large Language Models},
  author={Tang, Jiabin and Yang, Yuhao and Wei, Wei and Shi, Lei and Su, Lixin and Cheng, Suqi and Yin, Dawei and Huang, Chao},
  booktitle={Proceedings of the 47th International ACM SIGIR Conference on Research and Development in Information Retrieval (SIGIR)},
  year={2024},
  eprint={2310.13023},
  archivePrefix={arXiv}
}

@inproceedings{huang2023prodigy,
  title={{PRODIGY}: Enabling In-context Learning Over Graphs},
  author={Huang, Qian and Ren, Hongyu and Chen, Peng and Kr{\v{z}}manc, Gregor and Zeng, Daniel and Liang, Percy and Leskovec, Jure},
  booktitle={Advances in Neural Information Processing Systems (NeurIPS)},
  year={2023},
  eprint={2305.12600},
  archivePrefix={arXiv}
}

@inproceedings{li2024zerog,
  title={{ZeroG}: Investigating Cross-dataset Zero-shot Transferability in Graphs},
  author={Li, Yuhan and Wang, Peisong and Li, Zhixun and Yu, Jeffrey Xu and Li, Jia},
  booktitle={Proceedings of the 30th ACM SIGKDD Conference on Knowledge Discovery and Data Mining (KDD)},
  year={2024},
  eprint={2402.11235},
  archivePrefix={arXiv}
}

@inproceedings{velickovic2019dgi,
  title={Deep Graph Infomax},
  author={Veli{\v{c}}kovi{\'c}, Petar and Fedus, William and Hamilton, William L. and Li{\`o}, Pietro and Bengio, Yoshua and Hjelm, R Devon},
  booktitle={International Conference on Learning Representations (ICLR)},
  year={2019},
  eprint={1809.10341},
  archivePrefix={arXiv}
}

@inproceedings{you2020graphcl,
  title={Graph Contrastive Learning with Augmentations},
  author={You, Yuning and Chen, Tianlong and Sui, Yongduo and Chen, Ting and Wang, Zhangyang and Shen, Yang},
  booktitle={Advances in Neural Information Processing Systems (NeurIPS)},
  year={2020},
  eprint={2010.13902},
  archivePrefix={arXiv}
}

@inproceedings{zheng2021grb,
  title={Graph Robustness Benchmark: Benchmarking the Adversarial Robustness of Graph Machine Learning},
  author={Zheng, Qinkai and Zou, Xu and Dong, Yuxiao and Cen, Yukuo and Yin, Da and Xu, Jiarong and Yang, Yang and Tang, Jie},
  booktitle={Advances in Neural Information Processing Systems (NeurIPS) Datasets and Benchmarks Track},
  year={2021},
  eprint={2111.04314},
  archivePrefix={arXiv}
}

@inproceedings{li2020deeprobust,
  title={{DeepRobust}: A Platform for Adversarial Attacks and Defenses},
  author={Li, Yaxin and Jin, Wei and Xu, Han and Tang, Jiliang},
  booktitle={Proceedings of the AAAI Conference on Artificial Intelligence (AAAI), Demonstration Track},
  year={2021},
  eprint={2005.06149},
  archivePrefix={arXiv}
}

@article{ma2025cilg,
  title={Class-Imbalanced Learning on Graphs: A Survey},
  author={Ma, Yihong and Tian, Yijun and Moniz, Nuno and Chawla, Nitesh V.},
  journal={ACM Computing Surveys},
  volume={57},
  number={8},
  year={2025},
  doi={10.1145/3718734},
  eprint={2304.04300},
  archivePrefix={arXiv}
}

@inproceedings{zhao2021graphsmote,
  title={{GraphSMOTE}: Imbalanced Node Classification on Graphs with Graph Neural Networks},
  author={Zhao, Tianxiang and Zhang, Xiang and Wang, Suhang},
  booktitle={Proceedings of the 14th ACM International Conference on Web Search and Data Mining (WSDM)},
  year={2021},
  eprint={2103.08826},
  archivePrefix={arXiv}
}

@inproceedings{park2022graphens,
  title={{GraphENS}: Neighbor-Aware Ego Network Synthesis for Class-Imbalanced Node Classification},
  author={Park, Joonhyung and Song, Jaeyun and Yang, Eunho},
  booktitle={International Conference on Learning Representations (ICLR)},
  year={2022}
}

@article{bommasani2021foundation,
  title={On the Opportunities and Risks of Foundation Models},
  author={Bommasani, Rishi and Hudson, Drew A. and Adeli, Ehsan and Altman, Russ and Arora, Simran and von Arx, Sydney and Bernstein, Michael S. and Bohg, Jeannette and Bosselut, Antoine and Brunskill, Emma and others},
  journal={arXiv preprint arXiv:2108.07258},
  year={2021},
  eprint={2108.07258},
  archivePrefix={arXiv}
}

@article{stokes2020antibiotic,
  title={A Deep Learning Approach to Antibiotic Discovery},
  author={Stokes, Jonathan M. and Yang, Kevin and Swanson, Kyle and Jin, Wengong and Cubillos-Ruiz, Andres and Donghia, Nina M. and MacNair, Craig R. and French, Shawn and Carfrae, Lindsey A. and Bloom-Ackermann, Zohar and others},
  journal={Cell},
  volume={180},
  number={4},
  pages={688--702.e13},
  year={2020},
  doi={10.1016/j.cell.2020.01.021}
}

@inproceedings{ying2018pinsage,
  title={Graph Convolutional Neural Networks for Web-Scale Recommender Systems},
  author={Ying, Rex and He, Ruining and Chen, Kaifeng and Eksombatchai, Pong and Hamilton, William L. and Leskovec, Jure},
  booktitle={Proceedings of the 24th ACM SIGKDD International Conference on Knowledge Discovery and Data Mining (KDD)},
  year={2018},
  eprint={1806.01973},
  archivePrefix={arXiv}
}

@inproceedings{dou2020caregnn,
  title={Enhancing Graph Neural Network-based Fraud Detectors against Camouflaged Fraudsters},
  author={Dou, Yingtong and Liu, Zhiwei and Sun, Li and Deng, Yutong and Peng, Hao and Yu, Philip S.},
  booktitle={Proceedings of the 29th ACM International Conference on Information and Knowledge Management (CIKM)},
  year={2020},
  eprint={2008.08692},
  archivePrefix={arXiv}
}

@inproceedings{devlin2019bert,
  title={{BERT}: Pre-training of Deep Bidirectional Transformers for Language Understanding},
  author={Devlin, Jacob and Chang, Ming-Wei and Lee, Kenton and Toutanova, Kristina},
  booktitle={Proceedings of the 2019 Conference of the North American Chapter of the Association for Computational Linguistics: Human Language Technologies (NAACL-HLT)},
  year={2019},
  eprint={1810.04805},
  archivePrefix={arXiv}
}

@inproceedings{radford2021clip,
  title={Learning Transferable Visual Models From Natural Language Supervision},
  author={Radford, Alec and Kim, Jong Wook and Hallacy, Chris and Ramesh, Aditya and Goh, Gabriel and Agarwal, Sandhini and Sastry, Girish and Askell, Amanda and Mishkin, Pamela and Clark, Jack and Krueger, Gretchen and Sutskever, Ilya},
  booktitle={Proceedings of the 38th International Conference on Machine Learning (ICML)},
  year={2021},
  eprint={2103.00020},
  archivePrefix={arXiv}
}

@inproceedings{perozzi2014deepwalk,
  author={Perozzi, Bryan and Al-Rfou, Rami and Skiena, Steven},
  title={{DeepWalk}: Online Learning of Social Representations},
  booktitle={Proceedings of the 20th ACM SIGKDD International Conference on Knowledge Discovery and Data Mining (KDD)},
  pages={701--710},
  year={2014}
}

@inproceedings{grover2016node2vec,
  author={Grover, Aditya and Leskovec, Jure},
  title={node2vec: Scalable Feature Learning for Networks},
  booktitle={Proceedings of the 22nd ACM SIGKDD International Conference on Knowledge Discovery and Data Mining (KDD)},
  pages={855--864},
  year={2016}
}

@inproceedings{tang2015line,
  author={Tang, Jian and Qu, Meng and Wang, Mingzhe and Zhang, Ming and Yan, Jun and Mei, Qiaozhu},
  title={{LINE}: Large-scale Information Network Embedding},
  booktitle={Proceedings of the 24th International Conference on World Wide Web (WWW)},
  pages={1067--1077},
  year={2015}
}

@inproceedings{ribeiro2017struc2vec,
  author={Ribeiro, Leonardo F. R. and Saverese, Pedro H. P. and Figueiredo, Daniel R.},
  title={struc2vec: Learning Node Representations from Structural Identity},
  booktitle={Proceedings of the 23rd ACM SIGKDD International Conference on Knowledge Discovery and Data Mining (KDD)},
  pages={385--394},
  year={2017}
}

@inproceedings{zhang2019heterogeneous,
  title={Heterogeneous graph neural network},
  author={Zhang, Chuxu and Song, Dongjin and Huang, Chao and Swami, Ananthram and Chawla, Nitesh V},
  booktitle={Proceedings of the 25th ACM SIGKDD international conference on knowledge discovery and  data mining (KDD)},
  pages={793--803},
  year={2019}
}

@inproceedings{tian2023heterogeneous,
  title={Heterogeneous graph masked autoencoders},
  author={Tian, Yijun and Dong, Kaiwen and Zhang, Chunhui and Zhang, Chuxu and Chawla, Nitesh V},
  booktitle={Proceedings of the AAAI conference on artificial intelligence (AAAI)},
  volume={37},
  number={8},
  pages={9997--10005},
  year={2023}
}

@inproceedings{ju2023multi,
  title={Multi-task Self-supervised Graph Neural Networks Enable Stronger Task Generalization},
  author={Ju, Mingxuan and Zhao, Tong and Wen, Qianlong and Yu, Wenhao and Shah, Neil and Ye, Yanfang and Zhang, Chuxu},
  booktitle={International Conference on Learning Representations (ICLR)},
  year={2023}
}

@article{wang2026safety,
  title={Safety in graph machine learning: Threats and safeguards},
  author={Wang, Song and Dong, Yushun and Zhang, Binchi and Chen, Zihan and Fu, Xingbo and He, Yinhan and Shen, Cong and Zhang, Chuxu and Chawla, Nitesh V and Li, Jundong},
  journal={IEEE Transactions on Knowledge and Data Engineering},
  year={2026},
  publisher={IEEE}
}

@inproceedings{wanggenerative,
  title={Generative Graph Pattern Machine},
  author={Wang, Zehong and Zhang, Zheyuan and Ma, Tianyi and Zhang, Chuxu and Ye, Yanfang},
  booktitle={The Thirty-ninth Annual Conference on Neural Information Processing Systems},
  year={2025}
}

@inproceedings{wang2025beyond,
  title={Beyond Message Passing: Neural Graph Pattern Machine},
  author={Wang, Zehong and Zhang, Zheyuan and Ma, Tianyi and Chawla, Nitesh V and Zhang, Chuxu and Ye, Yanfang},
  booktitle={International Conference on Machine Learning},
  pages={65496--65517},
  year={2025},
  organization={PMLR}
}

@article{li2026graph,
  title={Graph is a Substrate Across Data Modalities},
  author={Li, Ziming and Wu, Xiaoming and Wang, Zehong and Li, Jiazheng and Tian, Yijun and Bi, Jinhe and Ma, Yunpu and Ye, Yanfang and Zhang, Chuxu},
  journal={arXiv preprint arXiv:2601.22384},
  year={2026}
}

@inproceedings{DBLP:conf/ijcai/LiLLLZ25,
  author       = {Ziming Li and
                  Youhuan Li and
                  Yuyu Luo and
                  Guoliang Li and
                  Chuxu Zhang},
  title        = {Graph Neural Networks for Databases: {A} Survey},
  booktitle    = {Proceedings of the Thirty-Fourth International Joint Conference on
                  Artificial Intelligence(IJCAI)},
  year         = {2025}
}

@inproceedings{DBLP:conf/icde/LiLCZLY024,
  author       = {Ziming Li and
                  Youhuan Li and
                  Xinhuan Chen and
                  Lei Zou and
                  Yang Li and
                  Xiaofeng Yang and
                  Hongbo Jiang},
  title        = {NewSP: {A} New Search Process for Continuous Subgraph Matching over
                  Dynamic Graphs},
  booktitle    = {{IEEE} International Conference on Data Engineering(ICDE)},
  year         = {2024}
}

@article{stemgnn2026,
  title={Generalizing GNNs with Tokenized Mixture of Experts},
  author={Guo, Xiaoguang and Wang, Zehong and Li, Jiazheng and Spitzel, Shawn and Yang, Qi and Ding, Kaize and Li, Jundong and Zhang, Chuxu},
  journal={arXiv preprint arXiv:2602.09258},
  year={2026}
}

@article{wang2025grm,
  title={Generative Risk Minimization for Out-of-Distribution Generalization on Graphs},
  author={Wang, Song and Tan, Zhen and Zhu, Yaochen and Zhang, Chuxu and Li, Jundong},
  journal={Transactions on Machine Learning Research},
  year={2025},
  note={arXiv:2502.07968}
}

@inproceedings{yuan2024dragon,
  title={Mitigating Emergent Robustness Degradation while Scaling Graph Learning},
  author={Yuan, Xiangchi and Zhang, Chunhui and Tian, Yijun and Ye, Yanfang and Zhang, Chuxu},
  booktitle={International Conference on Learning Representations},
  year={2024}
}

@inproceedings{yuan2024graphcsl,
  title={Graph Cross Supervised Learning via Generalized Knowledge},
  author={Yuan, Xiangchi and Tian, Yijun and Zhang, Chunhui and Ye, Yanfang and Chawla, Nitesh V and Zhang, Chuxu},
  booktitle={Proceedings of the 30th ACM SIGKDD Conference on Knowledge Discovery and Data Mining},
  pages={4083--4094},
  year={2024},
  doi={10.1145/3637528.3671830}
}

\clearpage
\appendix


\section{Datasets}
\label{app:datasets}

\paragraph{Dataset statistics.} We use twenty-five text-attributed
graphs spanning seven domains, sourced from
TSGFM~\citep{chen2024tsgfm} and TAGLAS~\citep{feng2024taglas}.
Per-dataset statistics (domain, task, node and edge counts, class
count, source) are reported in Table~\ref{tab:app_datasets}. We
use the canonical train/validation/test split provided by each
source repository, and repeat each method-dataset cell with five
random seeds that re-randomize model initialization. Because all
datasets are text-attributed, we use
Sentence-BERT~\citep{reimers2019sbert} to encode each canonical
text attribute (paper abstract, product description, item title,
atom symbol, or relation phrase) into a $768$-dimensional vector,
yielding a single feature space shared across methods, consistent
with the view of graphs as a unifying substrate across heterogeneous
data modalities~\citep{li2026graph}.

\begin{table}[!htbp]
\centering
\caption{Datasets used in \benchname. \textbf{Task}: NC = node
classification, LP = link prediction, GC = graph classification.
For graph-classification datasets, \#Nodes and \#Edges are
aggregated across all graphs in the dataset, and \#Classes
denotes the number of tasks in multi-task molecular benchmarks.}
\label{tab:app_datasets}
\small
\setlength{\tabcolsep}{4pt}
\begin{tabular}{llcrrrl}
\toprule
\textbf{Dataset} & \textbf{Domain} & \textbf{Task} & \textbf{\#Nodes} & \textbf{\#Edges} & \textbf{\#Classes} & \textbf{Source} \\
\midrule
Cora           & Academic        & NC & 2{,}708       & 10{,}556       & 7   & \citep{chen2024tsgfm} \\
CiteSeer       & Academic        & NC & 3{,}186       & 8{,}450        & 6   & \citep{chen2024tsgfm} \\
PubMed         & Academic        & NC & 19{,}717      & 88{,}648       & 3   & \citep{chen2024tsgfm} \\
arXiv          & Academic        & NC & 169{,}343     & 2{,}315{,}598  & 40  & \citep{chen2024tsgfm} \\
arXiv23        & Academic        & NC & 46{,}198      & 77{,}726       & 40  & \citep{chen2024tsgfm} \\
arXivYear      & Academic        & NC & 169{,}343     & 2{,}315{,}598  & 5   & \citep{chen2024tsgfm} \\
DBLP           & Academic        & NC & 14{,}376      & 431{,}326      & 4   & \citep{chen2024tsgfm} \\
\midrule
Elec-Computers & E-commerce      & NC & 87{,}229      & 1{,}256{,}548  & 10  & \citep{chen2024tsgfm} \\
ElePhoto       & E-commerce      & NC & 48{,}362      & 873{,}793      & 12  & \citep{chen2024tsgfm} \\
SportsFit      & E-commerce      & NC & 173{,}055     & 3{,}020{,}134  & 13  & \citep{chen2024tsgfm} \\
AmazonRatings  & E-commerce      & NC & 24{,}492      & 186{,}100      & 5   & \citep{chen2024tsgfm} \\
\midrule
BookChild      & Book            & NC & 76{,}875      & 2{,}325{,}044  & 24  & \citep{chen2024tsgfm} \\
BookHis        & Book            & NC & 41{,}551      & 503{,}180      & 12  & \citep{chen2024tsgfm} \\
\midrule
WikiCS         & Web platform    & NC & 11{,}701      & 431{,}726      & 10  & \citep{chen2024tsgfm} \\
Tolokers       & Web platform    & NC & 11{,}758      & 519{,}000      & 2   & \citep{feng2024taglas} \\
\midrule
WN18RR         & Knowledge graph & LP & 40{,}943      & 93{,}003       & 11  & \citep{feng2024taglas} \\
FB15K237       & Knowledge graph & LP & 14{,}541      & 310{,}116      & 237 & \citep{feng2024taglas} \\
\midrule
ML1M           & Recommendation  & LP & 9{,}923       & 2{,}000{,}418  & 5   & \citep{feng2024taglas} \\
\midrule
BACE           & Molecule        & GC & 51{,}577      & 111{,}536      & 1   & \citep{feng2024taglas} \\
BBBP           & Molecule        & GC & 49{,}068      & 105{,}842      & 1   & \citep{feng2024taglas} \\
ChemHIV        & Molecule        & GC & 1{,}049{,}163 & 2{,}259{,}376  & 1   & \citep{feng2024taglas} \\
CYP450         & Molecule        & GC & 414{,}367     & 895{,}886      & 5   & \citep{feng2024taglas} \\
MUV            & Molecule        & GC & 2{,}255{,}846 & 4{,}892{,}252  & 17  & \citep{feng2024taglas} \\
Tox21          & Molecule        & GC & 145{,}459     & 302{,}190      & 12  & \citep{feng2024taglas} \\
ToxCast        & Molecule        & GC & 161{,}002     & 330{,}180      & 588 & \citep{feng2024taglas} \\
\bottomrule
\end{tabular}
\end{table}

\paragraph{Note.}
Pretrained methods additionally use ChemBLPre and ChemPCBA as
pretraining-only molecular corpora. These corpora are excluded from
downstream evaluation and therefore not counted among the 25 benchmark
datasets in Table~\ref{tab:app_datasets}.

\paragraph{Licenses and credit.} All datasets are public academic
graphs accessed through the
TSGFM~\citep{chen2024tsgfm} and TAGLAS~\citep{feng2024taglas}
collections and are used in accordance with the licenses of their
original sources (for example, the OGB datasets~\citep{hu2020ogb}
under their MIT-style benchmark license, WikiCS under CC-BY-SA,
and MoleculeNet datasets under their respective open data
licenses). Method implementations (DeepWalk, Node2vec, GCN, GAT,
GraphSAGE, BGRL, GraphMAE, GFT, GIT, UniGraph2, OFA, LLaGA) are
cited at first mention and adapted from the authors' public
repositories under their respective open-source licenses (MIT,
Apache 2.0, or BSD as released); no license terms are modified.

\section{Methods}
\label{app:methods}

This appendix lists the twelve graph representation models evaluated
in \benchname, grouped by tier (\S\ref{sec:bench:models}). Pretraining
and downstream-adaptation hyperparameters for each method are
documented separately in
Appendix~\ref{app:training} (Implementation details).

\subsection{Shallow graph embeddings}

\paragraph{DeepWalk~\citep{perozzi2014deepwalk}.} A shallow node
embedding method that learns representations by applying Skip-gram
to uniform random walks over the graph.

\paragraph{Node2vec~\citep{grover2016node2vec}.} A shallow node
embedding method that extends DeepWalk with biased second-order
random walks interpolating between breadth- and depth-first
neighbourhood exploration.

\subsection{Supervised GNNs}

\paragraph{GCN~\citep{kipf2017gcn}.} A message-passing GNN that
aggregates neighbour features through symmetric normalised graph
convolutions.

\paragraph{GAT~\citep{velickovic2018gat}.} A GNN that aggregates
neighbour features through learned multi-head attention weights.

\paragraph{GraphSAGE~\citep{hamilton2017sage}.} A GNN that
generates node representations by sampling and aggregating
features from a node's local neighbourhood.

\subsection{Self-supervised graph models}

\paragraph{BGRL~\citep{thakoor2022bgrl}.} A self-supervised
graph representation method that learns by bootstrapping latent
target predictions across two augmented views, without negative
samples.

\paragraph{GraphMAE~\citep{hou2022graphmae}.} A self-supervised
graph representation method that uses masked-feature
reconstruction with a scaled cosine error to pretrain a
generative graph autoencoder.

\subsection{Graph foundation models}

\paragraph{GFT~\citep{wang2024gft}.} A graph foundation model
that converts cross-domain graph tasks into a shared
computation-tree vocabulary and adapts to downstream tasks via
prototype matching against learned tree codes.

\paragraph{GIT~\citep{wang2024git}.} A graph foundation model
pretrained with task-tree reconstruction over node-, edge-, and
graph-level tasks, learning generalities across heterogeneous
graphs.

\paragraph{UniGraph2~\citep{he2025unigraph2}.} A multi-domain
graph foundation model that aligns features from different
domains through routed mixture-of-experts on top of a shared
embedding space.

\paragraph{OFA~\citep{liu2024ofa}.} A graph foundation model
that unifies node, link, and graph tasks by reformulating each
query as a binary link-prediction problem over an augmented
prompt graph containing class and task-of-interest nodes.

\paragraph{LLaGA~\citep{chen2024llaga}.} An LLM-based graph
foundation model that projects a node's local neighbourhood into
a Vicuna-7B token sequence and reads predictions from the
language-model head.

\section{Implementation details}
\label{app:training}

This appendix documents the pretraining and downstream-adaptation
hyperparameter settings used in \benchname. Method descriptions
are in Appendix~\ref{app:methods}; per-axis stress-operator
definitions are in Appendix~\ref{app:safety}.

\subsection{Pretraining hyperparameters}
\label{app:training:pretrain}

The two self-supervised methods (BGRL, GraphMAE) and the five
graph foundation models (GFT, GIT, UniGraph2, OFA, LLaGA)
pretrain on the same nine text-attributed graphs (cora, pubmed,
arxiv, wikics, WN18RR, FB15K237, ChemHIV, ChemBLPre, ChemPCBA)
adopted by GFT~\citep{wang2024gft} and OFA~\citep{liu2024ofa}.
Each method optimizes its native objective on this shared corpus,
with hyperparameters following its source repository. The three
supervised baselines (GCN, GAT, GraphSAGE) and the two
topology-only methods (DeepWalk, Node2vec) have no pretraining
stage; they are trained end-to-end on each downstream graph
(see Appendix~\ref{app:training:downstream}). Per-method settings
for the pretrained methods are as follows.

\begin{enumerate}
\item For BGRL~\citep{thakoor2022bgrl}: \texttt{num\_layers=2,
num\_hidden=768, lr=1e-4, weight\_decay=1e-5, max\_epoch=50,
batch\_size=1024, ema\_tau=0.99, warmup\_frac=0.1, drop\_edge=0.2,
drop\_feat=0.2, predictor\_hidden=512, optimizer=AdamW}.

\item For GraphMAE~\citep{hou2022graphmae}:
\texttt{num\_layers=2, num\_hidden=768, num\_heads=4, lr=1e-3,
weight\_decay=2e-4, max\_epoch=50, batch\_size=1024,
mask\_rate=0.5, replace\_rate=0.0, loss=sce, alpha\_l=3.0,
in\_drop=0.2, attn\_drop=0.1, optimizer=AdamW}.

\item For GFT~\citep{wang2024gft}: \texttt{num\_layers=2,
hidden\_dim=768, dropout=0.15, pretrain\_lr=1e-4,
pretrain\_weight\_decay=1e-5, pretrain\_epochs=50,
pretrain\_batch\_size=1024, codebook\_size=128, codebook\_head=4,
codebook\_decay=0.8, feat\_p=0.2, edge\_p=0.2,
topo\_recon\_ratio=0.1, optimizer=AdamW}.

\item For GIT~\citep{wang2024git}: \texttt{num\_layers=2,
hidden\_dim=768, dropout=0.15, lr=1e-7, weight\_decay=1e-8,
epochs=20, batch\_size=4096, fanout=10, ema=0.99, feat\_p=0.2,
edge\_p=0.2, topo\_recon\_ratio=0.1, optimizer=AdamW}.

\item For UniGraph2~\citep{he2025unigraph2}: \texttt{num\_layers=3,
num\_hidden=768, num\_experts=8, num\_selected\_experts=2, lr=1e-4,
weight\_decay=1e-5, max\_epoch=50, batch\_size=64, gamma=2.0,
lambda\_spd=0.5, feat\_drop\_rate=0.1, edge\_mask\_rate=0.1,
fanout=10, optimizer=AdamW}.

\item For OFA~\citep{liu2024ofa}: \texttt{num\_layers=7,
num\_hidden=768, dropout=0.15, lr=1e-3, num\_epochs=50,
batch\_size=128, num\_relations=5, JK=none, optimizer=AdamW}.

\item For LLaGA~\citep{chen2024llaga}, following the original
two-stage recipe, only the projector is trained with the
Vicuna-7B language model frozen:
\texttt{llm\_backbone=Vicuna-7B (frozen), projector\_type=2-layer-mlp,
projector\_hidden=4096, lr=5e-4, weight\_decay=1e-4, epochs=10,
batch\_size=4, sample\_size=10, use\_hop=2, max\_text\_len=128,
max\_steps=5000, dtype=bfloat16, optimizer=AdamW}.
\end{enumerate}

\subsection{Downstream adaptation hyperparameters}
\label{app:training:downstream}

For each method-dataset cell we apply the method's native
downstream adaptation procedure end-to-end with AdamW and five
random seeds. Per-method settings are as follows.

\begin{enumerate}
\item For DeepWalk~\citep{perozzi2014deepwalk}, walk embeddings are
fitted on the downstream graph with Skip-gram on uniform random
walks ($p{=}q{=}1$): \texttt{emb\_dim=128, walk\_length=20,
walks\_per\_node=10, context\_size=10, num\_negative=1,
walk\_epochs=100, batch\_size=128, lr=1e-2,
optimizer=SparseAdam}. Predictions use a frozen-feature
$\ell_2$-regularised logistic-regression head (scikit-learn,
\texttt{C=1.0, max\_iter=1000, solver=lbfgs}).

\item For Node2vec~\citep{grover2016node2vec}: same Skip-gram and
logistic-regression configuration as DeepWalk, with biased
second-order walks (\texttt{p=1.0, q=0.5}, BFS-leaning).

\item For GCN, GAT, and GraphSAGE: \texttt{num\_layers=2,
num\_hidden=768, lr=1e-3, dropout=0.2, weight\_decay=1e-4,
max\_epochs=500, patience=200}.

\item For BGRL~\citep{thakoor2022bgrl}: \texttt{num\_layers=2,
num\_hidden=768, lr=5e-4, dropout=0.2, weight\_decay=1e-5,
max\_epochs=500, patience=200}.

\item For GraphMAE~\citep{hou2022graphmae}: \texttt{num\_layers=2,
num\_hidden=768, lr=1e-3, weight\_decay=1e-4, dropout=0.2,
max\_epochs=500, patience=200}.

\item For UniGraph2~\citep{he2025unigraph2}: \texttt{num\_layers=3,
num\_hidden=768, num\_experts=8, num\_selected\_experts=2,
lr=1e-3, weight\_decay=1e-4, dropout=0.2, max\_epochs=500,
patience=200}.

\item For GFT~\citep{wang2024gft}: \texttt{num\_layers=2,
hidden\_dim=768, dropout=0.15, finetune\_lr=5e-4,
finetune\_epochs=1000, early\_stop=200, batch\_size=full}.

\item For GIT~\citep{wang2024git}: \texttt{num\_layers=2,
hidden\_dim=768, dropout=0.15, lr=1e-4, sft\_epochs=100,
weight\_decay=0, batch\_size=full}.

\item For OFA~\citep{liu2024ofa}: \texttt{num\_layers=7,
num\_hidden=768, dropout=0.15, lr=1e-3, weight\_decay=0,
epochs=100, patience=30, prompt\_graph\_mode=subgraph,
subgraph\_hops=2}.

\item For LLaGA~\citep{chen2024llaga}: \texttt{lr=1e-3,
weight\_decay=1e-4, max\_epochs=15, patience=3, sample\_size=10,
max\_text\_len=128, batch\_size=8, dtype=bfloat16}.
\end{enumerate}

All methods use cross-entropy loss with their native prediction
head: GFM tier methods (GFT/GIT/UniGraph2) use prototype or
code-based classification; OFA uses class-node link scoring with
binary cross-entropy; LLaGA uses the language-model head over a
verbalised class vocabulary; the remaining methods use a linear
classification head. No class reweighting, focal loss, or
oversampling is applied.

\subsection{Compute resources}
\label{app:training:compute}

All experiments were run on a shared university HPC cluster
(SLURM scheduler, \texttt{general-gpu} partition) with single-GPU
jobs. The GPU pool consists of NVIDIA L40 (48\,GB VRAM) and NVIDIA
A100 (40\,GB VRAM) accelerators, allocated with 4--8 CPU cores and
32--192\,GB of system memory per job depending on dataset size.

Per method-dataset-seed cell, wall-clock time ranges from
approximately 5--30 minutes on small node-classification graphs
(Cora, CiteSeer, PubMed, ArXiv23, DBLP, AmazonRatings, WikiCS,
Tolokers, and BookHis), 30 minutes to 2 hours on medium graphs and
link-prediction datasets (ElePhoto, Elec-Computers, BookChild,
WN18RR, FB15K237, and ML1M) and on smaller molecular
graph-classification corpora (BACE, BBBP, Tox21, and ToxCast), and
2--6 hours on large node-classification graphs (arXiv, arXivYear,
and SportsFit) and larger molecular graph-classification corpora
(CYP450, ChemHIV, and MUV). Dataset statistics are reported in
Table~\ref{tab:app_datasets}.
Joint-corpus pretraining of each SSL backbone is a one-off cost on
the order of 8--24 GPU-hours.

The total compute for the experiments reported in this paper is
approximately on the order of $10^4$ GPU-hours, aggregating across
methods, datasets, safety axes, and seeds. The full research
project, including preliminary configurations and discarded
intermediate runs, required additional compute beyond the final
reported numbers.

\section{Safety dimension details}
\label{app:safety}

This appendix gives, for each of the five safety dimensions of
\benchname, the per-axis stress operator referenced in
\S\ref{sec:results}. Method-side hyperparameters are in
Appendix~\ref{app:training}; per-dimension extended numerical
results are in Appendix~\ref{app:extended_results}.

Dataset coverage is axis-specific. Degree-shift OOD is evaluated on
the node-classification datasets covered in this subcondition, with
the main text showing six representative datasets and the appendix
reporting the complete degree-shift results for this evaluated set.
Class imbalance is evaluated on eight categorical-label
node-classification datasets, while structural fairness is evaluated
on eight node-classification datasets with degree-based head-tail
groups. Interpretation is evaluated on eight node-classification
datasets for methods whose adaptation interface exposes edge-level
attribution. Temporal, scaffold, and inductive-entity shifts are
defined on datasets that provide the corresponding time, molecular
scaffold, or entity split.

Table~\ref{tab:app_coverage} summarises which (method, dimension)
cells are evaluated. All twelve methods are evaluated on the
corruption, OOD, imbalance, and fairness dimensions, with
sub-condition restrictions for the topology-only methods:
DeepWalk and Node2vec consume only graph structure, so under
corruption they are evaluated on the structure-noise channel only
(feature noise is inapplicable), and under OOD they are evaluated
on the node-classification shifts (degree and temporal) but not
on the graph-classification scaffold shift or the
link-prediction inductive-entity shift, which require
task-format adaptations they do not provide. The interpretation
dimension is restricted to the eight feature-consuming non-LLM
methods: OFA's prompt-graph reformulation reframes each query as
binary link prediction over an augmented prompt graph, which is
not directly compatible with the edge-saliency operator on the
original graph; LLaGA's frozen Vicuna-7B language-model head
precludes meaningful gradient saliency through the LLM; and
DeepWalk/Node2vec's inference path does not condition on
individual edges (random walks fold the graph into static node
embeddings before downstream prediction).

\providecommand{\cmark}{\ensuremath{\checkmark}}
\providecommand{\xmark}{\ensuremath{-}}

\begin{table}[!htbp]
\centering
\caption{Coverage of (method, safety dimension) cells in
\benchname. \cmark{} = evaluated; \xmark{} = not evaluated
(adaptation interface incompatible with the per-axis stress
operator; see paragraph above).}
\label{tab:app_coverage}
\small
\setlength{\tabcolsep}{6pt}
\renewcommand{\arraystretch}{1.05}
\begin{tabular}{lccccc}
\toprule
\textbf{Method} & \textbf{Corruption} & \textbf{OOD} & \textbf{Imbalance} & \textbf{Fairness} & \textbf{Interpretation} \\
\midrule
DeepWalk     & \cmark & \cmark & \cmark & \cmark & \xmark  \\
Node2vec     & \cmark & \cmark & \cmark & \cmark & \xmark  \\
\midrule
GCN          & \cmark & \cmark & \cmark & \cmark & \cmark \\
GAT          & \cmark & \cmark & \cmark & \cmark & \cmark \\
GraphSAGE    & \cmark & \cmark & \cmark & \cmark & \cmark \\
\midrule
BGRL         & \cmark & \cmark & \cmark & \cmark & \cmark \\
GraphMAE     & \cmark & \cmark & \cmark & \cmark & \cmark \\
\midrule
GFT          & \cmark & \cmark & \cmark & \cmark & \cmark \\
GIT          & \cmark & \cmark & \cmark & \cmark & \cmark \\
UniGraph2    & \cmark & \cmark & \cmark & \cmark & \cmark \\
OFA          & \cmark & \cmark & \cmark & \cmark & \xmark  \\
LLaGA        & \cmark & \cmark & \cmark & \cmark & \xmark  \\
\bottomrule
\end{tabular}
\end{table}

\subsection{Corruption}
\label{app:safety:corruption}

We evaluate robustness to two complementary input-corruption
operationalizations applied at inference time: \emph{feature
noise}, which perturbs node attributes
\citep{liu2022airgnn}, and \emph{structure noise}, which perturbs
graph topology \citep{rong2020dropedge}. In both cases the model
is trained on the clean graph, its parameters are frozen, and it
is then evaluated under six severity levels (severity~0 = clean
and five non-zero severities) with a deterministic per-(split,
severity) random seed for reproducibility. Performance is
measured with classification accuracy on node-classification
datasets, AUC-ROC on link-prediction datasets, and AUC on
graph-classification datasets.

\begin{figure}[h]
\centering
\includegraphics[width=\linewidth]{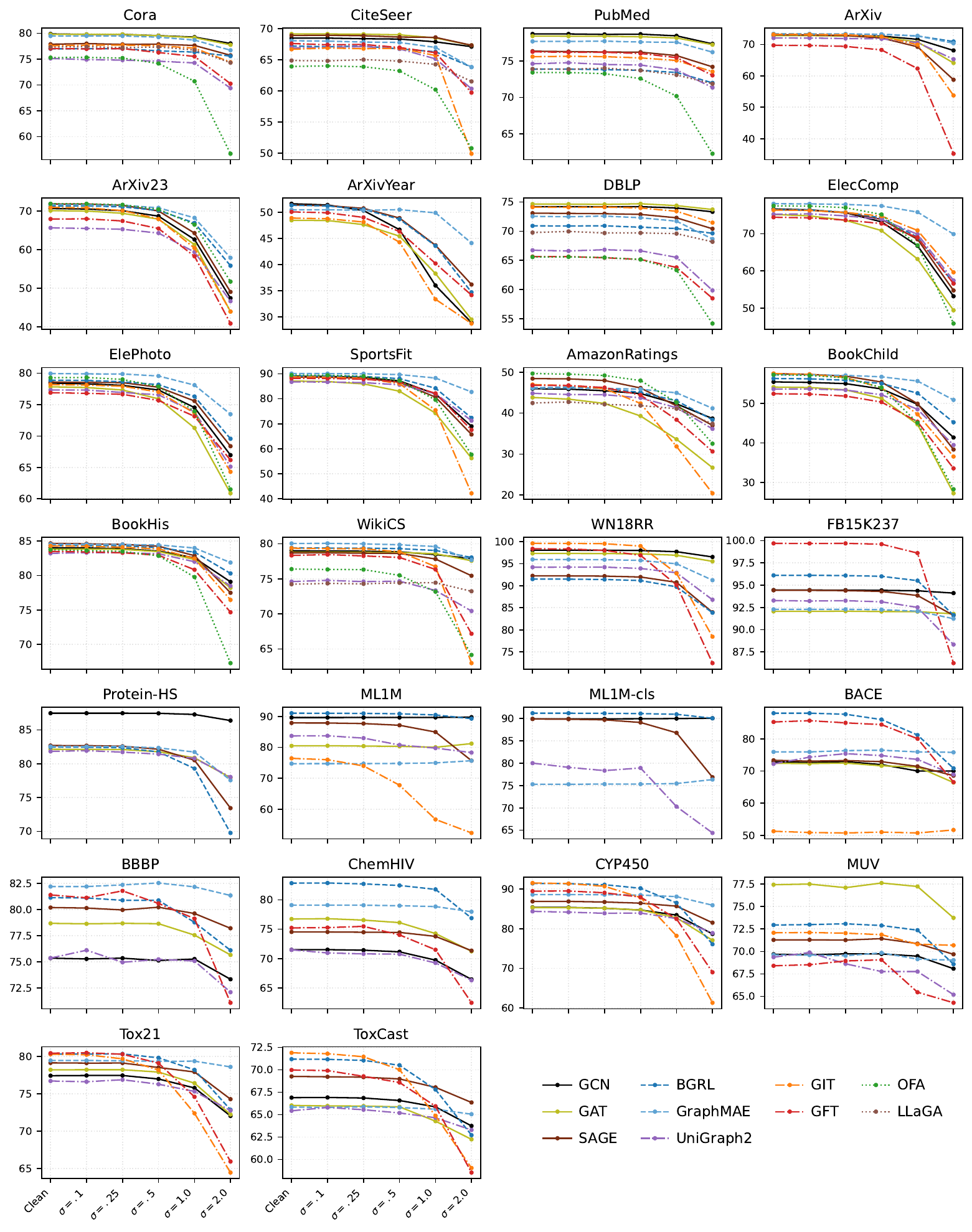}
\caption{Per-dataset feature-noise severity curves on \benchname{} datasets. Each panel reports test accuracy (NC) or AUC (LP, GC) at clean and the five relative noise scales ($\sigma_{\rm rel}{\in}\{0.1,0.25,0.5,1.0,2.0\}$).}
\label{fig:app_fn}
\end{figure}

\paragraph{Feature noise.}
At inference time, independent zero-mean Gaussian noise is added
to every node's feature vector, with the noise scaled per feature
dimension by the training-set per-dimension standard deviation:
\begin{equation*}
x_{\text{noisy}} = x + \sigma_{\text{rel}} \cdot
\mathrm{std}(X_{\text{train}}) \odot \varepsilon,
\qquad \varepsilon \sim \mathcal{N}(0, I).
\end{equation*}
The relative noise scale takes five values
$\sigma_{\text{rel}} \in \{0.1, 0.25, 0.5, 1.0, 2.0\}$. Edges,
labels, train/validation/test splits, and the trained model
parameters are not modified. Per-dataset severity curves across
all ten feature-consuming methods are reported in
Figure~\ref{fig:app_fn} (DeepWalk and Node2vec do not consume node
features and are therefore excluded from feature-noise evaluation).

\begin{figure}[h]
\centering
\includegraphics[width=\linewidth]{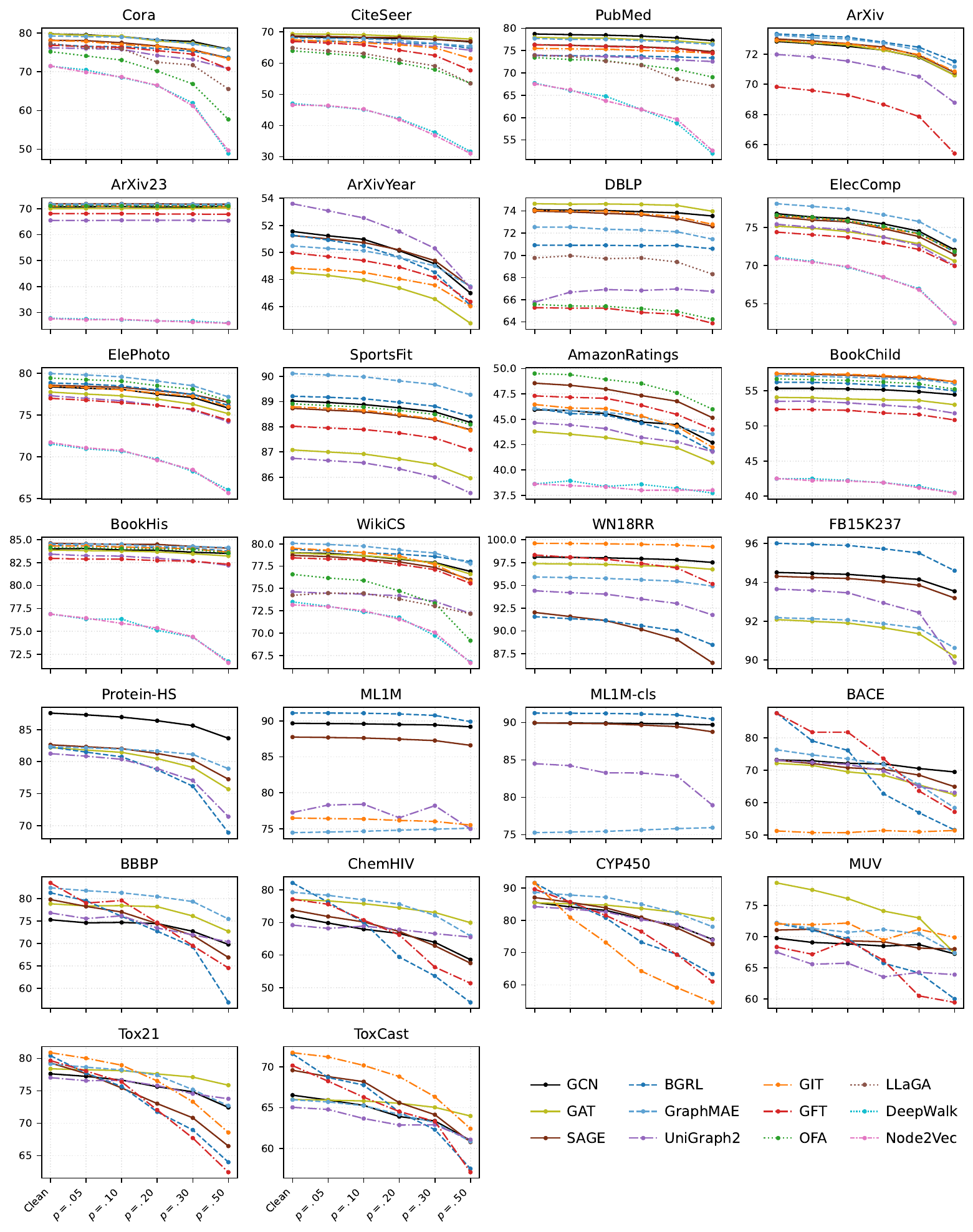}
\caption{Per-dataset edge-deletion severity curves on \benchname{} datasets. Each panel reports test accuracy (NC) or AUC (LP, GC) at clean and the five edge-drop probabilities ($p{\in}\{0.05,0.10,0.20,0.30,0.50\}$).}
\label{fig:app_ed}
\end{figure}

\paragraph{Structure noise.}
At inference time, a fraction $p$ of undirected edges is removed
uniformly at random; both directions of each selected edge are
dropped jointly so that the perturbed graph remains symmetric,
and self-loops (when present) are preserved. The drop rate takes
five values $p \in \{0.05, 0.10, 0.20, 0.30, 0.50\}$. Node
features, labels, train/validation/test splits, and the trained
model parameters are not modified. Per-dataset severity curves
across all twelve methods are reported in Figure~\ref{fig:app_ed}.

\subsection{Out-of-distribution generalization}
\label{app:safety:ood}

We evaluate four complementary graph-context shifts that target
different aspects of the input distribution: \emph{degree shift}
on node attributes' structural neighbourhood, \emph{temporal
shift} on publication time, \emph{scaffold shift} on molecular
substructure, and \emph{inductive KG shift} on the entity
vocabulary visible at test time. Following the spirit of
mechanism-level evaluation in graph OOD
benchmarks~\citep{gui2022good}, in every case the model is
trained on the in-distribution (ID) training partition and
evaluated on a held-out out-of-distribution (OOD) partition that
shares the task definition but differs in the targeted
distribution.

\paragraph{Degree shift.}
For each node-classification dataset, labelled nodes are sorted
by node degree in descending order; the top $60\%$ form the
training pool, the next $20\%$ form the OOD validation set, and
the bottom $20\%$ form the OOD test set. The protocol matches
GOOD's covariate-shift setup~\citep{gui2022good}. We report the
ID-to-OOD accuracy drop $\Delta = \mathrm{acc}_{\text{ID}} -
\mathrm{acc}_{\text{OOD}}$ per (method, dataset) cell. Per-dataset
heatmaps for the representative six-dataset subset are reported in
Figure~\ref{fig:ood_degree_heatmap} (main text), with full per-dataset
results in Appendix~\ref{app:ood_full}.

\paragraph{Temporal shift.}
On the arXiv citation network, training uses papers with
publication year $\leq 2010$ and OOD evaluation uses papers from
$\geq 2017$, with intermediate years 2011--2016 reserved as OOD
validation. The thirteen-year gap follows the aggressive
temporal-shift design of \citet{huang2023tgb} and reflects the
kind of evolving-graph context studied in continuous subgraph
search over dynamic
graphs~\citep{DBLP:conf/icde/LiLCZLY024} and, more broadly, in
graph neural networks for database
systems~\citep{DBLP:conf/ijcai/LiLLLZ25}. We report the
ID-to-OOD accuracy drop sorted across the twelve methods in
Figure~\ref{fig:ood_temporal_bars}.

\paragraph{Scaffold shift.}
On the molecular graph-classification datasets BACE and Tox21, we
compare random splits against Bemis--Murcko scaffold
splits~\citep{hu2020ogb}: molecules are grouped by their
Bemis--Murcko scaffold (computed from the canonical SMILES) and
the train, validation, and test sets receive disjoint scaffold
groups in an $80\!:\!10\!:\!10$ ratio. We report the AUC drop
$\Delta\mathrm{AUC} = \mathrm{AUC}_{\text{rand}} -
\mathrm{AUC}_{\text{scaffold}}$, so a smaller $\Delta\mathrm{AUC}$
indicates better generalization to unseen scaffolds.

\paragraph{Inductive KG shift.}
On the knowledge-graph link-prediction datasets WN18RR and
FB15K237, we construct an inductive split by randomly
partitioning the entity set into a training pool and a test pool
($75\!:\!25$). All facts whose head and tail are both in the
training pool are kept for downstream KG training; queries whose
candidate entity is in the test pool form the inductive test
queries. We report MRR and Hits@10, both standard ranking
metrics in inductive KG evaluation. Specifically, MRR is the
mean reciprocal rank of the correct entity over test queries,
and Hits@10 is the fraction of test queries where the correct
entity is ranked in the top 10 candidates. Test entities are
unseen during downstream training and require the method to
generalize beyond the entity vocabulary observed at training
time.

\subsection{Class imbalance}
\label{app:safety:imbalance}

We follow a step-imbalance protocol on each
node-classification dataset. Classes are sorted in ascending
order of their original training-set sample count; the lower
half are designated as \emph{minor} classes and the upper half
as \emph{major} classes. Let $n_{\text{major}} =
\max_{c \in \mathrm{major}} n_c$ denote the maximum training
count among major classes. Each minor class is then randomly
downsampled to
\begin{equation*}
n_{\text{minor}} = \max\!\bigl(1,\,
\lfloor n_{\text{major}} / \rho \rfloor\bigr)
\end{equation*}
training instances at imbalance ratio $\rho \in \{5, 10, 20\}$,
with the major-class training samples, the validation set, and
the test set left untouched. The per-cell sampling is
deterministic given the seed. We report balanced accuracy
(bAcc), macro-F1, and per-class recall on the test split,
averaged separately across major and minor classes. Per-dataset
per-class recall at all three imbalance ratios is reported in
Appendix~\ref{app:imb_full}.

\subsection{Fairness}
\label{app:safety:fairness}

We evaluate two complementary fairness operationalizations:
\emph{structural fairness}, which measures the disparity in
prediction quality between high- and low-degree nodes, and
\emph{demographic fairness}, which measures the disparity
between groups defined by a protected demographic attribute. In
both cases the model is trained without any fairness-aware
modification (no group reweighting, no constrained loss), so the
measurements isolate each method's intrinsic group disparity
under our unified fine-tuning protocol.

\paragraph{Structural fairness.}
Following the head/tail degree-disparity formulation of
\citet{tang2020investigating}, on each node-classification dataset
test nodes are partitioned by degree quantile: with
$q = 0.2$, the highest-degree $20\%$ of test nodes form the
\emph{head} group, and the lowest-degree $20\%$ form the
\emph{tail} group; the middle $60\%$ is not assigned to either
group for the gap measurement. We report per-group test accuracy
($\mathrm{acc}_{\text{head}}$, $\mathrm{acc}_{\text{tail}}$) and
the head$-$tail accuracy gap
$\Delta_{\text{HT}} = \mathrm{acc}_{\text{head}} -
\mathrm{acc}_{\text{tail}}$. A larger positive gap indicates a
more pronounced low-degree disadvantage. Per-dataset results
across the eight node-classification datasets are reported in
Appendix~\ref{app:fair_full}.

\paragraph{Demographic fairness.}
Demographic fairness is evaluated on the Tolokers
banned-user-prediction dataset, with the user education
attribute as a binary sensitive attribute ($s{=}1$ for users
with higher education, $s{=}0$ otherwise) following the protocol
of \citet{dong2024pygdebias}. We report three group-disparity
quantities at test time:
\begin{align*}
\Delta\mathrm{SP} &= |P(\hat{y}{=}1\,|\,s{=}0) -
                       P(\hat{y}{=}1\,|\,s{=}1)|,\\
\Delta\mathrm{EO} &= |P(\hat{y}{=}1\,|\,s{=}0,\,y{=}1) -
                       P(\hat{y}{=}1\,|\,s{=}1,\,y{=}1)|,\\
\Delta\mathrm{Util} &= |\mathrm{AUC}(s{=}0) -
                         \mathrm{AUC}(s{=}1)|.
\end{align*}
$\Delta\mathrm{SP}$ measures statistical-parity disparity,
$\Delta\mathrm{EO}$ measures equalized-odds disparity, and
$\Delta\mathrm{Util}$ measures the per-group AUC gap. Smaller
values indicate smaller group disparity. Full per-method results
are reported in Appendix~\ref{app:fair_full}.

\subsection{Interpretation}
\label{app:safety:interpret}

We evaluate post-hoc explanation fidelity at two task levels:
\emph{node-level interpretation} on node-classification datasets
via edge-saliency subgraph ablation, and \emph{graph-level
interpretation} on molecular graph-classification datasets via
atom-saliency node ablation. Both follow the GraphFramEx
characterization-score protocol of
\citet{amara2022graphframex}.

\paragraph{Node-level interpretation.}
For each test node we restrict the receptive field to the
$K$-hop computational subgraph ($K{=}2$) and compute per-edge
saliency by aggregating the gradient-based input attribution at
its two endpoints:
\begin{equation*}
S_{(u,v)} = \big\lVert \partial\,\mathrm{logit}_{\hat{y}}/
\partial x_u \big\rVert + \big\lVert
\partial\,\mathrm{logit}_{\hat{y}}/\partial x_v \big\rVert.
\end{equation*}
Edges are then ranked by $S_{(u,v)}$ and we evaluate at four
sparsity levels $k \in \{5, 10, 20, 50\}\%$. With $p_0$ the
clean predicted-class probability and $p_{+k}$, $p_{-k}$ the
probabilities after masking the top-$k$ and the complementary
edges respectively, we report
\begin{align*}
\mathrm{Fid}^+ &= p_0 - p_{+k}\quad\text{(higher better)},\\
\mathrm{Fid}^- &= p_0 - p_{-k}\quad\text{(lower better)},\\
\mathrm{char}  &= \frac{2\cdot\mathrm{Fid}^+\cdot
   (1-\mathrm{Fid}^-)}{\mathrm{Fid}^+ + (1-\mathrm{Fid}^-) +
   \epsilon}\quad\text{(higher better)},
\end{align*}
where $\mathrm{char}$ is the GraphFramEx characterization score,
a harmonic-style combination of $\mathrm{Fid}^+$ and
$(1-\mathrm{Fid}^-)$ with values in $[0,1]$. To control for
saliency-method specificity, every metric is reported for both
gradient saliency and a uniform-random baseline, and the
headline \emph{lift} is
$\Delta\mathrm{char} = \mathrm{char}_{\text{grad}} -
\mathrm{char}_{\text{random}}$. The full per-dataset sparsity
sweep and the per-dataset
$\mathrm{Fid}^+ / \mathrm{Fid}^-$ decomposition at $k{=}10\%$
are reported in Appendix~\ref{app:interpret_full}.

\paragraph{Graph-level interpretation.}
For each test molecule we compute per-atom saliency as the
$L_1$-norm gradient of the predicted-class logit with respect to
that atom's input features. Atoms are then ranked by saliency
and the same Fidelity quantities are computed by ablating the
top-$k$ or the complementary atoms (atom removal also drops the
incident bonds and pool entries). Edge saliency, when needed for
the subgraph ablation, is derived as the sum of the two endpoint
atom saliencies. Per-method graph-level interpretation results
on the molecular datasets are reported in
Appendix~\ref{app:interpret_full}.

\section{Extended Experimental Results}
\label{app:extended_results}

\subsection{Corruption}
\label{app:corruption_full}

\clearpage
\subsubsection{Feature Noise}
\label{app:corruption_fn_full}

\bigskip

\clearpage
\subsubsection{Edge Deletion}
\label{app:corruption_ed_full}

\FloatBarrier

\subsection{Out-of-Distribution Generalization --- full results}
\label{app:ood_full}







\begin{table}[!htbp]
\centering
\caption{Full per-dataset OOD degree-shift results across all 13 datasets (mean$\pm$std over 5 seeds). Each block shows up to four datasets side by side; per (dataset, column), \textbf{bold} = best; \underline{underline} = second. Gap $={\rm ID}-{\rm OOD}$, smaller $|{\rm Gap}|$ more robust. ``--'' indicates the method was not evaluated on that dataset.}
\label{tab:app_ood_degree_combined}
\scriptsize
\setlength{\tabcolsep}{1.6pt}
\renewcommand{\arraystretch}{1.0}
\centerline{%
\begin{tabular}{l|ccc|ccc|ccc|ccc}
\toprule
\textbf{Method} & \multicolumn{3}{c}{\textbf{cora}} & \multicolumn{3}{c}{\textbf{citeseer}} & \multicolumn{3}{c}{\textbf{pubmed}} & \multicolumn{3}{c}{\textbf{arxiv}} \\
\cmidrule(lr){2-4} \cmidrule(lr){5-7} \cmidrule(lr){8-10} \cmidrule(lr){11-13}
 & ID & OOD & Gap & ID & OOD & Gap & ID & OOD & Gap & ID & OOD & Gap \\
\midrule
DeepWalk & 86.79$_{\pm1.57}$ & 66.49$_{\pm1.04}$ & +20.30 & 71.22$_{\pm1.12}$ & 40.75$_{\pm1.98}$ & +30.47 & 83.13$_{\pm0.36}$ & 73.14$_{\pm0.55}$ & +9.99 & 73.67$_{\pm0.24}$ & 49.14$_{\pm0.16}$ & +24.53 \\
Node2Vec & 87.23$_{\pm1.66}$ & 65.83$_{\pm1.69}$ & +21.40 & 71.35$_{\pm1.75}$ & 41.07$_{\pm1.31}$ & +30.28 & 83.67$_{\pm0.39}$ & 73.35$_{\pm0.56}$ & +10.33 & 73.53$_{\pm0.35}$ & 49.07$_{\pm0.12}$ & +24.46 \\
\cmidrule(lr){2-13}
GCN & \underline{88.94}$_{\pm1.98}$ & 80.37$_{\pm1.90}$ & +8.58 & 80.69$_{\pm0.96}$ & 68.30$_{\pm1.51}$ & +12.39 & 88.99$_{\pm0.57}$ & 85.79$_{\pm0.58}$ & +3.20 & 78.63$_{\pm0.21}$ & 62.75$_{\pm0.13}$ & +15.88 \\
GAT & \textbf{89.00}$_{\pm1.90}$ & 81.20$_{\pm1.15}$ & +7.80 & 80.65$_{\pm1.26}$ & 67.72$_{\pm1.77}$ & +12.93 & 88.60$_{\pm0.44}$ & 84.27$_{\pm0.21}$ & +4.33 & 78.61$_{\pm0.16}$ & 61.88$_{\pm0.13}$ & +16.72 \\
SAGE & 87.13$_{\pm2.20}$ & 80.16$_{\pm0.96}$ & +6.97 & \underline{81.05}$_{\pm0.60}$ & 67.39$_{\pm1.86}$ & +13.66 & \textbf{90.27}$_{\pm0.58}$ & \underline{88.23}$_{\pm0.37}$ & \underline{+2.04} & 79.10$_{\pm0.29}$ & 61.86$_{\pm0.23}$ & +17.24 \\
\cmidrule(lr){2-13}
BGRL & 88.87$_{\pm1.79}$ & \textbf{82.21}$_{\pm0.91}$ & +6.66 & 80.34$_{\pm0.89}$ & \textbf{70.16}$_{\pm0.99}$ & \textbf{+10.18} & 89.42$_{\pm0.55}$ & 86.80$_{\pm0.55}$ & +2.62 & \underline{79.36}$_{\pm0.21}$ & \textbf{64.58}$_{\pm0.85}$ & \textbf{+14.78} \\
GraphMAE & 88.92$_{\pm2.06}$ & 81.67$_{\pm0.95}$ & +7.25 & 80.74$_{\pm1.47}$ & \underline{69.51}$_{\pm1.01}$ & \underline{+11.23} & 88.80$_{\pm0.44}$ & 85.07$_{\pm0.31}$ & +3.73 & \textbf{79.40}$_{\pm0.20}$ & \underline{63.58}$_{\pm0.44}$ & \underline{+15.82} \\
UniGraph2 & 85.80$_{\pm2.57}$ & 78.07$_{\pm1.63}$ & +7.73 & 78.19$_{\pm1.59}$ & 66.83$_{\pm1.47}$ & +11.37 & 87.72$_{\pm0.63}$ & 84.51$_{\pm0.47}$ & +3.21 & 77.68$_{\pm0.26}$ & 60.53$_{\pm0.25}$ & +17.16 \\
GIT & 87.56$_{\pm1.39}$ & 81.31$_{\pm1.31}$ & \underline{+6.24} & \textbf{81.47}$_{\pm0.65}$ & 68.40$_{\pm1.20}$ & +13.07 & 89.62$_{\pm0.31}$ & 86.84$_{\pm0.17}$ & +2.78 & 78.79$_{\pm0.24}$ & 62.47$_{\pm0.13}$ & +16.32 \\
GFT & 86.92$_{\pm1.76}$ & \underline{81.90}$_{\pm1.33}$ & \textbf{+5.03} & 80.60$_{\pm1.69}$ & 68.32$_{\pm1.25}$ & +12.28 & \underline{90.26}$_{\pm0.74}$ & \textbf{88.82}$_{\pm0.48}$ & \textbf{+1.44} & 75.91$_{\pm0.38}$ & 58.01$_{\pm0.44}$ & +17.91 \\
OFA & 87.08$_{\pm2.81}$ & 77.93$_{\pm1.32}$ & +9.15 & 78.18$_{\pm2.13}$ & 60.88$_{\pm2.06}$ & +17.30 & 85.29$_{\pm0.93}$ & 81.06$_{\pm1.09}$ & +4.24 & 74.06$_{\pm0.51}$ & 59.70$_{\pm0.23}$ & +14.36 \\
LLaGA & 84.35$_{\pm3.80}$ & 76.57$_{\pm1.40}$ & +7.78 & 78.37$_{\pm1.85}$ & 60.47$_{\pm2.36}$ & +17.90 & 86.86$_{\pm1.74}$ & 81.66$_{\pm0.73}$ & +5.20 & \multicolumn{3}{c}{--} \\
\bottomrule
\end{tabular}%
}
\par\vspace{4pt}
\centerline{%
\begin{tabular}{l|ccc|ccc|ccc|ccc}
\toprule
\textbf{Method} & \multicolumn{3}{c}{\textbf{arxiv23}} & \multicolumn{3}{c}{\textbf{dblp}} & \multicolumn{3}{c}{\textbf{wikics}} & \multicolumn{3}{c}{\textbf{elecomp}} \\
\cmidrule(lr){2-4} \cmidrule(lr){5-7} \cmidrule(lr){8-10} \cmidrule(lr){11-13}
 & ID & OOD & Gap & ID & OOD & Gap & ID & OOD & Gap & ID & OOD & Gap \\
\midrule
DeepWalk & 32.83$_{\pm0.84}$ & 19.08$_{\pm0.45}$ & +13.74 & 76.08$_{\pm0.64}$ & 44.45$_{\pm1.11}$ & +31.63 & 82.41$_{\pm1.25}$ & 61.64$_{\pm0.55}$ & +20.77 & 78.86$_{\pm0.42}$ & 62.95$_{\pm0.15}$ & +15.91 \\
Node2Vec & 32.61$_{\pm0.94}$ & 19.01$_{\pm0.15}$ & +13.60 & 76.23$_{\pm0.54}$ & 43.93$_{\pm1.15}$ & +32.30 & 83.16$_{\pm0.82}$ & 61.63$_{\pm1.62}$ & +21.53 & 79.01$_{\pm0.47}$ & 62.72$_{\pm0.14}$ & +16.29 \\
\cmidrule(lr){2-13}
GCN & 71.17$_{\pm0.56}$ & 66.88$_{\pm0.21}$ & \underline{+4.29} & 79.66$_{\pm1.13}$ & 75.24$_{\pm1.81}$ & \textbf{+4.42} & 85.74$_{\pm0.86}$ & 68.41$_{\pm3.01}$ & +17.33 & 86.91$_{\pm0.40}$ & 68.37$_{\pm0.42}$ & +18.54 \\
GAT & 71.05$_{\pm0.36}$ & 66.50$_{\pm0.54}$ & +4.55 & 79.64$_{\pm0.82}$ & 71.77$_{\pm3.33}$ & +7.88 & 85.99$_{\pm0.63}$ & 72.24$_{\pm1.77}$ & +13.75 & 86.66$_{\pm0.29}$ & 67.19$_{\pm0.24}$ & +19.47 \\
SAGE & \underline{72.86}$_{\pm0.48}$ & 67.75$_{\pm0.22}$ & +5.11 & 81.76$_{\pm0.85}$ & \textbf{76.54}$_{\pm1.70}$ & +5.22 & 86.49$_{\pm0.47}$ & \underline{76.15}$_{\pm1.00}$ & \underline{+10.34} & 87.35$_{\pm0.35}$ & 68.82$_{\pm0.35}$ & +18.53 \\
\cmidrule(lr){2-13}
BGRL & 72.57$_{\pm0.54}$ & \textbf{68.76}$_{\pm0.25}$ & \textbf{+3.81} & 79.50$_{\pm1.12}$ & 73.25$_{\pm4.08}$ & +6.25 & \textbf{87.10}$_{\pm0.82}$ & 65.79$_{\pm3.48}$ & +21.31 & 87.34$_{\pm0.43}$ & 68.84$_{\pm0.56}$ & +18.51 \\
GraphMAE & \textbf{72.96}$_{\pm0.36}$ & \underline{68.13}$_{\pm0.18}$ & +4.83 & 79.99$_{\pm1.05}$ & 75.54$_{\pm0.68}$ & \underline{+4.46} & 85.64$_{\pm0.70}$ & \textbf{76.51}$_{\pm0.84}$ & \textbf{+9.13} & \textbf{88.34}$_{\pm0.33}$ & \textbf{70.12}$_{\pm0.18}$ & +18.22 \\
UniGraph2 & 67.12$_{\pm0.69}$ & 60.45$_{\pm0.49}$ & +6.67 & 79.22$_{\pm1.07}$ & 74.52$_{\pm0.89}$ & +4.70 & 83.35$_{\pm0.61}$ & 71.80$_{\pm0.89}$ & +11.55 & 85.68$_{\pm0.32}$ & 64.38$_{\pm0.62}$ & +21.29 \\
GIT & 71.73$_{\pm0.50}$ & 66.31$_{\pm0.14}$ & +5.42 & \underline{81.78}$_{\pm1.11}$ & \underline{76.24}$_{\pm1.19}$ & +5.54 & 85.87$_{\pm0.49}$ & 73.84$_{\pm0.83}$ & +12.03 & \underline{87.46}$_{\pm0.40}$ & \underline{69.15}$_{\pm0.38}$ & +18.30 \\
GFT & 70.56$_{\pm0.65}$ & 64.47$_{\pm0.44}$ & +6.09 & \textbf{81.81}$_{\pm0.81}$ & 74.43$_{\pm0.62}$ & +7.38 & \underline{86.75}$_{\pm0.84}$ & 72.81$_{\pm2.02}$ & +13.94 & 85.94$_{\pm0.18}$ & 67.92$_{\pm0.55}$ & \underline{+18.02} \\
OFA & 65.44$_{\pm0.70}$ & 55.97$_{\pm0.43}$ & +9.47 & 79.03$_{\pm0.77}$ & 70.55$_{\pm2.16}$ & +8.48 & 83.42$_{\pm1.23}$ & 68.65$_{\pm0.61}$ & +14.77 & 87.02$_{\pm0.66}$ & 74.51$_{\pm0.40}$ & +12.51 \\
LLaGA & 69.90$_{\pm2.89}$ & 65.13$_{\pm1.18}$ & +4.77 & 76.92$_{\pm0.61}$ & 65.50$_{\pm1.88}$ & +11.42 & 82.54$_{\pm1.52}$ & 67.68$_{\pm1.64}$ & +14.86 & \multicolumn{3}{c}{--} \\
\bottomrule
\end{tabular}%
}
\par\vspace{4pt}
\centerline{%
\begin{tabular}{l|ccc|ccc|ccc|ccc}
\toprule
\textbf{Method} & \multicolumn{3}{c}{\textbf{elephoto}} & \multicolumn{3}{c}{\textbf{sportsfit}} & \multicolumn{3}{c}{\textbf{amazonratings}} & \multicolumn{3}{c}{\textbf{bookchild}} \\
\cmidrule(lr){2-4} \cmidrule(lr){5-7} \cmidrule(lr){8-10} \cmidrule(lr){11-13}
 & ID & OOD & Gap & ID & OOD & Gap & ID & OOD & Gap & ID & OOD & Gap \\
\midrule
DeepWalk & 74.93$_{\pm0.71}$ & 60.20$_{\pm0.23}$ & +14.72 & 88.84$_{\pm0.17}$ & 63.61$_{\pm0.23}$ & +25.23 & 38.39$_{\pm0.50}$ & 37.29$_{\pm0.28}$ & +1.10 & 47.97$_{\pm0.40}$ & 32.20$_{\pm0.44}$ & +15.77 \\
Node2Vec & 75.14$_{\pm0.37}$ & 60.71$_{\pm0.30}$ & +14.43 & 88.97$_{\pm0.18}$ & 63.69$_{\pm0.13}$ & +25.28 & 38.31$_{\pm0.63}$ & 36.99$_{\pm0.46}$ & +1.32 & 47.96$_{\pm0.28}$ & 32.26$_{\pm0.47}$ & +15.70 \\
\cmidrule(lr){2-13}
GCN & 84.03$_{\pm0.34}$ & 70.28$_{\pm0.76}$ & +13.76 & 92.33$_{\pm0.20}$ & 77.77$_{\pm0.22}$ & \underline{+14.56} & \textbf{43.11}$_{\pm0.85}$ & 43.59$_{\pm0.59}$ & -0.47 & 57.33$_{\pm0.39}$ & 43.41$_{\pm0.46}$ & +13.92 \\
GAT & 83.87$_{\pm0.51}$ & 68.29$_{\pm0.94}$ & +15.59 & 91.54$_{\pm0.16}$ & 75.17$_{\pm0.12}$ & +16.37 & 40.21$_{\pm0.90}$ & 41.08$_{\pm1.29}$ & -0.87 & 57.75$_{\pm0.34}$ & 43.59$_{\pm0.22}$ & +14.16 \\
SAGE & 84.76$_{\pm0.46}$ & \underline{71.50}$_{\pm0.76}$ & \underline{+13.26} & 92.91$_{\pm0.14}$ & 77.47$_{\pm0.20}$ & +15.44 & 41.27$_{\pm1.13}$ & \underline{44.15}$_{\pm2.97}$ & -2.89 & \textbf{60.85}$_{\pm0.33}$ & 45.47$_{\pm0.53}$ & +15.38 \\
\cmidrule(lr){2-13}
BGRL & \underline{84.84}$_{\pm0.23}$ & 71.49$_{\pm0.53}$ & +13.35 & \underline{92.95}$_{\pm0.15}$ & \underline{78.25}$_{\pm0.24}$ & +14.70 & 42.59$_{\pm0.96}$ & 43.34$_{\pm0.68}$ & -0.75 & 58.66$_{\pm0.41}$ & \underline{45.49}$_{\pm0.76}$ & \underline{+13.17} \\
GraphMAE & \textbf{85.72}$_{\pm0.24}$ & \textbf{71.88}$_{\pm0.28}$ & +13.84 & \textbf{93.48}$_{\pm0.15}$ & \textbf{80.51}$_{\pm0.21}$ & \textbf{+12.97} & \underline{42.74}$_{\pm0.72}$ & 43.70$_{\pm0.87}$ & -0.96 & 59.52$_{\pm0.32}$ & 45.21$_{\pm0.32}$ & +14.31 \\
UniGraph2 & 82.96$_{\pm0.47}$ & 67.86$_{\pm0.78}$ & +15.10 & 91.63$_{\pm0.21}$ & 74.27$_{\pm0.31}$ & +17.36 & 41.54$_{\pm0.71}$ & 41.98$_{\pm0.87}$ & \underline{-0.43} & 56.25$_{\pm0.45}$ & 43.28$_{\pm0.42}$ & \textbf{+12.97} \\
GIT & 84.58$_{\pm0.57}$ & 70.89$_{\pm0.80}$ & +13.68 & 92.91$_{\pm0.15}$ & 77.12$_{\pm0.18}$ & +15.80 & 40.14$_{\pm0.91}$ & \textbf{45.24}$_{\pm1.32}$ & -5.10 & \underline{60.63}$_{\pm0.18}$ & \textbf{45.86}$_{\pm0.37}$ & +14.76 \\
GFT & 83.32$_{\pm0.45}$ & 68.26$_{\pm0.65}$ & +15.05 & 92.35$_{\pm0.21}$ & 75.68$_{\pm0.40}$ & +16.66 & 40.86$_{\pm1.00}$ & 41.82$_{\pm1.58}$ & -0.96 & 55.23$_{\pm0.47}$ & 40.19$_{\pm0.54}$ & +15.04 \\
OFA & 84.18$_{\pm0.48}$ & 72.08$_{\pm0.58}$ & +12.11 & 87.14$_{\pm0.34}$ & 68.65$_{\pm0.64}$ & +18.49 & 37.10$_{\pm0.64}$ & 36.16$_{\pm1.01}$ & +0.94 & 41.95$_{\pm0.87}$ & 32.46$_{\pm1.31}$ & +9.49 \\
LLaGA & \multicolumn{3}{c}{--} & \multicolumn{3}{c}{--} & 39.60$_{\pm1.95}$ & 39.93$_{\pm2.27}$ & -0.33 & \multicolumn{3}{c}{--} \\
\bottomrule
\end{tabular}%
}
\par\vspace{4pt}
\centerline{%
\begin{tabular}{l|ccc}
\toprule
\textbf{Method} & \multicolumn{3}{c}{\textbf{bookhis}} \\
\cmidrule(lr){2-4}
 & ID & OOD & Gap \\
\midrule
DeepWalk & 79.18$_{\pm0.59}$ & 69.43$_{\pm0.25}$ & +9.75 \\
Node2Vec & 78.96$_{\pm0.52}$ & 69.10$_{\pm0.50}$ & +9.85 \\
\cmidrule(lr){2-4}
GCN & 83.61$_{\pm0.51}$ & 81.51$_{\pm0.57}$ & +2.09 \\
GAT & 83.67$_{\pm0.34}$ & 80.60$_{\pm0.98}$ & +3.07 \\
SAGE & \textbf{84.79}$_{\pm0.45}$ & \textbf{82.64}$_{\pm0.24}$ & +2.15 \\
\cmidrule(lr){2-4}
BGRL & 84.17$_{\pm0.66}$ & 82.26$_{\pm0.38}$ & \textbf{+1.91} \\
GraphMAE & \underline{84.40}$_{\pm0.59}$ & \underline{82.31}$_{\pm0.24}$ & +2.08 \\
UniGraph2 & 83.04$_{\pm0.63}$ & 80.10$_{\pm0.43}$ & +2.94 \\
GIT & 84.29$_{\pm0.43}$ & 82.25$_{\pm0.31}$ & \underline{+2.04} \\
GFT & 83.33$_{\pm0.49}$ & 80.48$_{\pm0.46}$ & +2.85 \\
OFA & 79.91$_{\pm0.83}$ & 75.89$_{\pm0.76}$ & +4.02 \\
LLaGA & 82.40$_{\pm1.95}$ & 80.26$_{\pm0.40}$ & +2.14 \\
\bottomrule
\end{tabular}%
}
\end{table}

\FloatBarrier

\clearpage
\subsection{Class imbalance --- full results}
\label{app:imb_full}



\begin{table}[H]
\centering
\caption{Per-class recall on 8 NC datasets (\%): \emph{major / minor} at imbalance ratios $\rho \in \{5,10,20\}$, mean$\pm$std over 5 seeds. Major and minor groups are ranked independently; \textbf{bold} / \underline{underline} mark the best and second-best methods in each column. Cells marked `--' indicate the method collapsed to majority-class predictions and minor recall is uninformative.}
\label{tab:app_imb_per_class}
\scriptsize
\setlength{\tabcolsep}{2pt}
\renewcommand{\arraystretch}{0.85}
\centerline{%
\begin{tabular}{l|ccc|ccc}
\toprule
\textbf{Method} & \multicolumn{3}{c}{\textbf{cora}} & \multicolumn{3}{c}{\textbf{citeseer}} \\
\cmidrule(lr){2-4} \cmidrule(lr){5-7}
 & $\rho{=}5$ & $\rho{=}10$ & $\rho{=}20$ & $\rho{=}5$ & $\rho{=}10$ & $\rho{=}20$ \\
\midrule
DeepWalk & 84.0$_{\pm0.6}$/24.4$_{\pm4.2}$ & 84.9$_{\pm1.0}$/7.1$_{\pm2.6}$ & 85.0$_{\pm0.9}$/1.9$_{\pm1.7}$ & 53.2$_{\pm1.1}$/9.9$_{\pm3.5}$ & 53.7$_{\pm1.5}$/1.6$_{\pm0.7}$ & 54.1$_{\pm1.3}$/0.2$_{\pm0.1}$ \\
Node2vec &84.7$_{\pm0.6}$/23.4$_{\pm4.1}$ & 85.7$_{\pm0.6}$/7.0$_{\pm2.1}$ & 85.8$_{\pm0.8}$/1.9$_{\pm1.5}$ & 53.5$_{\pm1.9}$/9.9$_{\pm3.4}$ & 54.4$_{\pm2.1}$/1.6$_{\pm0.7}$ & 54.3$_{\pm2.0}$/0.3$_{\pm0.2}$ \\
\midrule
GCN & 85.8$_{\pm2.0}$/58.3$_{\pm4.4}$ & 87.9$_{\pm1.3}$/41.3$_{\pm8.2}$ & 89.9$_{\pm1.0}$/24.7$_{\pm10.2}$ & 73.7$_{\pm1.8}$/41.1$_{\pm4.9}$ & 75.4$_{\pm2.3}$/28.3$_{\pm8.6}$ & 76.9$_{\pm1.2}$/17.5$_{\pm6.6}$ \\
GAT & 83.8$_{\pm1.6}$/\underline{62.3}$_{\pm5.0}$ & 85.1$_{\pm2.9}$/\underline{47.4}$_{\pm9.4}$ & 88.2$_{\pm2.0}$/\underline{27.6}$_{\pm12.6}$ & 72.5$_{\pm2.1}$/\underline{45.0}$_{\pm5.8}$ & 73.5$_{\pm2.0}$/\textbf{34.2}$_{\pm8.2}$ & 76.3$_{\pm1.8}$/18.2$_{\pm9.1}$ \\
GraphSAGE & 86.8$_{\pm1.3}$/49.4$_{\pm4.5}$ & 86.7$_{\pm1.2}$/38.3$_{\pm5.3}$ & 87.9$_{\pm0.8}$/24.2$_{\pm6.6}$ & 74.6$_{\pm1.1}$/38.9$_{\pm3.0}$ & 73.9$_{\pm1.5}$/\underline{33.5}$_{\pm4.4}$ & 75.2$_{\pm1.3}$/\underline{23.8}$_{\pm5.3}$ \\
\midrule
BGRL & \underline{88.2}$_{\pm0.9}$/50.5$_{\pm4.2}$ & \underline{89.8}$_{\pm0.8}$/35.4$_{\pm4.9}$ & \underline{91.0}$_{\pm0.8}$/20.6$_{\pm9.0}$ & 73.9$_{\pm1.7}$/36.4$_{\pm3.7}$ & 75.8$_{\pm1.0}$/23.4$_{\pm6.4}$ & \underline{77.6}$_{\pm1.0}$/11.4$_{\pm4.3}$ \\
GraphMAE & \textbf{88.4}$_{\pm1.2}$/56.8$_{\pm6.0}$ & \textbf{90.3}$_{\pm0.7}$/40.7$_{\pm7.8}$ & \textbf{91.0}$_{\pm0.8}$/19.7$_{\pm11.9}$ & \textbf{77.2}$_{\pm1.4}$/36.4$_{\pm4.2}$ & \textbf{79.0}$_{\pm1.3}$/21.8$_{\pm7.6}$ & \textbf{80.6}$_{\pm0.9}$/9.2$_{\pm4.9}$ \\
\midrule
UniGraph2 & 77.8$_{\pm4.4}$/55.5$_{\pm7.5}$ & 76.9$_{\pm7.0}$/42.3$_{\pm5.5}$ & 81.2$_{\pm3.3}$/23.1$_{\pm9.2}$ & 63.9$_{\pm5.8}$/\textbf{45.6}$_{\pm10.8}$ & 69.0$_{\pm4.1}$/32.1$_{\pm5.7}$ & 68.3$_{\pm4.3}$/\textbf{24.9}$_{\pm5.0}$ \\
GIT & 87.8$_{\pm1.2}$/44.3$_{\pm5.5}$ & 88.8$_{\pm1.5}$/29.4$_{\pm5.2}$ & 89.1$_{\pm0.5}$/17.6$_{\pm5.1}$ & \underline{75.3}$_{\pm1.0}$/36.3$_{\pm2.9}$ & 75.3$_{\pm1.4}$/29.4$_{\pm5.9}$ & 77.1$_{\pm1.0}$/19.1$_{\pm5.2}$ \\
GFT & 84.6$_{\pm1.6}$/53.6$_{\pm5.5}$ & 87.6$_{\pm1.6}$/36.1$_{\pm6.6}$ & 88.1$_{\pm1.8}$/20.7$_{\pm6.2}$ & 73.6$_{\pm2.1}$/39.7$_{\pm5.5}$ & 75.6$_{\pm1.8}$/28.9$_{\pm5.3}$ & 77.1$_{\pm1.5}$/18.4$_{\pm3.9}$ \\
\midrule
OFA & 83.0$_{\pm0.2}$/47.3$_{\pm8.2}$ & 84.2$_{\pm0.8}$/34.7$_{\pm8.0}$ & 85.5$_{\pm0.5}$/18.0$_{\pm6.6}$ & 68.1$_{\pm8.2}$/36.7$_{\pm8.0}$ & 69.3$_{\pm3.1}$/25.5$_{\pm4.1}$ & 73.7$_{\pm2.5}$/10.5$_{\pm7.5}$ \\
LLaGA & 83.6$_{\pm2.0}$/\textbf{63.9}$_{\pm7.9}$ & 85.4$_{\pm1.1}$/\textbf{54.1}$_{\pm10.0}$ & 87.8$_{\pm1.3}$/\textbf{37.1}$_{\pm8.0}$ & 74.0$_{\pm2.8}$/29.2$_{\pm8.2}$ & \underline{77.4}$_{\pm2.1}$/10.8$_{\pm9.5}$ & 75.3$_{\pm3.2}$/8.3$_{\pm5.7}$ \\
\bottomrule
\end{tabular}%
}
\par\vspace{6pt}
\centerline{%
\begin{tabular}{l|ccc|ccc}
\toprule
\textbf{Method} & \multicolumn{3}{c}{\textbf{pubmed}} & \multicolumn{3}{c}{\textbf{dblp}} \\
\cmidrule(lr){2-4} \cmidrule(lr){5-7}
 & $\rho{=}5$ & $\rho{=}10$ & $\rho{=}20$ & $\rho{=}5$ & $\rho{=}10$ & $\rho{=}20$ \\
\midrule
DeepWalk & -- & -- & -- & 65.6$_{\pm1.0}$/3.8$_{\pm1.5}$ & 66.0$_{\pm1.1}$/0.6$_{\pm0.2}$ & 66.0$_{\pm1.0}$/0.0$_{\pm0.1}$ \\
Node2vec &-- & -- & -- & 65.3$_{\pm1.3}$/4.1$_{\pm2.3}$ & 65.7$_{\pm1.4}$/0.7$_{\pm0.3}$ & 65.7$_{\pm1.2}$/0.1$_{\pm0.1}$ \\
\midrule
GCN & 91.4$_{\pm4.4}$/50.2$_{\pm7.5}$ & \underline{96.8}$_{\pm2.2}$/32.1$_{\pm7.5}$ & \underline{99.2}$_{\pm0.5}$/12.4$_{\pm9.7}$ & 83.8$_{\pm1.5}$/20.0$_{\pm6.0}$ & 85.5$_{\pm2.3}$/8.3$_{\pm7.8}$ & 86.4$_{\pm2.2}$/4.8$_{\pm6.8}$ \\
GAT & 86.0$_{\pm9.1}$/56.5$_{\pm6.8}$ & 93.2$_{\pm7.0}$/39.6$_{\pm13.4}$ & 97.5$_{\pm3.4}$/19.6$_{\pm16.6}$ & \underline{84.1}$_{\pm1.7}$/24.9$_{\pm8.5}$ & 84.7$_{\pm3.4}$/12.5$_{\pm11.5}$ & 86.5$_{\pm1.9}$/5.9$_{\pm9.0}$ \\
GraphSAGE & 92.0$_{\pm4.0}$/44.7$_{\pm7.3}$ & 96.7$_{\pm1.9}$/31.1$_{\pm9.5}$ & \textbf{99.4}$_{\pm0.6}$/10.1$_{\pm5.7}$ & 83.4$_{\pm1.7}$/22.4$_{\pm5.8}$ & 84.1$_{\pm1.8}$/14.2$_{\pm5.5}$ & 84.1$_{\pm1.4}$/10.9$_{\pm4.5}$ \\
\midrule
BGRL & 89.0$_{\pm4.3}$/50.6$_{\pm5.4}$ & 93.2$_{\pm3.1}$/42.0$_{\pm8.5}$ & 94.2$_{\pm3.9}$/\underline{33.0}$_{\pm9.4}$ & 83.3$_{\pm1.9}$/27.4$_{\pm6.5}$ & 85.2$_{\pm1.7}$/\underline{17.1}$_{\pm9.2}$ & 83.9$_{\pm6.1}$/\underline{11.6}$_{\pm8.7}$ \\
GraphMAE & \textbf{93.4}$_{\pm3.4}$/39.3$_{\pm9.2}$ & \textbf{98.3}$_{\pm1.4}$/21.1$_{\pm7.0}$ & 98.9$_{\pm2.7}$/8.3$_{\pm6.8}$ & \textbf{84.7}$_{\pm1.4}$/19.6$_{\pm7.2}$ & \textbf{86.5}$_{\pm1.4}$/7.7$_{\pm8.3}$ & \textbf{87.2}$_{\pm1.2}$/3.1$_{\pm5.9}$ \\
\midrule
UniGraph2 & 79.3$_{\pm6.3}$/\underline{59.3}$_{\pm3.6}$ & 84.2$_{\pm4.3}$/\underline{52.9}$_{\pm5.1}$ & 80.0$_{\pm7.6}$/\textbf{49.5}$_{\pm8.2}$ & 67.7$_{\pm11.3}$/\textbf{40.0}$_{\pm13.6}$ & 71.3$_{\pm5.4}$/\textbf{30.4}$_{\pm13.3}$ & 76.1$_{\pm6.7}$/\textbf{14.9}$_{\pm11.5}$ \\
GIT & \underline{92.8}$_{\pm3.4}$/43.8$_{\pm7.3}$ & 96.6$_{\pm2.2}$/30.7$_{\pm10.4}$ & 99.1$_{\pm0.8}$/12.3$_{\pm7.4}$ & 84.0$_{\pm1.8}$/17.6$_{\pm5.9}$ & 85.3$_{\pm1.7}$/8.7$_{\pm4.6}$ & 86.6$_{\pm0.7}$/3.6$_{\pm2.4}$ \\
GFT & 90.0$_{\pm5.2}$/53.4$_{\pm8.6}$ & 96.7$_{\pm1.8}$/36.2$_{\pm8.3}$ & 99.1$_{\pm0.4}$/16.5$_{\pm9.9}$ & 81.4$_{\pm2.0}$/31.1$_{\pm8.0}$ & 82.9$_{\pm2.2}$/15.2$_{\pm6.4}$ & 84.6$_{\pm1.1}$/8.3$_{\pm3.5}$ \\
\midrule
OFA & 73.3$_{\pm1.6}$/\textbf{66.2}$_{\pm7.7}$ & 72.9$_{\pm2.2}$/\textbf{57.6}$_{\pm8.5}$ & 75.4$_{\pm2.2}$/29.6$_{\pm21.9}$ & 69.7$_{\pm7.0}$/\underline{31.2}$_{\pm6.9}$ & 79.6$_{\pm5.3}$/12.5$_{\pm8.5}$ & 82.3$_{\pm3.5}$/3.6$_{\pm7.9}$ \\
LLaGA & 86.7$_{\pm12.9}$/50.5$_{\pm11.5}$ & 94.3$_{\pm4.7}$/34.3$_{\pm13.5}$ & 98.5$_{\pm1.6}$/21.4$_{\pm9.9}$ & 80.9$_{\pm1.0}$/22.3$_{\pm9.0}$ & \underline{85.8}$_{\pm2.5}$/4.1$_{\pm7.1}$ & \underline{86.7}$_{\pm1.6}$/0.0$_{\pm0.0}$ \\
\bottomrule
\end{tabular}%
}
\par\vspace{6pt}
\centerline{%
\begin{tabular}{l|ccc|ccc}
\toprule
\textbf{Method} & \multicolumn{3}{c}{\textbf{elecomp}} & \multicolumn{3}{c}{\textbf{elephoto}} \\
\cmidrule(lr){2-4} \cmidrule(lr){5-7}
 & $\rho{=}5$ & $\rho{=}10$ & $\rho{=}20$ & $\rho{=}5$ & $\rho{=}10$ & $\rho{=}20$ \\
\midrule
DeepWalk & 73.1$_{\pm0.5}$/54.0$_{\pm3.5}$ & 73.9$_{\pm0.5}$/49.0$_{\pm3.1}$ & 74.3$_{\pm0.5}$/44.3$_{\pm2.8}$ & 65.4$_{\pm0.2}$/55.1$_{\pm1.1}$ & 65.5$_{\pm0.3}$/52.9$_{\pm1.4}$ & 65.9$_{\pm0.4}$/46.0$_{\pm0.7}$ \\
Node2vec &73.1$_{\pm0.4}$/52.2$_{\pm2.1}$ & 73.9$_{\pm0.4}$/48.6$_{\pm1.6}$ & 74.2$_{\pm0.4}$/43.3$_{\pm1.9}$ & 65.2$_{\pm0.4}$/55.8$_{\pm1.5}$ & 65.4$_{\pm0.3}$/54.1$_{\pm1.5}$ & 65.7$_{\pm0.4}$/47.4$_{\pm1.5}$ \\
\midrule
GCN & 77.6$_{\pm0.3}$/55.0$_{\pm1.3}$ & 78.0$_{\pm0.2}$/51.1$_{\pm0.9}$ & 78.2$_{\pm0.3}$/46.8$_{\pm0.8}$ & 77.0$_{\pm0.5}$/57.7$_{\pm1.3}$ & 76.8$_{\pm0.4}$/56.7$_{\pm0.6}$ & 77.4$_{\pm0.5}$/\underline{54.0}$_{\pm1.2}$ \\
GAT & 76.3$_{\pm0.1}$/54.6$_{\pm0.3}$ & 76.9$_{\pm0.2}$/51.8$_{\pm1.0}$ & 77.2$_{\pm0.1}$/48.3$_{\pm0.5}$ & 76.1$_{\pm0.5}$/55.3$_{\pm0.9}$ & 76.2$_{\pm0.4}$/54.0$_{\pm1.0}$ & 76.8$_{\pm0.5}$/50.3$_{\pm1.4}$ \\
GraphSAGE & 76.7$_{\pm0.4}$/54.3$_{\pm1.1}$ & 77.4$_{\pm0.2}$/51.5$_{\pm0.7}$ & 77.6$_{\pm0.5}$/47.7$_{\pm0.9}$ & 76.3$_{\pm0.7}$/56.9$_{\pm1.2}$ & 76.0$_{\pm0.5}$/55.2$_{\pm1.0}$ & 76.7$_{\pm0.4}$/50.9$_{\pm1.1}$ \\
\midrule
BGRL & 75.4$_{\pm0.4}$/56.0$_{\pm1.6}$ & 75.9$_{\pm0.6}$/54.8$_{\pm1.5}$ & 76.3$_{\pm0.6}$/50.3$_{\pm2.3}$ & 75.8$_{\pm0.5}$/56.1$_{\pm1.5}$ & 76.0$_{\pm0.3}$/53.9$_{\pm1.3}$ & 76.7$_{\pm0.6}$/50.6$_{\pm1.4}$ \\
GraphMAE & \underline{78.8}$_{\pm0.2}$/\underline{58.9}$_{\pm0.8}$ & \underline{79.1}$_{\pm0.4}$/\underline{56.6}$_{\pm1.1}$ & \underline{79.5}$_{\pm0.3}$/\underline{53.4}$_{\pm0.9}$ & \underline{77.2}$_{\pm0.2}$/57.8$_{\pm0.4}$ & \underline{77.5}$_{\pm0.2}$/56.6$_{\pm0.5}$ & \underline{78.0}$_{\pm0.2}$/53.4$_{\pm0.5}$ \\
\midrule
UniGraph2 & 75.1$_{\pm0.7}$/\textbf{59.2}$_{\pm2.5}$ & 75.0$_{\pm1.1}$/\textbf{58.7}$_{\pm3.2}$ & 75.6$_{\pm1.4}$/\textbf{57.3}$_{\pm2.5}$ & 75.1$_{\pm1.3}$/\textbf{60.1}$_{\pm2.1}$ & 75.1$_{\pm1.3}$/\textbf{59.0}$_{\pm1.5}$ & 75.9$_{\pm1.3}$/\textbf{56.0}$_{\pm2.4}$ \\
GIT & 77.0$_{\pm0.3}$/54.2$_{\pm0.9}$ & 77.7$_{\pm0.4}$/51.7$_{\pm1.2}$ & 77.9$_{\pm0.4}$/48.5$_{\pm1.2}$ & 76.4$_{\pm0.7}$/57.7$_{\pm1.0}$ & 76.5$_{\pm0.9}$/56.4$_{\pm0.9}$ & 77.0$_{\pm0.8}$/50.9$_{\pm0.9}$ \\
GFT & 77.4$_{\pm0.3}$/56.9$_{\pm1.5}$ & 77.8$_{\pm0.2}$/54.6$_{\pm1.4}$ & 78.0$_{\pm0.3}$/51.5$_{\pm1.5}$ & \textbf{77.8}$_{\pm0.6}$/\underline{59.3}$_{\pm0.9}$ & \textbf{77.9}$_{\pm0.4}$/\underline{58.5}$_{\pm1.6}$ & \textbf{78.1}$_{\pm0.5}$/53.4$_{\pm1.3}$ \\
\midrule
OFA & 77.0$_{\pm0.1}$/36.3$_{\pm0.5}$ & 77.5$_{\pm0.1}$/29.2$_{\pm0.7}$ & 77.5$_{\pm0.1}$/20.5$_{\pm1.0}$ & 75.1$_{\pm0.1}$/28.3$_{\pm0.2}$ & 75.2$_{\pm0.4}$/18.4$_{\pm1.3}$ & 75.2$_{\pm0.3}$/8.3$_{\pm0.8}$ \\
LLaGA & \textbf{81.1}$_{\pm2.8}$/46.9$_{\pm4.5}$ & \textbf{81.1}$_{\pm2.8}$/41.2$_{\pm8.4}$ & \textbf{81.8}$_{\pm2.2}$/41.8$_{\pm8.3}$ & 75.4$_{\pm6.6}$/55.2$_{\pm8.5}$ & 75.3$_{\pm6.5}$/56.3$_{\pm12.2}$ & 74.5$_{\pm6.9}$/46.6$_{\pm7.7}$ \\
\bottomrule
\end{tabular}%
}
\par\vspace{6pt}
\centerline{%
\begin{tabular}{l|ccc|ccc}
\toprule
\textbf{Method} & \multicolumn{3}{c}{\textbf{sportsfit}} & \multicolumn{3}{c}{\textbf{wikics}} \\
\cmidrule(lr){2-4} \cmidrule(lr){5-7}
 & $\rho{=}5$ & $\rho{=}10$ & $\rho{=}20$ & $\rho{=}5$ & $\rho{=}10$ & $\rho{=}20$ \\
\midrule
DeepWalk & 82.1$_{\pm0.2}$/45.0$_{\pm0.2}$ & 82.2$_{\pm0.2}$/44.9$_{\pm0.2}$ & 83.0$_{\pm0.2}$/42.0$_{\pm0.2}$ & 76.0$_{\pm0.3}$/58.0$_{\pm1.9}$ & 77.3$_{\pm0.5}$/44.0$_{\pm1.2}$ & 78.0$_{\pm0.5}$/20.9$_{\pm2.0}$ \\
Node2vec &82.1$_{\pm0.1}$/45.5$_{\pm0.3}$ & 82.2$_{\pm0.1}$/45.4$_{\pm0.3}$ & 83.0$_{\pm0.1}$/42.6$_{\pm0.3}$ & 75.3$_{\pm0.5}$/60.2$_{\pm1.8}$ & 76.6$_{\pm0.6}$/44.3$_{\pm3.8}$ & 77.3$_{\pm0.6}$/20.6$_{\pm2.4}$ \\
\midrule
GCN & \underline{89.7}$_{\pm0.2}$/72.8$_{\pm0.7}$ & \underline{89.6}$_{\pm0.2}$/73.3$_{\pm1.3}$ & \underline{90.1}$_{\pm0.2}$/70.1$_{\pm1.3}$ & 79.9$_{\pm0.3}$/69.7$_{\pm1.4}$ & 81.3$_{\pm0.8}$/62.2$_{\pm2.9}$ & 81.1$_{\pm0.7}$/49.3$_{\pm2.9}$ \\
GAT & 87.4$_{\pm0.2}$/67.8$_{\pm0.4}$ & 87.4$_{\pm0.2}$/67.9$_{\pm0.7}$ & 87.9$_{\pm0.2}$/65.0$_{\pm0.5}$ & 79.6$_{\pm0.4}$/\underline{72.3}$_{\pm1.3}$ & 80.3$_{\pm0.6}$/\underline{65.2}$_{\pm2.4}$ & 81.0$_{\pm0.7}$/\textbf{53.0}$_{\pm3.3}$ \\
GraphSAGE & 88.9$_{\pm0.1}$/72.8$_{\pm0.4}$ & 89.0$_{\pm0.1}$/72.4$_{\pm0.5}$ & 89.4$_{\pm0.1}$/69.5$_{\pm0.5}$ & 78.9$_{\pm0.3}$/70.2$_{\pm0.8}$ & 80.2$_{\pm0.4}$/63.3$_{\pm2.0}$ & 81.3$_{\pm0.3}$/48.0$_{\pm1.6}$ \\
\midrule
BGRL & 88.5$_{\pm0.5}$/\textbf{74.0}$_{\pm0.8}$ & 88.5$_{\pm0.5}$/\underline{73.9}$_{\pm0.6}$ & 89.1$_{\pm0.3}$/\underline{70.9}$_{\pm0.6}$ & \textbf{81.8}$_{\pm0.3}$/68.8$_{\pm1.3}$ & \textbf{82.9}$_{\pm0.3}$/59.1$_{\pm1.6}$ & \textbf{83.0}$_{\pm0.6}$/43.5$_{\pm3.4}$ \\
GraphMAE & \textbf{90.7}$_{\pm0.1}$/73.5$_{\pm0.4}$ & \textbf{90.7}$_{\pm0.1}$/73.4$_{\pm0.4}$ & \textbf{91.1}$_{\pm0.1}$/70.8$_{\pm0.5}$ & \underline{80.8}$_{\pm0.5}$/\textbf{73.6}$_{\pm1.0}$ & \underline{81.7}$_{\pm0.4}$/\textbf{67.9}$_{\pm1.5}$ & \underline{82.7}$_{\pm0.2}$/\underline{52.7}$_{\pm2.1}$ \\
\midrule
UniGraph2 & 84.5$_{\pm1.6}$/73.3$_{\pm2.1}$ & 84.8$_{\pm1.2}$/71.3$_{\pm3.0}$ & 85.7$_{\pm0.6}$/69.5$_{\pm1.9}$ & 74.0$_{\pm1.3}$/67.6$_{\pm2.2}$ & 76.3$_{\pm1.3}$/59.9$_{\pm3.8}$ & 76.1$_{\pm1.7}$/47.7$_{\pm2.7}$ \\
GIT & 89.3$_{\pm0.2}$/73.4$_{\pm0.4}$ & 89.3$_{\pm0.2}$/73.2$_{\pm0.3}$ & 89.6$_{\pm0.2}$/70.2$_{\pm0.7}$ & 78.8$_{\pm0.6}$/70.5$_{\pm1.1}$ & 79.9$_{\pm0.4}$/62.8$_{\pm1.5}$ & 81.3$_{\pm0.2}$/46.5$_{\pm2.1}$ \\
GFT & 89.6$_{\pm0.1}$/\underline{73.9}$_{\pm0.7}$ & 89.5$_{\pm0.2}$/\textbf{74.0}$_{\pm0.8}$ & 89.9$_{\pm0.3}$/\textbf{71.4}$_{\pm1.3}$ & 78.4$_{\pm0.5}$/71.5$_{\pm0.7}$ & 79.4$_{\pm0.7}$/64.2$_{\pm1.6}$ & 80.8$_{\pm0.2}$/49.9$_{\pm3.1}$ \\
\midrule
OFA & -- & -- & -- & 78.5$_{\pm0.6}$/49.3$_{\pm3.1}$ & 79.1$_{\pm0.3}$/36.4$_{\pm4.4}$ & 79.7$_{\pm0.3}$/20.2$_{\pm6.5}$ \\
LLaGA & 84.8$_{\pm2.6}$/41.9$_{\pm12.0}$ & 85.7$_{\pm2.4}$/43.7$_{\pm13.4}$ & 84.3$_{\pm1.8}$/35.0$_{\pm10.6}$ & 76.0$_{\pm1.8}$/62.7$_{\pm10.0}$ & 77.6$_{\pm2.2}$/51.5$_{\pm7.4}$ & 78.8$_{\pm1.7}$/40.8$_{\pm9.0}$ \\
\bottomrule
\end{tabular}%
}
\end{table}


\FloatBarrier

\subsection{Fairness --- full per-dataset results}
\label{app:fair_full}

The main-text fairness section reports structural fairness. Full
per-dataset structural-fairness results are reported below for
eight NC datasets, including the four representative datasets shown
in Figure~\ref{fig:fair_structural_v2}. Demographic fairness is
evaluated on tolokers (banned-user prediction; sensitive attribute:
education) and reported in
Table~\ref{tab:app_fair_demographic_full}.

\begin{table}[!htbp]
\centering
\caption{\textbf{Demographic fairness.} Test AUC and the
three fairness gaps for each method on tolokers (banned-user
prediction; sensitive attribute: education). Higher AUC is better;
lower $\Delta$ is more fair. All values in $\%$ ($\Delta$ values
$\times 100$). Mean$\pm$std over 5 seeds. \textbf{Bold} /
\underline{underline} = best / second per column.
$\Delta_{\text{SP}}{=}|P(\hat{Y}{=}1|S{=}0){-}P(\hat{Y}{=}1|S{=}1)|$,
$\Delta_{\text{EO}}{=}|\text{TPR}_{S=0}{-}\text{TPR}_{S=1}|$,
$\Delta_{\text{Util}}{=}|\text{AUC}_{S=0}{-}\text{AUC}_{S=1}|$.}
\label{tab:app_fair_demographic_full}
\label{tab:fair_demographic}
\footnotesize
\setlength{\tabcolsep}{6pt}
\renewcommand{\arraystretch}{1.05}
\begin{tabular}{lcccc}
\toprule
\textbf{Method} & \textbf{AUC}$\uparrow$ & $\boldsymbol{\Delta_{\text{SP}}}\!\downarrow$ & $\boldsymbol{\Delta_{\text{EO}}}\!\downarrow$ & $\boldsymbol{\Delta_{\text{Util}}}\!\downarrow$ \\
\midrule
GCN      & 74.78$_{\pm0.31}$ & 9.02$_{\pm0.73}$ & 11.80$_{\pm0.81}$ & 10.16$_{\pm0.41}$ \\
GAT      & 70.01$_{\pm2.27}$ & \underline{1.24}$_{\pm1.53}$ & 2.31$_{\pm1.76}$ & \textbf{7.78}$_{\pm1.69}$ \\
SAGE     & 75.59$_{\pm0.45}$ & 6.59$_{\pm2.52}$ & 9.21$_{\pm2.59}$ & 13.65$_{\pm0.34}$ \\
\midrule
BGRL     & \textbf{80.41}$_{\pm0.36}$ & 9.65$_{\pm1.74}$ & 13.33$_{\pm1.88}$ & \underline{8.30}$_{\pm0.80}$ \\
GraphMAE & 68.90$_{\pm3.46}$ & \textbf{1.12}$_{\pm1.48}$ & \textbf{1.89}$_{\pm2.44}$ & 11.73$_{\pm4.45}$ \\
\midrule
GFT      & 72.42$_{\pm0.61}$ & \underline{1.24}$_{\pm0.74}$ & \underline{2.03}$_{\pm1.15}$ & 11.43$_{\pm0.77}$ \\
GIT      & \underline{76.66}$_{\pm0.44}$ & 6.86$_{\pm1.88}$ & 9.77$_{\pm2.48}$ & 14.06$_{\pm0.31}$ \\
UniGraph2 & 69.34$_{\pm5.20}$ & 2.14$_{\pm3.40}$ & 3.53$_{\pm5.52}$ & 9.12$_{\pm4.01}$ \\
\bottomrule
\end{tabular}
\end{table}

\begin{table}[!htbp]
\centering
\caption{Full \textbf{structural-fairness} per-dataset results, one block per dataset.
Columns: overall accuracy, accuracy on head (high-degree) nodes, accuracy on tail
(low-degree) nodes, and head$-$tail gap. Within each dataset block, \textbf{bold} =
method with smallest $|\text{gap}|$; \underline{underline} = second.
Cells report mean$_{\pm\text{std}}$ over 5 seeds.}
\label{app:fair_structural}
\scriptsize
\setlength{\tabcolsep}{1.8pt}
\renewcommand{\arraystretch}{1.05}

\begin{minipage}[t]{0.49\linewidth}\centering
\textit{Cora}\\[1pt]
\begin{tabular}{lcccc}
\toprule
\textbf{Method} & \textbf{Acc} & \textbf{Acc(h)} & \textbf{Acc(t)} & \textbf{Gap}$\!\downarrow$ \\
\midrule
DeepWalk  & 71.47$_{\pm1.60}$ & 78.83$_{\pm0.72}$ & 63.12$_{\pm3.05}$ & +15.71$_{\pm3.28}$ \\
Node2vec  & 71.45$_{\pm1.39}$ & 79.14$_{\pm1.26}$ & 62.84$_{\pm2.21}$ & +16.30$_{\pm2.29}$ \\
\midrule
GCN       & 79.34$_{\pm0.66}$ & 81.73$_{\pm1.58}$ & 76.02$_{\pm0.62}$ & +5.71$_{\pm2.08}$ \\
GAT       & 79.20$_{\pm0.96}$ & 81.75$_{\pm1.24}$ & 76.06$_{\pm1.02}$ & +5.68$_{\pm0.97}$ \\
SAGE      & 78.60$_{\pm1.22}$ & 80.63$_{\pm2.45}$ & 75.42$_{\pm0.88}$ & +5.21$_{\pm2.16}$ \\
\midrule
BGRL      & 76.68$_{\pm1.04}$ & 80.14$_{\pm2.02}$ & 71.81$_{\pm1.36}$ & +8.34$_{\pm3.05}$ \\
GraphMAE  & 79.89$_{\pm1.49}$ & 82.33$_{\pm2.32}$ & 76.40$_{\pm1.51}$ & +5.93$_{\pm2.52}$ \\
\midrule
UniGraph2 & 75.93$_{\pm1.52}$ & 76.14$_{\pm2.33}$ & 72.92$_{\pm1.99}$ & \textbf{+3.22}$_{\pm3.25}$ \\
GIT       & 78.59$_{\pm1.13}$ & 80.36$_{\pm1.56}$ & 75.39$_{\pm0.91}$ & +4.97$_{\pm1.39}$ \\
GFT       & 79.16$_{\pm0.53}$ & 82.00$_{\pm1.37}$ & 74.83$_{\pm0.59}$ & +7.17$_{\pm1.80}$ \\
\midrule
OFA       & 75.17$_{\pm1.64}$ & 77.44$_{\pm2.87}$ & 73.70$_{\pm1.10}$ & \underline{+3.74}$_{\pm2.33}$ \\
LLaGA     & 76.31$_{\pm1.38}$ & 79.19$_{\pm2.58}$ & 72.35$_{\pm1.37}$ & +6.84$_{\pm3.02}$ \\
\bottomrule
\end{tabular}
\end{minipage}\hfill
\begin{minipage}[t]{0.49\linewidth}\centering
\textit{CiteSeer}\\[1pt]
\begin{tabular}{lcccc}
\toprule
\textbf{Method} & \textbf{Acc} & \textbf{Acc(h)} & \textbf{Acc(t)} & \textbf{Gap}$\!\downarrow$ \\
\midrule
DeepWalk  & 47.04$_{\pm0.66}$ & 68.79$_{\pm0.80}$ & 33.36$_{\pm0.71}$ & +35.43$_{\pm0.23}$ \\
Node2vec  & 46.56$_{\pm0.74}$ & 68.07$_{\pm0.66}$ & 33.30$_{\pm1.36}$ & +34.76$_{\pm0.96}$ \\
\midrule
GCN       & 68.51$_{\pm0.64}$ & 77.35$_{\pm1.26}$ & 62.74$_{\pm0.75}$ & +14.60$_{\pm1.48}$ \\
GAT       & 69.62$_{\pm0.58}$ & 77.54$_{\pm0.53}$ & 64.28$_{\pm0.61}$ & +13.26$_{\pm0.51}$ \\
SAGE      & 69.16$_{\pm0.40}$ & 77.35$_{\pm0.35}$ & 63.90$_{\pm0.51}$ & +13.45$_{\pm0.59}$ \\
\midrule
BGRL      & 66.31$_{\pm0.55}$ & 76.93$_{\pm0.84}$ & 58.95$_{\pm0.88}$ & +17.98$_{\pm0.79}$ \\
GraphMAE  & 67.92$_{\pm0.46}$ & 78.52$_{\pm0.49}$ & 61.23$_{\pm1.04}$ & +17.30$_{\pm0.97}$ \\
\midrule
UniGraph2 & 68.08$_{\pm1.10}$ & 75.64$_{\pm3.32}$ & 63.16$_{\pm1.30}$ & \underline{+12.49}$_{\pm3.88}$ \\
GIT       & 69.31$_{\pm0.54}$ & 77.12$_{\pm0.47}$ & 64.19$_{\pm0.54}$ & +12.93$_{\pm0.49}$ \\
GFT       & 66.77$_{\pm0.45}$ & 78.48$_{\pm0.89}$ & 59.57$_{\pm1.19}$ & +18.92$_{\pm1.98}$ \\
\midrule
OFA       & 63.83$_{\pm1.01}$ & 70.02$_{\pm1.43}$ & 60.79$_{\pm0.93}$ & \textbf{+9.23}$_{\pm0.99}$ \\
LLaGA     & 64.12$_{\pm1.04}$ & 74.37$_{\pm2.14}$ & 56.42$_{\pm2.05}$ & +17.95$_{\pm3.72}$ \\
\bottomrule
\end{tabular}
\end{minipage}

\vspace{6pt}

\begin{minipage}[t]{0.49\linewidth}\centering
\textit{PubMed}\\[1pt]
\begin{tabular}{lcccc}
\toprule
\textbf{Method} & \textbf{Acc} & \textbf{Acc(h)} & \textbf{Acc(t)} & \textbf{Gap}$\!\downarrow$ \\
\midrule
DeepWalk  & 67.73$_{\pm2.18}$ & 76.82$_{\pm1.73}$ & 61.12$_{\pm2.44}$ & +15.71$_{\pm1.23}$ \\
Node2vec  & 67.42$_{\pm2.60}$ & 76.47$_{\pm3.03}$ & 60.79$_{\pm2.56}$ & +15.68$_{\pm1.14}$ \\
\midrule
GCN       & 78.45$_{\pm1.75}$ & 79.49$_{\pm2.08}$ & 76.92$_{\pm1.64}$ & \underline{+2.56}$_{\pm0.73}$ \\
GAT       & 78.49$_{\pm0.87}$ & 80.51$_{\pm0.81}$ & 76.43$_{\pm1.05}$ & +4.08$_{\pm0.64}$ \\
SAGE      & 76.20$_{\pm0.99}$ & 78.46$_{\pm1.21}$ & 73.55$_{\pm1.00}$ & +4.91$_{\pm0.90}$ \\
\midrule
BGRL      & 73.10$_{\pm2.83}$ & 72.22$_{\pm4.11}$ & 72.51$_{\pm2.56}$ & \textbf{$-$0.28}$_{\pm2.07}$ \\
GraphMAE  & 77.92$_{\pm1.75}$ & 80.39$_{\pm2.23}$ & 75.54$_{\pm1.85}$ & +4.85$_{\pm0.92}$ \\
\midrule
UniGraph2 & 75.43$_{\pm1.37}$ & 77.66$_{\pm2.24}$ & 73.19$_{\pm1.38}$ & +4.47$_{\pm2.43}$ \\
GIT       & 76.29$_{\pm0.76}$ & 78.80$_{\pm0.97}$ & 73.57$_{\pm1.05}$ & +5.23$_{\pm1.31}$ \\
GFT       & 78.03$_{\pm0.56}$ & 80.55$_{\pm1.30}$ & 75.19$_{\pm0.70}$ & +5.36$_{\pm1.46}$ \\
\midrule
OFA       & 73.41$_{\pm2.56}$ & 75.99$_{\pm3.22}$ & 71.89$_{\pm2.20}$ & +4.10$_{\pm1.41}$ \\
LLaGA     & 73.56$_{\pm2.63}$ & 75.67$_{\pm3.64}$ & 70.94$_{\pm2.38}$ & +4.73$_{\pm1.47}$ \\
\bottomrule
\end{tabular}
\end{minipage}\hfill
\begin{minipage}[t]{0.49\linewidth}\centering
\textit{WikiCS}\\[1pt]
\begin{tabular}{lcccc}
\toprule
\textbf{Method} & \textbf{Acc} & \textbf{Acc(h)} & \textbf{Acc(t)} & \textbf{Gap}$\!\downarrow$ \\
\midrule
DeepWalk  & 73.49$_{\pm0.96}$ & 81.81$_{\pm1.17}$ & 61.72$_{\pm0.94}$ & +20.09$_{\pm1.00}$ \\
Node2vec  & 73.31$_{\pm0.54}$ & 82.25$_{\pm1.09}$ & 61.36$_{\pm0.98}$ & +20.89$_{\pm0.85}$ \\
\midrule
GCN       & 78.81$_{\pm0.73}$ & 84.14$_{\pm1.42}$ & 73.29$_{\pm0.74}$ & +10.85$_{\pm1.62}$ \\
GAT       & 78.93$_{\pm0.48}$ & 84.18$_{\pm1.56}$ & 73.15$_{\pm0.45}$ & +11.03$_{\pm1.54}$ \\
SAGE      & 78.46$_{\pm0.34}$ & 84.83$_{\pm1.19}$ & 72.13$_{\pm0.51}$ & +12.70$_{\pm1.23}$ \\
\midrule
BGRL      & 79.25$_{\pm0.53}$ & 83.97$_{\pm1.47}$ & 74.26$_{\pm1.07}$ & +9.70$_{\pm1.26}$ \\
GraphMAE  & 79.82$_{\pm0.55}$ & 84.59$_{\pm1.18}$ & 75.07$_{\pm0.94}$ & +9.51$_{\pm1.28}$ \\
\midrule
UniGraph2 & 74.94$_{\pm1.09}$ & 78.70$_{\pm2.19}$ & 70.09$_{\pm1.19}$ & \underline{+8.61}$_{\pm2.13}$ \\
GIT       & 78.74$_{\pm0.42}$ & 84.85$_{\pm1.01}$ & 72.36$_{\pm0.59}$ & +12.49$_{\pm1.00}$ \\
GFT       & 78.51$_{\pm0.58}$ & 84.74$_{\pm0.78}$ & 70.16$_{\pm1.44}$ & +14.58$_{\pm1.25}$ \\
\midrule
OFA       & 76.55$_{\pm0.68}$ & 80.16$_{\pm0.88}$ & 73.23$_{\pm0.67}$ & \textbf{+6.93}$_{\pm0.52}$ \\
LLaGA     & 72.84$_{\pm1.01}$ & 78.03$_{\pm1.64}$ & 68.33$_{\pm3.59}$ & +9.70$_{\pm4.45}$ \\
\bottomrule
\end{tabular}
\end{minipage}

\vspace{6pt}

\begin{minipage}[t]{0.49\linewidth}\centering
\textit{ElecComp}\\[1pt]
\begin{tabular}{lcccc}
\toprule
\textbf{Method} & \textbf{Acc} & \textbf{Acc(h)} & \textbf{Acc(t)} & \textbf{Gap}$\!\downarrow$ \\
\midrule
DeepWalk  & 71.10$_{\pm0.39}$ & 78.70$_{\pm0.83}$ & 57.74$_{\pm0.82}$ & +20.96$_{\pm1.46}$ \\
Node2vec  & 71.00$_{\pm0.09}$ & 78.27$_{\pm0.33}$ & 57.55$_{\pm0.65}$ & +20.72$_{\pm0.84}$ \\
\midrule
GCN       & 76.79$_{\pm0.13}$ & 86.78$_{\pm0.20}$ & 66.47$_{\pm0.37}$ & +20.31$_{\pm0.46}$ \\
GAT       & 75.20$_{\pm0.17}$ & 85.75$_{\pm0.32}$ & 64.30$_{\pm0.36}$ & +21.45$_{\pm0.30}$ \\
SAGE      & 76.37$_{\pm0.16}$ & 86.35$_{\pm0.33}$ & 65.74$_{\pm0.24}$ & +20.61$_{\pm0.32}$ \\
\midrule
BGRL      & 76.32$_{\pm0.12}$ & 86.22$_{\pm0.22}$ & 65.41$_{\pm0.25}$ & +20.81$_{\pm0.41}$ \\
GraphMAE  & 78.06$_{\pm0.16}$ & 87.38$_{\pm0.26}$ & 67.64$_{\pm0.46}$ & +19.75$_{\pm0.37}$ \\
\midrule
UniGraph2 & 75.47$_{\pm0.26}$ & 85.93$_{\pm0.11}$ & 64.18$_{\pm0.71}$ & +21.75$_{\pm0.72}$ \\
GIT       & 76.54$_{\pm0.17}$ & 86.48$_{\pm0.23}$ & 66.20$_{\pm0.29}$ & +20.28$_{\pm0.35}$ \\
GFT       & 77.36$_{\pm0.15}$ & 87.03$_{\pm0.29}$ & 66.86$_{\pm0.48}$ & +20.17$_{\pm0.55}$ \\
\midrule
OFA       & 78.08$_{\pm1.11}$ & 83.23$_{\pm1.00}$ & 73.32$_{\pm1.17}$ & \textbf{+9.91}$_{\pm0.50}$ \\
LLaGA     & 77.31$_{\pm2.53}$ & 81.20$_{\pm3.51}$ & 68.78$_{\pm3.96}$ & \underline{+12.42}$_{\pm5.56}$ \\
\bottomrule
\end{tabular}
\end{minipage}\hfill
\begin{minipage}[t]{0.49\linewidth}\centering
\textit{ElePhoto}\\[1pt]
\begin{tabular}{lcccc}
\toprule
\textbf{Method} & \textbf{Acc} & \textbf{Acc(h)} & \textbf{Acc(t)} & \textbf{Gap}$\!\downarrow$ \\
\midrule
DeepWalk  & 71.56$_{\pm0.14}$ & 74.75$_{\pm0.39}$ & 64.80$_{\pm0.46}$ & +9.95$_{\pm0.33}$ \\
Node2vec  & 71.71$_{\pm0.22}$ & 74.93$_{\pm0.58}$ & 65.31$_{\pm0.56}$ & +9.62$_{\pm1.03}$ \\
\midrule
GCN       & 78.42$_{\pm0.13}$ & 80.52$_{\pm0.26}$ & 76.89$_{\pm0.33}$ & +3.63$_{\pm0.52}$ \\
GAT       & 77.75$_{\pm0.14}$ & 80.12$_{\pm0.38}$ & 75.33$_{\pm0.17}$ & +4.79$_{\pm0.44}$ \\
SAGE      & 78.64$_{\pm0.07}$ & 80.37$_{\pm0.20}$ & 77.74$_{\pm0.19}$ & +2.63$_{\pm0.26}$ \\
\midrule
BGRL      & 78.90$_{\pm0.19}$ & 80.96$_{\pm0.39}$ & 77.59$_{\pm0.20}$ & +3.37$_{\pm0.43}$ \\
GraphMAE  & 79.89$_{\pm0.16}$ & 82.53$_{\pm0.19}$ & 78.12$_{\pm0.33}$ & +4.42$_{\pm0.30}$ \\
\midrule
UniGraph2 & 77.42$_{\pm0.12}$ & 80.07$_{\pm0.62}$ & 75.64$_{\pm0.31}$ & +4.43$_{\pm0.68}$ \\
GIT       & 78.70$_{\pm0.12}$ & 80.13$_{\pm0.25}$ & 77.79$_{\pm0.16}$ & \underline{+2.34}$_{\pm0.24}$ \\
GFT       & 79.11$_{\pm0.11}$ & 81.54$_{\pm0.38}$ & 77.92$_{\pm0.22}$ & +3.62$_{\pm0.58}$ \\
\midrule
OFA       & 79.24$_{\pm1.41}$ & 80.64$_{\pm1.83}$ & 78.13$_{\pm1.22}$ & +2.51$_{\pm0.90}$ \\
LLaGA     & 79.03$_{\pm1.18}$ & 76.35$_{\pm1.58}$ & 77.19$_{\pm2.68}$ & \textbf{$-$0.83}$_{\pm2.94}$ \\
\bottomrule
\end{tabular}
\end{minipage}

\vspace{6pt}

\begin{minipage}[t]{0.49\linewidth}\centering
\textit{AmazonRatings}\\[1pt]
\begin{tabular}{lcccc}
\toprule
\textbf{Method} & \textbf{Acc} & \textbf{Acc(h)} & \textbf{Acc(t)} & \textbf{Gap}$\!\downarrow$ \\
\midrule
DeepWalk  & 38.63$_{\pm0.44}$ & 38.91$_{\pm0.31}$ & 39.36$_{\pm0.65}$ & \underline{$-$0.45}$_{\pm0.81}$ \\
Node2vec  & 38.62$_{\pm0.23}$ & 39.16$_{\pm0.70}$ & 39.47$_{\pm0.42}$ & \textbf{$-$0.31}$_{\pm0.75}$ \\
\midrule
GCN       & 46.17$_{\pm0.31}$ & 43.10$_{\pm0.49}$ & 47.43$_{\pm0.44}$ & $-$4.33$_{\pm0.42}$ \\
GAT       & 43.34$_{\pm0.40}$ & 39.56$_{\pm0.68}$ & 45.67$_{\pm0.28}$ & $-$6.11$_{\pm0.62}$ \\
SAGE      & 49.16$_{\pm0.34}$ & 44.58$_{\pm0.46}$ & 51.91$_{\pm0.64}$ & $-$7.33$_{\pm0.78}$ \\
\midrule
BGRL      & 46.43$_{\pm0.07}$ & 40.07$_{\pm0.09}$ & 48.63$_{\pm0.05}$ & $-$8.55$_{\pm0.10}$ \\
GraphMAE  & 46.63$_{\pm0.00}$ & 41.19$_{\pm0.00}$ & 49.22$_{\pm0.00}$ & $-$8.03$_{\pm0.00}$ \\
\midrule
UniGraph2 & 45.16$_{\pm0.00}$ & 41.24$_{\pm0.00}$ & 47.50$_{\pm0.00}$ & $-$6.26$_{\pm0.00}$ \\
GIT       & 48.78$_{\pm0.04}$ & 43.62$_{\pm0.52}$ & 51.77$_{\pm0.28}$ & $-$8.14$_{\pm0.59}$ \\
GFT       & 47.43$_{\pm0.22}$ & 40.77$_{\pm0.24}$ & 50.94$_{\pm0.47}$ & $-$10.17$_{\pm0.51}$ \\
\midrule
OFA       & 36.56$_{\pm0.94}$ & 35.70$_{\pm1.08}$ & 37.27$_{\pm1.20}$ & $-$1.58$_{\pm1.38}$ \\
LLaGA     & 37.04$_{\pm2.14}$ & 36.70$_{\pm4.30}$ & 38.19$_{\pm1.29}$ & $-$1.49$_{\pm4.70}$ \\
\bottomrule
\end{tabular}
\end{minipage}\hfill
\begin{minipage}[t]{0.49\linewidth}\centering
\textit{Tolokers}\\[1pt]
\begin{tabular}{lcccc}
\toprule
\textbf{Method} & \textbf{Acc} & \textbf{Acc(h)} & \textbf{Acc(t)} & \textbf{Gap}$\!\downarrow$ \\
\midrule
DeepWalk  & 80.41$_{\pm0.29}$ & 75.78$_{\pm0.57}$ & 83.55$_{\pm0.32}$ & $-$7.77$_{\pm0.62}$ \\
Node2vec  & 80.19$_{\pm0.15}$ & 75.07$_{\pm1.24}$ & 83.64$_{\pm0.39}$ & $-$8.58$_{\pm1.52}$ \\
\midrule
GCN       & 78.66$_{\pm0.09}$ & 74.66$_{\pm0.49}$ & 83.42$_{\pm0.10}$ & $-$8.77$_{\pm0.54}$ \\
GAT       & 78.14$_{\pm0.04}$ & 72.13$_{\pm0.00}$ & 83.40$_{\pm0.18}$ & $-$11.27$_{\pm0.18}$ \\
SAGE      & 78.60$_{\pm0.33}$ & 72.32$_{\pm0.45}$ & 83.25$_{\pm0.68}$ & $-$10.92$_{\pm0.62}$ \\
\midrule
BGRL      & 80.42$_{\pm0.33}$ & 76.16$_{\pm0.85}$ & 82.63$_{\pm1.02}$ & \underline{$-$6.47}$_{\pm1.26}$ \\
GraphMAE  & 78.16$_{\pm0.00}$ & 72.13$_{\pm0.00}$ & 83.49$_{\pm0.00}$ & $-$11.36$_{\pm0.00}$ \\
\midrule
UniGraph2 & 78.14$_{\pm0.45}$ & 72.26$_{\pm1.47}$ & 83.08$_{\pm0.74}$ & $-$10.82$_{\pm1.86}$ \\
GIT       & 78.33$_{\pm0.33}$ & 72.21$_{\pm0.18}$ & 83.24$_{\pm0.52}$ & $-$11.03$_{\pm0.69}$ \\
GFT       & 78.16$_{\pm0.01}$ & 72.13$_{\pm0.00}$ & 83.49$_{\pm0.04}$ & $-$11.36$_{\pm0.04}$ \\
\midrule
OFA       & 78.60$_{\pm0.57}$ & 77.22$_{\pm0.24}$ & 79.85$_{\pm0.50}$ & \textbf{$-$2.63}$_{\pm0.61}$ \\
LLaGA     & 78.16$_{\pm0.39}$ & 72.62$_{\pm0.77}$ & 83.84$_{\pm0.27}$ & $-$11.21$_{\pm0.73}$ \\
\bottomrule
\end{tabular}
\end{minipage}

\end{table}

\FloatBarrier

\subsection{Interpretation --- full per-dataset results}
\label{app:interpret_full}

The main-text interpretation section reports node-level
subgraph-ablation attribution fidelity at top-$k{=}10\%$ edges.
Here we provide the complementary details: a sparsity sweep over
multiple top-$k$ levels, a per-dataset decomposition of
Fid$^+$ / Fid$^-$ / characterization score at top-$10\%$, and
atom-level graph-classification results on molecular datasets.



\begin{table}[!htbp]
\centering
\caption{\textbf{Subgraph-ablation sparsity sweep} ($\Delta_{\text{char}}\!\times\!100$,
higher is better). $\Delta_{\text{char}} = \text{char}_{\text{grad}} - \text{char}_{\text{rand}}$
at four edge sparsity levels in each test node's $K{=}2$-hop receptive field. Cells:
mean$_{\pm\text{std}}$ across 5 seeds; std for $\Delta$ is propagated as
$\sqrt{\sigma_{\text{grad}}^2 + \sigma_{\text{rand}}^2}$.
\textbf{Bold} / \underline{underline} = best / second per dataset and sparsity.}
\label{app:interp_v2_sparsity}
\scriptsize
\setlength{\tabcolsep}{1.8pt}
\renewcommand{\arraystretch}{1.05}

\begin{minipage}[t]{0.49\linewidth}\centering
\textit{Cora}\\[1pt]
\begin{tabular}{lcccc}
\toprule
\textbf{Method} & \textbf{$k{=}5\%$} & \textbf{$k{=}10\%$} & \textbf{$k{=}20\%$} & \textbf{$k{=}50\%$} \\
\midrule
GCN       & +9.7$_{\pm1.2}$ & +11.6$_{\pm2.4}$ & +13.0$_{\pm2.3}$ & +9.5$_{\pm3.2}$ \\
GAT       & +9.0$_{\pm2.2}$ & +11.2$_{\pm2.0}$ & +14.7$_{\pm1.5}$ & +13.0$_{\pm3.9}$ \\
SAGE      & \underline{+18.9}$_{\pm5.4}$ & \underline{+30.2}$_{\pm1.7}$ & \underline{+39.1}$_{\pm3.5}$ & \underline{+40.0}$_{\pm4.7}$ \\
\midrule
BGRL      & +12.3$_{\pm1.6}$ & +14.2$_{\pm1.7}$ & +16.0$_{\pm2.6}$ & +11.6$_{\pm3.2}$ \\
GraphMAE  & +9.0$_{\pm1.9}$ & +12.8$_{\pm0.6}$ & +15.7$_{\pm1.2}$ & +12.5$_{\pm2.5}$ \\
\midrule
UniGraph2 & +4.3$_{\pm4.0}$ & +5.4$_{\pm2.5}$ & +5.4$_{\pm2.3}$ & +8.1$_{\pm1.4}$ \\
GFT       & \textbf{+20.3}$_{\pm5.3}$ & \textbf{+30.6}$_{\pm3.5}$ & \textbf{+40.5}$_{\pm2.8}$ & \textbf{+40.9}$_{\pm4.8}$ \\
GIT       & +14.5$_{\pm13.8}$ & +21.3$_{\pm18.6}$ & +27.1$_{\pm23.6}$ & +28.1$_{\pm28.7}$ \\
\bottomrule
\end{tabular}
\end{minipage}\hfill
\begin{minipage}[t]{0.49\linewidth}\centering
\textit{CiteSeer}\\[1pt]
\begin{tabular}{lcccc}
\toprule
\textbf{Method} & \textbf{$k{=}5\%$} & \textbf{$k{=}10\%$} & \textbf{$k{=}20\%$} & \textbf{$k{=}50\%$} \\
\midrule
GCN       & +4.5$_{\pm5.3}$ & +5.4$_{\pm6.4}$ & +5.3$_{\pm7.1}$ & +4.5$_{\pm9.0}$ \\
GAT       & +3.0$_{\pm1.8}$ & +3.4$_{\pm1.9}$ & +4.2$_{\pm2.5}$ & +3.7$_{\pm2.9}$ \\
SAGE      & +8.7$_{\pm6.9}$ & +12.0$_{\pm8.6}$ & +15.0$_{\pm9.5}$ & +17.8$_{\pm12.1}$ \\
\midrule
BGRL      & +8.0$_{\pm1.2}$ & +8.9$_{\pm0.7}$ & +10.0$_{\pm1.6}$ & +7.3$_{\pm1.0}$ \\
GraphMAE  & +5.1$_{\pm4.4}$ & +6.6$_{\pm4.5}$ & +7.6$_{\pm5.4}$ & +7.5$_{\pm7.3}$ \\
\midrule
UniGraph2 & +1.0$_{\pm3.3}$ & +2.7$_{\pm3.4}$ & +5.4$_{\pm3.2}$ & +4.4$_{\pm3.0}$ \\
GFT       & \textbf{+13.9}$_{\pm20.5}$ & \textbf{+19.5}$_{\pm23.9}$ & \textbf{+22.9}$_{\pm26.0}$ & \textbf{+27.6}$_{\pm35.2}$ \\
GIT       & \underline{+10.3}$_{\pm3.7}$ & \underline{+13.6}$_{\pm4.0}$ & \underline{+16.8}$_{\pm2.9}$ & \underline{+19.6}$_{\pm3.0}$ \\
\bottomrule
\end{tabular}
\end{minipage}
\vspace{6pt}\par
\begin{minipage}[t]{0.49\linewidth}\centering
\textit{PubMed}\\[1pt]
\begin{tabular}{lcccc}
\toprule
\textbf{Method} & \textbf{$k{=}5\%$} & \textbf{$k{=}10\%$} & \textbf{$k{=}20\%$} & \textbf{$k{=}50\%$} \\
\midrule
GCN       & +12.8$_{\pm3.4}$ & +12.7$_{\pm2.9}$ & +12.7$_{\pm2.3}$ & +8.3$_{\pm2.6}$ \\
GAT       & +17.9$_{\pm5.4}$ & +18.4$_{\pm3.6}$ & +17.4$_{\pm2.0}$ & +12.2$_{\pm3.8}$ \\
SAGE      & \underline{+31.0}$_{\pm3.5}$ & \underline{+33.7}$_{\pm4.1}$ & \underline{+34.0}$_{\pm3.7}$ & \underline{+25.8}$_{\pm3.7}$ \\
\midrule
BGRL      & +11.8$_{\pm4.9}$ & +11.0$_{\pm4.8}$ & +10.4$_{\pm4.6}$ & +6.2$_{\pm5.4}$ \\
GraphMAE  & +17.3$_{\pm2.0}$ & +17.3$_{\pm1.2}$ & +17.2$_{\pm0.4}$ & +12.5$_{\pm1.6}$ \\
\midrule
UniGraph2 & +7.7$_{\pm6.4}$ & +10.7$_{\pm7.5}$ & +14.3$_{\pm5.7}$ & +12.9$_{\pm6.5}$ \\
GFT       & \textbf{+36.4}$_{\pm4.7}$ & \textbf{+40.1}$_{\pm5.2}$ & \textbf{+40.1}$_{\pm4.7}$ & \textbf{+31.7}$_{\pm5.2}$ \\
GIT       & +27.8$_{\pm5.8}$ & +30.1$_{\pm6.0}$ & +30.1$_{\pm5.8}$ & +22.4$_{\pm6.0}$ \\
\bottomrule
\end{tabular}
\end{minipage}\hfill
\begin{minipage}[t]{0.49\linewidth}\centering
\textit{WikiCS}\\[1pt]
\begin{tabular}{lcccc}
\toprule
\textbf{Method} & \textbf{$k{=}5\%$} & \textbf{$k{=}10\%$} & \textbf{$k{=}20\%$} & \textbf{$k{=}50\%$} \\
\midrule
GCN       & +26.1$_{\pm1.3}$ & +23.7$_{\pm2.2}$ & +21.7$_{\pm2.2}$ & +18.6$_{\pm2.2}$ \\
GAT       & +26.8$_{\pm3.5}$ & +25.8$_{\pm4.6}$ & +24.4$_{\pm4.8}$ & +19.7$_{\pm4.8}$ \\
SAGE      & +52.5$_{\pm1.2}$ & +51.8$_{\pm1.6}$ & +49.9$_{\pm1.8}$ & +43.1$_{\pm4.2}$ \\
\midrule
BGRL      & +23.4$_{\pm0.4}$ & +20.1$_{\pm2.3}$ & +18.8$_{\pm2.4}$ & +16.3$_{\pm2.4}$ \\
GraphMAE  & +25.3$_{\pm2.7}$ & +25.4$_{\pm2.2}$ & +23.2$_{\pm2.6}$ & +16.5$_{\pm2.2}$ \\
\midrule
UniGraph2 & +22.4$_{\pm4.5}$ & +24.6$_{\pm4.5}$ & +22.3$_{\pm4.0}$ & +20.5$_{\pm3.6}$ \\
GFT       & \textbf{+64.8}$_{\pm1.6}$ & \textbf{+63.7}$_{\pm1.8}$ & \textbf{+62.2}$_{\pm2.0}$ & \textbf{+55.7}$_{\pm3.1}$ \\
GIT       & \underline{+54.1}$_{\pm2.5}$ & \underline{+53.5}$_{\pm3.1}$ & \underline{+51.7}$_{\pm3.0}$ & \underline{+45.9}$_{\pm4.7}$ \\
\bottomrule
\end{tabular}
\end{minipage}
\vspace{6pt}\par
\begin{minipage}[t]{0.49\linewidth}\centering
\textit{ElecComp}\\[1pt]
\begin{tabular}{lcccc}
\toprule
\textbf{Method} & \textbf{$k{=}5\%$} & \textbf{$k{=}10\%$} & \textbf{$k{=}20\%$} & \textbf{$k{=}50\%$} \\
\midrule
GCN       & +13.3$_{\pm5.0}$ & +16.1$_{\pm6.1}$ & +17.6$_{\pm6.9}$ & +18.9$_{\pm7.6}$ \\
GAT       & +6.8$_{\pm5.4}$ & +8.1$_{\pm5.4}$ & +12.6$_{\pm5.0}$ & +17.8$_{\pm4.8}$ \\
SAGE      & \underline{+23.1}$_{\pm7.6}$ & \underline{+25.7}$_{\pm7.1}$ & \underline{+30.0}$_{\pm5.6}$ & \underline{+27.3}$_{\pm7.0}$ \\
\midrule
BGRL      & +14.4$_{\pm7.0}$ & +18.2$_{\pm7.4}$ & +20.4$_{\pm7.7}$ & +21.6$_{\pm7.2}$ \\
GraphMAE  & +7.1$_{\pm7.3}$ & +9.1$_{\pm6.4}$ & +14.7$_{\pm7.3}$ & +18.2$_{\pm7.2}$ \\
\midrule
UniGraph2 & +4.5$_{\pm6.7}$ & +7.3$_{\pm6.1}$ & +8.5$_{\pm8.2}$ & +10.1$_{\pm7.8}$ \\
GFT       & +23.0$_{\pm6.7}$ & +24.9$_{\pm5.9}$ & +28.6$_{\pm5.1}$ & +26.6$_{\pm5.2}$ \\
GIT       & \textbf{+24.3}$_{\pm6.6}$ & \textbf{+26.5}$_{\pm6.3}$ & \textbf{+30.3}$_{\pm4.5}$ & \textbf{+27.7}$_{\pm5.6}$ \\
\bottomrule
\end{tabular}
\end{minipage}\hfill
\begin{minipage}[t]{0.49\linewidth}\centering
\textit{ElePhoto}\\[1pt]
\begin{tabular}{lcccc}
\toprule
\textbf{Method} & \textbf{$k{=}5\%$} & \textbf{$k{=}10\%$} & \textbf{$k{=}20\%$} & \textbf{$k{=}50\%$} \\
\midrule
GCN       & +12.2$_{\pm4.3}$ & +16.5$_{\pm4.9}$ & +18.6$_{\pm4.6}$ & +16.9$_{\pm4.8}$ \\
GAT       & +7.8$_{\pm3.7}$ & +9.6$_{\pm4.4}$ & +15.1$_{\pm1.8}$ & +19.6$_{\pm3.8}$ \\
SAGE      & \textbf{+23.1}$_{\pm6.2}$ & \underline{+24.8}$_{\pm6.6}$ & \underline{+24.9}$_{\pm6.1}$ & \textbf{+24.1}$_{\pm6.5}$ \\
\midrule
BGRL      & +13.9$_{\pm3.9}$ & +17.0$_{\pm4.3}$ & +18.1$_{\pm4.6}$ & +16.3$_{\pm5.7}$ \\
GraphMAE  & +8.0$_{\pm3.7}$ & +10.6$_{\pm4.5}$ & +14.2$_{\pm3.3}$ & +19.8$_{\pm3.6}$ \\
\midrule
UniGraph2 & +5.8$_{\pm1.6}$ & +6.4$_{\pm2.6}$ & +10.3$_{\pm3.6}$ & +11.4$_{\pm3.1}$ \\
GFT       & +22.2$_{\pm4.1}$ & +24.2$_{\pm4.6}$ & +24.9$_{\pm4.5}$ & +23.3$_{\pm5.0}$ \\
GIT       & \underline{+22.7}$_{\pm5.0}$ & \textbf{+25.0}$_{\pm4.8}$ & \textbf{+25.6}$_{\pm4.4}$ & \underline{+23.9}$_{\pm5.4}$ \\
\bottomrule
\end{tabular}
\end{minipage}
\vspace{6pt}\par
\begin{minipage}[t]{0.49\linewidth}\centering
\textit{AmazonRatings}\\[1pt]
\begin{tabular}{lcccc}
\toprule
\textbf{Method} & \textbf{$k{=}5\%$} & \textbf{$k{=}10\%$} & \textbf{$k{=}20\%$} & \textbf{$k{=}50\%$} \\
\midrule
GCN       & +3.4$_{\pm2.7}$ & +3.5$_{\pm3.6}$ & +3.6$_{\pm4.5}$ & +2.9$_{\pm5.9}$ \\
GAT       & +1.9$_{\pm3.4}$ & +2.0$_{\pm4.0}$ & +2.1$_{\pm6.4}$ & +2.2$_{\pm10.1}$ \\
SAGE      & \underline{+12.4}$_{\pm5.6}$ & \underline{+16.0}$_{\pm5.7}$ & \underline{+20.4}$_{\pm4.8}$ & \underline{+26.2}$_{\pm6.0}$ \\
\midrule
BGRL      & +6.4$_{\pm5.3}$ & +5.9$_{\pm5.6}$ & +7.8$_{\pm6.6}$ & +6.4$_{\pm6.7}$ \\
GraphMAE  & +5.8$_{\pm2.9}$ & +6.7$_{\pm3.4}$ & +6.6$_{\pm4.8}$ & +7.0$_{\pm5.9}$ \\
\midrule
UniGraph2 & +2.6$_{\pm5.5}$ & +5.0$_{\pm3.7}$ & +6.2$_{\pm4.8}$ & +8.2$_{\pm2.4}$ \\
GFT       & \textbf{+15.8}$_{\pm2.6}$ & \textbf{+17.3}$_{\pm2.3}$ & \textbf{+24.3}$_{\pm1.4}$ & \textbf{+29.8}$_{\pm3.2}$ \\
GIT       & +10.3$_{\pm5.3}$ & +14.2$_{\pm3.1}$ & +19.2$_{\pm5.2}$ & +20.6$_{\pm6.1}$ \\
\bottomrule
\end{tabular}
\end{minipage}\hfill
\begin{minipage}[t]{0.49\linewidth}\centering
\textit{Tolokers}\\[1pt]
\begin{tabular}{lcccc}
\toprule
\textbf{Method} & \textbf{$k{=}5\%$} & \textbf{$k{=}10\%$} & \textbf{$k{=}20\%$} & \textbf{$k{=}50\%$} \\
\midrule
GCN       & +22.4$_{\pm0.3}$ & +22.4$_{\pm0.3}$ & +22.3$_{\pm0.3}$ & +21.1$_{\pm0.3}$ \\
GAT       & +0.2$_{\pm0.0}$ & +0.2$_{\pm0.0}$ & +0.2$_{\pm0.0}$ & +0.1$_{\pm0.0}$ \\
SAGE      & \underline{+23.6}$_{\pm0.6}$ & \underline{+23.6}$_{\pm0.6}$ & \underline{+23.7}$_{\pm0.6}$ & \underline{+23.2}$_{\pm0.8}$ \\
\midrule
BGRL      & +8.5$_{\pm1.7}$ & +8.2$_{\pm1.8}$ & +7.5$_{\pm1.9}$ & +3.7$_{\pm2.0}$ \\
GraphMAE  & +1.2$_{\pm0.8}$ & +1.2$_{\pm0.8}$ & +1.2$_{\pm0.8}$ & +1.2$_{\pm0.8}$ \\
\midrule
UniGraph2 & +11.4$_{\pm4.6}$ & +14.2$_{\pm3.9}$ & +17.2$_{\pm4.6}$ & +16.1$_{\pm5.7}$ \\
GFT       & \textbf{+30.2}$_{\pm1.5}$ & \textbf{+30.2}$_{\pm1.5}$ & \textbf{+30.3}$_{\pm1.4}$ & \textbf{+29.8}$_{\pm1.4}$ \\
GIT       & +9.9$_{\pm1.0}$ & +9.8$_{\pm1.0}$ & +9.8$_{\pm1.1}$ & +9.3$_{\pm1.3}$ \\
\bottomrule
\end{tabular}
\end{minipage}

\end{table}

\begin{table}[!htbp]
\centering
\caption{\textbf{Subgraph-ablation Fid$^+$ / Fid$^-$ / char decomposition at $k\!=\!10\%$ edges} ($\times 100$).
For each dataset/method, columns report the gradient-saliency value (G) and the random-edge baseline (R)
for Fid$^+$ (drop in confidence when salient edges are masked; higher is better) and Fid$^-$
(preservation when non-salient edges are masked; lower is better). char $=$ harmonic mean of Fid$^+$
and $1{-}$Fid$^-$. $\Delta = $ char(G) $-$ char(R), the quantity reported in the main-text table;
all cells report mean$_{\pm\text{std}}$ over 5 seeds ($\Delta$ uses propagated std).
\textbf{Bold} / \underline{underline} = best / second per dataset on $\Delta$.}
\label{app:interp_v2_decomp}
\scriptsize
\setlength{\tabcolsep}{1.2pt}
\renewcommand{\arraystretch}{1.05}

\begin{minipage}[t]{0.49\linewidth}\centering
\textit{Cora}\\[1pt]
\resizebox{\linewidth}{!}{%
\begin{tabular}{lcccccc|c}
\toprule
\multirow{2}{*}{\textbf{Method}} & \multicolumn{2}{c}{\textbf{Fid$^+$}} & \multicolumn{2}{c}{\textbf{Fid$^-$}} & \multicolumn{2}{c|}{\textbf{char}} & \multirow{2}{*}{\textbf{$\Delta$}} \\
& G & R & G & R & G & R & \\
\midrule
GCN       & 5.4$_{\pm2.4}$ & $-$0.0$_{\pm0.6}$ & $-$0.9$_{\pm0.4}$ & 5.2$_{\pm2.3}$ & 14.8$_{\pm2.2}$ & 3.2$_{\pm1.1}$ & +11.6$_{\pm2.4}$ \\
GAT       & 11.0$_{\pm3.7}$ & 2.1$_{\pm1.1}$ & 4.9$_{\pm2.4}$ & 12.9$_{\pm3.7}$ & 16.2$_{\pm1.8}$ & 5.0$_{\pm0.9}$ & +11.2$_{\pm2.0}$ \\
SAGE      & 24.8$_{\pm1.2}$ & 3.2$_{\pm1.1}$ & 1.8$_{\pm1.9}$ & 27.6$_{\pm3.4}$ & 36.2$_{\pm0.7}$ & 5.9$_{\pm1.5}$ & \underline{+30.2}$_{\pm1.7}$ \\
\midrule
BGRL      & 7.8$_{\pm1.5}$ & 0.5$_{\pm1.2}$ & $-$1.6$_{\pm3.3}$ & 7.6$_{\pm2.9}$ & 18.7$_{\pm1.1}$ & 4.5$_{\pm1.4}$ & +14.2$_{\pm1.7}$ \\
GraphMAE  & 9.0$_{\pm0.9}$ & 0.3$_{\pm0.5}$ & 5.1$_{\pm3.2}$ & 11.6$_{\pm2.8}$ & 15.9$_{\pm0.2}$ & 3.1$_{\pm0.6}$ & +12.8$_{\pm0.6}$ \\
\midrule
UniGraph2 & 7.4$_{\pm2.4}$ & 3.5$_{\pm2.4}$ & 13.2$_{\pm0.5}$ & 14.6$_{\pm2.2}$ & 12.9$_{\pm1.7}$ & 7.5$_{\pm1.8}$ & +5.4$_{\pm2.5}$ \\
GFT       & 30.3$_{\pm3.7}$ & 5.4$_{\pm1.7}$ & 6.3$_{\pm2.4}$ & 34.8$_{\pm3.5}$ & 38.5$_{\pm3.0}$ & 7.8$_{\pm1.8}$ & \textbf{+30.6}$_{\pm3.5}$ \\
GIT       & 17.2$_{\pm13.1}$ & 2.2$_{\pm2.0}$ & 0.9$_{\pm2.5}$ & 18.5$_{\pm14.5}$ & 25.4$_{\pm18.3}$ & 4.2$_{\pm3.4}$ & +21.3$_{\pm18.6}$ \\
\bottomrule
\end{tabular}}
\end{minipage}\hfill
\begin{minipage}[t]{0.49\linewidth}\centering
\textit{CiteSeer}\\[1pt]
\resizebox{\linewidth}{!}{%
\begin{tabular}{lcccccc|c}
\toprule
\multirow{2}{*}{\textbf{Method}} & \multicolumn{2}{c}{\textbf{Fid$^+$}} & \multicolumn{2}{c}{\textbf{Fid$^-$}} & \multicolumn{2}{c|}{\textbf{char}} & \multirow{2}{*}{\textbf{$\Delta$}} \\
& G & R & G & R & G & R & \\
\midrule
GCN       & 4.0$_{\pm6.7}$ & 1.4$_{\pm2.9}$ & $-$0.8$_{\pm2.6}$ & 2.6$_{\pm4.9}$ & 9.5$_{\pm5.7}$ & 4.1$_{\pm2.8}$ & +5.4$_{\pm6.4}$ \\
GAT       & 0.9$_{\pm0.4}$ & $-$0.0$_{\pm0.4}$ & 0.5$_{\pm0.5}$ & 1.5$_{\pm0.9}$ & 5.7$_{\pm1.8}$ & 2.3$_{\pm0.7}$ & +3.4$_{\pm1.9}$ \\
SAGE      & 12.1$_{\pm5.7}$ & 4.5$_{\pm2.4}$ & 0.5$_{\pm0.4}$ & 10.8$_{\pm4.1}$ & 19.8$_{\pm7.9}$ & 7.8$_{\pm3.4}$ & +12.0$_{\pm8.6}$ \\
\midrule
BGRL      & 4.9$_{\pm0.8}$ & 0.2$_{\pm0.1}$ & $-$2.1$_{\pm2.0}$ & 3.5$_{\pm1.1}$ & 14.2$_{\pm0.6}$ & 5.3$_{\pm0.4}$ & +9.0$_{\pm0.7}$ \\
GraphMAE  & 4.9$_{\pm4.4}$ & 1.2$_{\pm2.8}$ & 1.9$_{\pm1.7}$ & 3.9$_{\pm3.3}$ & 11.6$_{\pm3.6}$ & 5.0$_{\pm2.6}$ & +6.6$_{\pm4.5}$ \\
\midrule
UniGraph2 & 5.3$_{\pm4.9}$ & 4.2$_{\pm4.3}$ & 6.4$_{\pm3.9}$ & 9.2$_{\pm5.9}$ & 12.7$_{\pm2.4}$ & 9.9$_{\pm2.4}$ & +2.8$_{\pm3.4}$ \\
GFT       & 27.4$_{\pm21.1}$ & 10.2$_{\pm9.2}$ & 2.6$_{\pm2.7}$ & 23.0$_{\pm16.3}$ & 32.6$_{\pm21.7}$ & 13.0$_{\pm10.0}$ & \textbf{+19.5}$_{\pm23.9}$ \\
GIT       & 13.7$_{\pm2.6}$ & 4.9$_{\pm1.2}$ & 0.5$_{\pm0.2}$ & 12.4$_{\pm0.7}$ & 22.2$_{\pm3.8}$ & 8.6$_{\pm1.4}$ & \underline{+13.6}$_{\pm4.0}$ \\
\bottomrule
\end{tabular}}
\end{minipage}
\vspace{6pt}\par
\begin{minipage}[t]{0.49\linewidth}\centering
\textit{PubMed}\\[1pt]
\resizebox{\linewidth}{!}{%
\begin{tabular}{lcccccc|c}
\toprule
\multirow{2}{*}{\textbf{Method}} & \multicolumn{2}{c}{\textbf{Fid$^+$}} & \multicolumn{2}{c}{\textbf{Fid$^-$}} & \multicolumn{2}{c|}{\textbf{char}} & \multirow{2}{*}{\textbf{$\Delta$}} \\
& G & R & G & R & G & R & \\
\midrule
GCN       & 8.2$_{\pm3.4}$ & 0.9$_{\pm1.0}$ & 0.2$_{\pm2.0}$ & 9.5$_{\pm1.8}$ & 15.1$_{\pm2.7}$ & 2.4$_{\pm1.0}$ & +12.7$_{\pm2.9}$ \\
GAT       & 18.8$_{\pm2.8}$ & 1.9$_{\pm1.4}$ & 3.9$_{\pm0.9}$ & 17.0$_{\pm0.8}$ & 21.6$_{\pm3.3}$ & 3.2$_{\pm1.5}$ & +18.4$_{\pm3.6}$ \\
SAGE      & 25.3$_{\pm2.6}$ & 1.9$_{\pm1.0}$ & 1.6$_{\pm0.8}$ & 20.7$_{\pm2.4}$ & 37.2$_{\pm4.0}$ & 3.4$_{\pm1.0}$ & \underline{+33.7}$_{\pm4.1}$ \\
\midrule
BGRL      & 3.8$_{\pm4.7}$ & $-$0.0$_{\pm0.7}$ & 0.2$_{\pm0.9}$ & 2.7$_{\pm3.7}$ & 13.2$_{\pm4.8}$ & 2.1$_{\pm0.8}$ & +11.0$_{\pm4.8}$ \\
GraphMAE  & 13.1$_{\pm2.0}$ & 0.5$_{\pm0.8}$ & 3.9$_{\pm0.4}$ & 11.5$_{\pm0.9}$ & 19.4$_{\pm0.9}$ & 2.1$_{\pm0.7}$ & +17.3$_{\pm1.2}$ \\
\midrule
UniGraph2 & 12.3$_{\pm7.1}$ & 0.4$_{\pm2.0}$ & 8.5$_{\pm3.1}$ & 21.1$_{\pm2.4}$ & 15.5$_{\pm7.3}$ & 4.7$_{\pm1.8}$ & +10.7$_{\pm7.5}$ \\
GFT       & 31.9$_{\pm3.2}$ & 1.6$_{\pm1.0}$ & 1.4$_{\pm1.5}$ & 23.5$_{\pm0.4}$ & 43.2$_{\pm5.0}$ & 3.1$_{\pm1.3}$ & \textbf{+40.1}$_{\pm5.2}$ \\
GIT       & 23.3$_{\pm4.0}$ & 1.8$_{\pm0.6}$ & 1.0$_{\pm0.9}$ & 19.2$_{\pm2.4}$ & 33.0$_{\pm6.0}$ & 2.9$_{\pm0.2}$ & +30.1$_{\pm6.0}$ \\
\bottomrule
\end{tabular}}
\end{minipage}\hfill
\begin{minipage}[t]{0.49\linewidth}\centering
\textit{WikiCS}\\[1pt]
\resizebox{\linewidth}{!}{%
\begin{tabular}{lcccccc|c}
\toprule
\multirow{2}{*}{\textbf{Method}} & \multicolumn{2}{c}{\textbf{Fid$^+$}} & \multicolumn{2}{c}{\textbf{Fid$^-$}} & \multicolumn{2}{c|}{\textbf{char}} & \multirow{2}{*}{\textbf{$\Delta$}} \\
& G & R & G & R & G & R & \\
\midrule
GCN       & 16.2$_{\pm2.6}$ & 0.3$_{\pm0.3}$ & $-$2.7$_{\pm0.3}$ & 7.3$_{\pm3.5}$ & 26.2$_{\pm2.1}$ & 2.5$_{\pm0.8}$ & +23.7$_{\pm2.2}$ \\
GAT       & 18.4$_{\pm5.7}$ & $-$0.1$_{\pm0.5}$ & $-$0.1$_{\pm1.0}$ & 14.3$_{\pm4.7}$ & 28.0$_{\pm4.5}$ & 2.2$_{\pm0.4}$ & +25.8$_{\pm4.6}$ \\
SAGE      & 41.8$_{\pm1.5}$ & 0.7$_{\pm0.7}$ & 0.1$_{\pm0.1}$ & 20.3$_{\pm2.6}$ & 54.2$_{\pm1.6}$ & 2.4$_{\pm0.5}$ & +51.8$_{\pm1.6}$ \\
\midrule
BGRL      & 11.0$_{\pm3.8}$ & 0.4$_{\pm0.4}$ & $-$0.9$_{\pm0.4}$ & 6.4$_{\pm1.9}$ & 22.8$_{\pm2.3}$ & 2.6$_{\pm0.6}$ & +20.1$_{\pm2.3}$ \\
GraphMAE  & 19.1$_{\pm2.6}$ & 0.9$_{\pm1.3}$ & $-$0.1$_{\pm0.2}$ & 10.6$_{\pm3.2}$ & 28.1$_{\pm1.8}$ & 2.7$_{\pm1.3}$ & +25.4$_{\pm2.2}$ \\
\midrule
UniGraph2 & 26.6$_{\pm4.2}$ & 3.1$_{\pm0.7}$ & 2.9$_{\pm2.5}$ & 18.6$_{\pm2.8}$ & 31.2$_{\pm4.4}$ & 6.6$_{\pm1.0}$ & +24.6$_{\pm4.5}$ \\
GFT       & 58.3$_{\pm1.7}$ & 1.6$_{\pm0.7}$ & 0.1$_{\pm0.2}$ & 24.6$_{\pm2.2}$ & 66.8$_{\pm1.5}$ & 3.2$_{\pm0.9}$ & \textbf{+63.7}$_{\pm1.8}$ \\
GIT       & 42.9$_{\pm2.4}$ & 0.6$_{\pm0.8}$ & 0.0$_{\pm0.1}$ & 20.7$_{\pm3.2}$ & 55.9$_{\pm2.9}$ & 2.4$_{\pm0.9}$ & \underline{+53.5}$_{\pm3.1}$ \\
\bottomrule
\end{tabular}}
\end{minipage}
\vspace{6pt}\par
\begin{minipage}[t]{0.49\linewidth}\centering
\textit{ElecComp}\\[1pt]
\resizebox{\linewidth}{!}{%
\begin{tabular}{lcccccc|c}
\toprule
\multirow{2}{*}{\textbf{Method}} & \multicolumn{2}{c}{\textbf{Fid$^+$}} & \multicolumn{2}{c}{\textbf{Fid$^-$}} & \multicolumn{2}{c|}{\textbf{char}} & \multirow{2}{*}{\textbf{$\Delta$}} \\
& G & R & G & R & G & R & \\
\midrule
GCN       & 18.1$_{\pm5.2}$ & 5.7$_{\pm2.7}$ & 3.2$_{\pm1.5}$ & 18.8$_{\pm6.8}$ & 26.0$_{\pm5.2}$ & 9.9$_{\pm3.0}$ & +16.1$_{\pm6.1}$ \\
GAT       & 10.8$_{\pm1.6}$ & 5.9$_{\pm3.3}$ & 11.4$_{\pm4.1}$ & 20.9$_{\pm0.9}$ & 16.8$_{\pm3.2}$ & 8.7$_{\pm4.3}$ & +8.1$_{\pm5.4}$ \\
SAGE      & 27.9$_{\pm4.2}$ & 7.3$_{\pm4.1}$ & 5.2$_{\pm2.3}$ & 23.4$_{\pm1.7}$ & 36.4$_{\pm5.5}$ & 10.7$_{\pm4.5}$ & \underline{+25.7}$_{\pm7.1}$ \\
\midrule
BGRL      & 20.2$_{\pm4.9}$ & 6.3$_{\pm3.9}$ & 4.5$_{\pm1.2}$ & 20.5$_{\pm4.7}$ & 29.0$_{\pm5.8}$ & 10.8$_{\pm4.5}$ & +18.2$_{\pm7.4}$ \\
GraphMAE  & 12.5$_{\pm4.6}$ & 6.0$_{\pm2.9}$ & 14.1$_{\pm4.8}$ & 25.4$_{\pm2.1}$ & 17.6$_{\pm5.4}$ & 8.5$_{\pm3.5}$ & +9.1$_{\pm6.4}$ \\
\midrule
UniGraph2 & 12.5$_{\pm3.8}$ & 5.1$_{\pm3.8}$ & 16.8$_{\pm2.4}$ & 27.7$_{\pm1.6}$ & 16.9$_{\pm4.4}$ & 9.6$_{\pm4.2}$ & +7.3$_{\pm6.1}$ \\
GFT       & 28.1$_{\pm3.4}$ & 7.7$_{\pm3.2}$ & 4.4$_{\pm1.9}$ & 22.7$_{\pm0.5}$ & 35.6$_{\pm4.5}$ & 10.7$_{\pm3.9}$ & +24.9$_{\pm5.9}$ \\
GIT       & 28.5$_{\pm3.7}$ & 7.3$_{\pm3.7}$ & 4.1$_{\pm1.6}$ & 23.2$_{\pm1.4}$ & 37.1$_{\pm4.8}$ & 10.6$_{\pm4.0}$ & \textbf{+26.5}$_{\pm6.3}$ \\
\bottomrule
\end{tabular}}
\end{minipage}\hfill
\begin{minipage}[t]{0.49\linewidth}\centering
\textit{ElePhoto}\\[1pt]
\resizebox{\linewidth}{!}{%
\begin{tabular}{lcccccc|c}
\toprule
\multirow{2}{*}{\textbf{Method}} & \multicolumn{2}{c}{\textbf{Fid$^+$}} & \multicolumn{2}{c}{\textbf{Fid$^-$}} & \multicolumn{2}{c|}{\textbf{char}} & \multirow{2}{*}{\textbf{$\Delta$}} \\
& G & R & G & R & G & R & \\
\midrule
GCN       & 16.2$_{\pm3.7}$ & 5.2$_{\pm3.3}$ & 4.0$_{\pm0.9}$ & 14.5$_{\pm2.2}$ & 25.2$_{\pm2.6}$ & 8.7$_{\pm4.2}$ & +16.5$_{\pm4.9}$ \\
GAT       & 11.0$_{\pm4.0}$ & 5.2$_{\pm2.3}$ & 8.5$_{\pm3.2}$ & 15.6$_{\pm3.0}$ & 17.6$_{\pm3.6}$ & 8.0$_{\pm2.5}$ & +9.6$_{\pm4.4}$ \\
SAGE      & 23.3$_{\pm3.8}$ & 4.5$_{\pm4.0}$ & 2.1$_{\pm0.9}$ & 14.0$_{\pm0.2}$ & 32.1$_{\pm4.4}$ & 7.2$_{\pm4.9}$ & \underline{+24.8}$_{\pm6.6}$ \\
\midrule
BGRL      & 16.2$_{\pm2.1}$ & 5.1$_{\pm2.5}$ & 5.4$_{\pm2.4}$ & 15.9$_{\pm0.7}$ & 25.4$_{\pm2.1}$ & 8.4$_{\pm3.7}$ & +17.0$_{\pm4.3}$ \\
GraphMAE  & 13.1$_{\pm2.8}$ & 6.0$_{\pm3.9}$ & 10.0$_{\pm3.7}$ & 18.4$_{\pm0.8}$ & 18.8$_{\pm2.4}$ & 8.2$_{\pm3.8}$ & +10.6$_{\pm4.5}$ \\
\midrule
UniGraph2 & 9.5$_{\pm2.2}$ & 3.8$_{\pm2.4}$ & 15.6$_{\pm3.5}$ & 21.4$_{\pm2.9}$ & 14.8$_{\pm2.1}$ & 8.4$_{\pm1.5}$ & +6.4$_{\pm2.6}$ \\
GFT       & 22.5$_{\pm1.2}$ & 4.5$_{\pm3.1}$ & 2.5$_{\pm0.6}$ & 13.7$_{\pm1.4}$ & 31.0$_{\pm1.2}$ & 6.8$_{\pm4.4}$ & +24.2$_{\pm4.6}$ \\
GIT       & 22.4$_{\pm2.7}$ & 4.0$_{\pm3.1}$ & 2.1$_{\pm0.7}$ & 14.1$_{\pm2.4}$ & 31.6$_{\pm2.4}$ & 6.6$_{\pm4.1}$ & \textbf{+25.0}$_{\pm4.8}$ \\
\bottomrule
\end{tabular}}
\end{minipage}
\vspace{6pt}\par
\begin{minipage}[t]{0.49\linewidth}\centering
\textit{AmazonRatings}\\[1pt]
\resizebox{\linewidth}{!}{%
\begin{tabular}{lcccccc|c}
\toprule
\multirow{2}{*}{\textbf{Method}} & \multicolumn{2}{c}{\textbf{Fid$^+$}} & \multicolumn{2}{c}{\textbf{Fid$^-$}} & \multicolumn{2}{c|}{\textbf{char}} & \multirow{2}{*}{\textbf{$\Delta$}} \\
& G & R & G & R & G & R & \\
\midrule
GCN       & 3.9$_{\pm3.4}$ & 3.5$_{\pm2.1}$ & 17.7$_{\pm4.3}$ & 15.0$_{\pm4.0}$ & 12.0$_{\pm2.6}$ & 8.5$_{\pm2.5}$ & +3.5$_{\pm3.6}$ \\
GAT       & 1.9$_{\pm2.3}$ & 1.1$_{\pm1.0}$ & 5.3$_{\pm6.4}$ & 4.8$_{\pm5.7}$ & 5.6$_{\pm3.7}$ & 3.6$_{\pm1.6}$ & +2.0$_{\pm4.0}$ \\
SAGE      & 18.6$_{\pm5.2}$ & 6.3$_{\pm1.3}$ & 15.2$_{\pm1.5}$ & 29.2$_{\pm6.9}$ & 25.8$_{\pm5.5}$ & 9.8$_{\pm1.6}$ & \underline{+16.0}$_{\pm5.7}$ \\
\midrule
BGRL      & 8.4$_{\pm2.2}$ & 4.9$_{\pm3.3}$ & 22.8$_{\pm0.8}$ & 22.7$_{\pm5.0}$ & 17.3$_{\pm3.1}$ & 11.4$_{\pm4.6}$ & +5.9$_{\pm5.6}$ \\
GraphMAE  & 9.2$_{\pm2.8}$ & 4.2$_{\pm2.2}$ & 15.7$_{\pm0.9}$ & 17.1$_{\pm2.0}$ & 16.4$_{\pm2.7}$ & 9.7$_{\pm2.2}$ & +6.7$_{\pm3.4}$ \\
\midrule
UniGraph2 & 8.1$_{\pm1.6}$ & 4.6$_{\pm4.2}$ & 22.7$_{\pm5.3}$ & 24.4$_{\pm7.1}$ & 17.3$_{\pm1.5}$ & 12.2$_{\pm3.4}$ & +5.0$_{\pm3.7}$ \\
GFT       & 22.0$_{\pm0.9}$ & 6.7$_{\pm2.7}$ & 16.4$_{\pm1.7}$ & 37.1$_{\pm2.1}$ & 26.6$_{\pm0.9}$ & 9.3$_{\pm2.1}$ & \textbf{+17.3}$_{\pm2.3}$ \\
GIT       & 16.9$_{\pm2.0}$ & 6.3$_{\pm1.6}$ & 14.0$_{\pm2.6}$ & 27.4$_{\pm1.6}$ & 24.2$_{\pm2.8}$ & 10.0$_{\pm1.2}$ & +14.2$_{\pm3.1}$ \\
\bottomrule
\end{tabular}}
\end{minipage}\hfill
\begin{minipage}[t]{0.49\linewidth}\centering
\textit{Tolokers}\\[1pt]
\resizebox{\linewidth}{!}{%
\begin{tabular}{lcccccc|c}
\toprule
\multirow{2}{*}{\textbf{Method}} & \multicolumn{2}{c}{\textbf{Fid$^+$}} & \multicolumn{2}{c}{\textbf{Fid$^-$}} & \multicolumn{2}{c|}{\textbf{char}} & \multirow{2}{*}{\textbf{$\Delta$}} \\
& G & R & G & R & G & R & \\
\midrule
GCN       & 11.5$_{\pm0.2}$ & 0.1$_{\pm0.1}$ & 3.0$_{\pm0.6}$ & 4.5$_{\pm0.2}$ & 22.9$_{\pm0.2}$ & 0.4$_{\pm0.2}$ & +22.4$_{\pm0.3}$ \\
GAT       & 0.1$_{\pm0.0}$ & 0.0$_{\pm0.0}$ & 0.0$_{\pm0.0}$ & 0.0$_{\pm0.0}$ & 0.2$_{\pm0.0}$ & 0.0$_{\pm0.0}$ & +0.2$_{\pm0.0}$ \\
SAGE      & 13.6$_{\pm0.3}$ & 0.2$_{\pm0.2}$ & 0.0$_{\pm0.0}$ & 3.3$_{\pm0.2}$ & 23.9$_{\pm0.5}$ & 0.3$_{\pm0.4}$ & \underline{+23.6}$_{\pm0.6}$ \\
\midrule
BGRL      & $-$4.1$_{\pm3.9}$ & $-$0.1$_{\pm0.1}$ & $-$2.3$_{\pm0.9}$ & 1.1$_{\pm3.5}$ & 9.5$_{\pm1.7}$ & 1.4$_{\pm0.4}$ & +8.2$_{\pm1.8}$ \\
GraphMAE  & 0.3$_{\pm0.5}$ & 0.0$_{\pm0.0}$ & 0.0$_{\pm0.1}$ & 0.1$_{\pm0.2}$ & 1.2$_{\pm0.8}$ & 0.0$_{\pm0.0}$ & +1.2$_{\pm0.8}$ \\
\midrule
UniGraph2 & 11.0$_{\pm2.8}$ & 0.9$_{\pm2.2}$ & 12.4$_{\pm0.6}$ & 13.7$_{\pm4.3}$ & 20.1$_{\pm3.6}$ & 5.9$_{\pm1.7}$ & +14.2$_{\pm3.9}$ \\
GFT       & 18.1$_{\pm1.0}$ & 0.2$_{\pm0.3}$ & 0.0$_{\pm0.0}$ & 4.3$_{\pm0.3}$ & 30.6$_{\pm1.4}$ & 0.4$_{\pm0.5}$ & \textbf{+30.2}$_{\pm1.5}$ \\
GIT       & 4.5$_{\pm2.0}$ & 0.1$_{\pm0.1}$ & 0.1$_{\pm0.1}$ & 0.4$_{\pm1.6}$ & 10.1$_{\pm1.0}$ & 0.3$_{\pm0.2}$ & +9.9$_{\pm1.0}$ \\
\bottomrule
\end{tabular}}
\end{minipage}

\end{table}

\begin{figure}[!htbp]
\centering
\includegraphics[width=\linewidth]{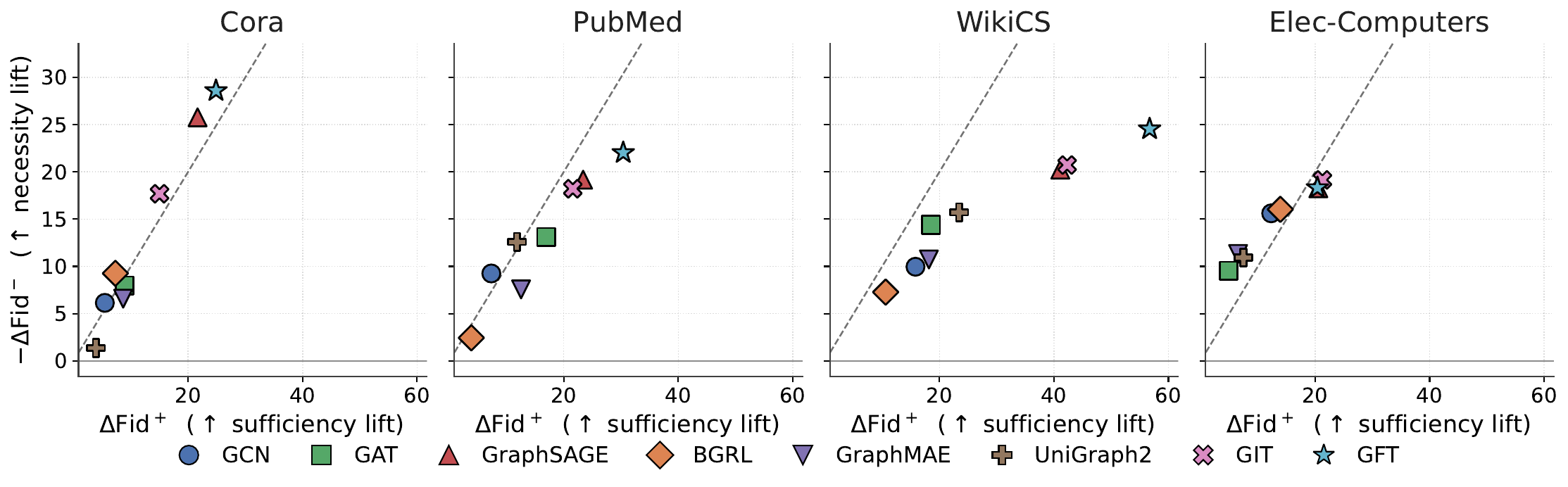}
\caption{\textbf{Node-level saliency lift in fidelity space.}
Each marker denotes one method on a representative NC dataset,
plotted as the improvement of gradient saliency over a random-edge
baseline: $(\Delta\mathrm{Fid}^+,\,-\Delta\mathrm{Fid}^-)$.
The dashed line marks equal lift on the two fidelity components.
Full per-dataset decomposition is in
Table~\ref{app:interp_v2_decomp}.}
\label{fig:interp_main}
\end{figure}


\begin{table}[H]
\centering
\caption{\textbf{Graph-level interpretation: $\Delta_{\text{char}} \times 100$ at top-$k{=}20\%$ atom removal.}
$\Delta_{\text{char}} = \mathrm{char}_{\text{sal}} - \mathrm{char}_{\text{rand}}$, where char is the
GraphFramEx~\citep{amara2022graphframex} harmonic mean of Fid+ and $1\!-\!$Fid$-$, computed via
gradient saliency on FT'd model and node ablation (atom removal with incident edges
dropped, atom excluded from pool; not feature zeroing). Random baseline is mandatory per
GInX-Eval~\citep{amara2023ginxeval}. Higher is better; positive $\Delta_{\text{char}}$ indicates
saliency identifies more predictive atoms than random. Eval set: clean-correct positives
(fallback to all $y{=}1$ when $n_{\mathrm{correct}}{<}10$). Mean$\pm$std over 5 seeds.
``---'' indicates the axis is incompatible with the method's native adaptation pipeline.
\textbf{Bold} / \underline{underline} = best / second per column.}
\label{tab:interpret_graph_main}
\footnotesize
\setlength{\tabcolsep}{4pt}
\begin{tabular}{l|cc}
\toprule
\textbf{Method} & Tox21 & ChemHIV \\
\midrule
GCN & \textbf{59.9}$_{\pm7.2}$ & 2.6$_{\pm1.0}$ \\
GAT & 48.5$_{\pm8.6}$ & 9.7$_{\pm9.0}$ \\
GraphSAGE & \underline{57.0}$_{\pm11.5}$ & 2.4$_{\pm2.1}$ \\
\midrule
BGRL & 37.4$_{\pm6.9}$ & \textbf{33.9}$_{\pm5.7}$ \\
GraphMAE & 41.0$_{\pm9.7}$ & \underline{33.7}$_{\pm3.3}$ \\
\midrule
UniGraph2 & 30.9$_{\pm8.0}$ & 2.7$_{\pm1.1}$ \\
GFT & 45.4$_{\pm16.9}$ & 16.1$_{\pm19.2}$ \\
GIT & 54.7$_{\pm6.9}$ & 4.9$_{\pm1.7}$ \\
\bottomrule
\end{tabular}
\end{table}


\section{Broader impacts}
\label{app:broader_impacts}

\benchname is intended to inform model selection and risk
assessment when graph representation methods are deployed in
domains such as citation triage, recommendation, fraud detection,
and molecular screening. By making per-axis safety failures
visible alongside clean accuracy, the benchmark can help
practitioners avoid overestimating robustness from leaderboard
accuracy alone, and can surface capability gaps that motivate new
robustness, adaptation, and training objectives.

We do not anticipate direct negative societal impacts. The
benchmark relies on publicly available academic datasets, the
stress operators are standard noise and distribution-shift
constructions rather than novel attacks, and no new pretrained
model that meaningfully expands capabilities is released. A
residual risk is that practitioners over-interpret aggregate
rankings as evidence that any single method is uniformly safe;
the paper emphasizes throughout that safety behavior is
condition-dependent and cross-axis comparison is required.



\end{document}